%% file: main.tex
\documentclass{article}

\usepackage{arxiv}

\usepackage[utf8]{inputenc}
\usepackage[T1]{fontenc}
\PassOptionsToPackage{numbers,square,sort&compress}{natbib}
\usepackage{hyperref}
\usepackage{url}
\usepackage{booktabs}
\usepackage{placeins}
\usepackage{float}
\usepackage{amsfonts}
\usepackage{nicefrac}
\usepackage{microtype}
\usepackage{xcolor}
\usepackage{graphicx}
\usepackage{natbib}
\usepackage{doi}

\usepackage{CJKutf8}
\usepackage{enumitem}
\usepackage{algorithm}
\usepackage{algorithmic}
\usepackage{tikz}
\usetikzlibrary{positioning,arrows.meta}
\usepackage{subcaption}
\usepackage{threeparttable}
\usepackage{tabularx}
\usepackage{amsmath}
\usepackage{amssymb}
\usepackage{mathtools}
\usepackage{amsthm}
\usepackage{bbm}

\newcommand{\method}{\mbox{TabCausal}}
\newcommand{\scorestd}[1]{{\textcolor{black!55}{\fontsize{5}{6}\selectfont\,(#1)}}}

\providecommand{\NA}{\text{--}}

\usepackage[capitalize,noabbrev]{cleveref}

\theoremstyle{plain}

\theoremstyle{definition}

\theoremstyle{remark}

\usepackage[textsize=tiny]{todonotes}

\title{\texorpdfstring{\method{}: Pretraining Across Causal Environments for Tabular Causal Discovery}{TabCausal: Pretraining Across Causal Environments for Tabular Causal Discovery}}

\renewcommand{\shorttitle}{\method{}: Pretraining Across Causal Environments for Tabular Causal Discovery}

\author{%
  Zi-Rong Li \\
  Nanjing University
  \And
  Si-Yang Liu \\
  Nanjing University
  \And
  Tian-Zuo Wang \\
  Nanjing University
  \And
  Han-Jia Ye \\
  Nanjing University
  \thanks{Code and benchmark materials: \url{https://github.com/LAMDA-Tabular/TabCausal}.}
}

\hypersetup{
  pdftitle={TabCausal: Pretraining Across Causal Environments for Tabular Causal Discovery},
  pdfauthor={Zi-Rong Li, Si-Yang Liu, Tian-Zuo Wang, Han-Jia Ye},
  pdfkeywords={causal discovery, foundation models, tabular data, pretraining},
}

\begin{document}
\begin{CJK*}{UTF8}{gbsn}

\maketitle

\input{abstract}
\input{introduction}
\input{related_work}
\input{preliminary}
\input{method}
\input{semantic_benchmark}

\input{experiments}

\input{analysis}

\input{conclusion}

\bibliographystyle{unsrtnat}
\bibliography{cite}
\appendix
\input{appendix}

\end{CJK*}
\end{document}

%% file: abstract.tex
\begin{abstract}
Causal discovery aims to recover directed causal relations from observational and interventional data, providing a basis for mechanistic understanding and reliable decision-making. 
Causal discovery foundation models (CDFMs) amortize this problem by mapping a dataset to a causal graph in one forward pass, without per-dataset testing, search, or optimization.
Existing CDFMs still often fall short of strong classical methods. We attribute much of this gap to how causal pretraining tasks are built.
We introduce \method{}, a data-driven CDFM pretrained over diverse graph priors, structural mechanisms, noise models, dimensions, sample sizes, and intervention regimes.
A dynamic task construction strategy composes these environments into varied discovery tasks so structural cues can transfer on observational and mixed-interventional data.
On large-scale synthetic benchmarks, \method{} has better macro-averaged scores than a diverse set of causal discovery baselines.
We also introduce a protocol-guided and LLM-audited semantic causal environment benchmark, where domain-grounded SCMs generate named observational and interventional tables for out-of-distribution analysis.
Across synthetic, semantic, and real-data settings, \method{} recovers structure more reliably when interventional evidence is present.\looseness=-1
\end{abstract}

%% file: introduction.tex
\section{Introduction}
Recovering causal relations from observational and interventional data is a central problem in scientific discovery and reliable decision-making. Across domains such as biology~\citep{conf/nips/ZhangGSSSU23}, Earth science~\citep{runge2019inferring}, and finance~\citep{sadeghi2024causaldiscoveryfinancialmarkets}, researchers often need to know which variables directly influence others, and how interventions propagate. In many such settings the evidence is tabular: columns are variables and rows are observational samples or experimental draws. Tabular causal discovery infers a directed graph of those direct relations, which can then be used for mechanistic interpretation, intervention planning, and downstream modeling~\citep{Peters2017}.

Classical causal discovery methods have developed along several complementary lines, including constraint-based~\citep{books/spirtes2000causation}, score-based~\citep{journals/jmlr/Chickering02a}, and functional-based approaches~\citep{conf/nips/HoyerJMPS08}. These methods infer structure through conditional-independence tests, combinatorial or continuous score optimization, or assumptions about structural equation mechanisms and noise. When their assumptions are well matched and sufficient data are available, they can be highly effective. However, they often rely on restrictive modeling conditions, large sample sizes, carefully tuned hyperparameters, or dataset-specific testing/search/optimization. These requirements make them brittle in complex, nonlinear, high-dimensional, or data-limited regimes~\citep{reviewAll,journals/uhler2013geometry}.

Foundation models for broad AI tasks and for tabular prediction suggest a similar approach for causal discovery: a causal discovery foundation model (CDFM) that reuses structural inference skills across many causal environments. Such a model amortizes dataset-to-graph inference: given a table, optionally with intervention indicators, it predicts directed edge probabilities or an adjacency matrix in one forward pass~\citep{Montagna2024Demystifying}. Early systems train neural predictors on synthetic causal tasks~\citep{lorch2022amortizedinferencecausalstructure}. Tabular foundation models are strong at supervised prediction~\citep{Hollmann2022TabPFN,hollmann2025tabpfn}, which predicts a target column; causal discovery has to recover directions among the columns. Existing CDFMs still do not consistently beat strong classical baselines across mechanisms, regimes, and scales.

We focus on how causal pretraining tasks are constructed. A narrow synthetic generator can teach shortcuts tied to a small family of graph priors, structural equation mechanisms, noise distributions, dimensionalities, or intervention protocols. A CDFM needs many causal environments and task compositions during pretraining if the structural cues are to transfer.

We introduce \method{}, a data-driven foundation model for tabular causal discovery. It takes a table with optional intervention indicators and predicts directed edge probabilities, or a decoded adjacency matrix, in one forward pass without dataset-specific retraining. The core is a broad causal pretraining setup: a causal environment engine jointly samples graph priors, structural equation mechanisms, nonlinear functional families, noise distributions, dimensionalities, sample sizes, and observational/interventional regimes. Dynamic causal task construction then composes these environments into varied discovery tasks so the learned cues still work under distribution shift.

We evaluate \method{} on seven synthetic families under observational and mixed-interventional regimes, from small graphs to high-dimensional cases. Against classical, neural, and amortized baselines, it has the best overall average rank in both regimes, with the largest gains when interventions are present. We also introduce a domain-grounded semantic causal environment benchmark: blueprint-guided scene specifications are compiled into SCMs with known DAGs and named variables, then turned into observational and mixed-interventional tables after LLM validation. Additional checks use public tables with complete directed ground truth, fully held-out synthetic OOD families absent from pretraining, and sample-size and dimension scaling. Broad causal pretraining helps structure recovery in these settings; large graphs and hard observational cases remain difficult.

Our contributions are:
\begin{itemize}[leftmargin=*,nosep]
    \item We treat how large and diverse the pretraining environments are, and how those tasks are built, as a design problem for causal discovery foundation models.
    \item We propose \method{}, a tabular CDFM with a broad causal environment engine and dynamic task construction, mapping values and optional intervention indicators to a graph in one pass.\looseness=-1
    \item We evaluate on synthetic, semantic, real-data, scaling, and held-out OOD settings, and introduce a domain-grounded semantic causal environment benchmark with SCM-controlled ground truth.
\end{itemize}

%% file: related_work.tex
\section{Related Work}
\label{sec:related}

\noindent{\bf Classical causal discovery methods}.
Classical causal discovery methods are commonly grouped into three classes~\citep{VowelsCB23Survey}: constraint-based approaches grounded in conditional-independence (CI) relations~\citep{books/spirtes2000causation}, score-based approaches that optimize a graph score over a structure space~\citep{journals/jmlr/Chickering02a}, and functional-based approaches that identify directions under additional assumptions on the data-generating mechanisms~\citep{conf/nips/HoyerJMPS08}.
A representative constraint-based method is the PC algorithm, which first recovers an undirected skeleton via CI tests and then applies orientation rules to obtain a partially directed equivalence-class representation~\citep{CPDAG}.
When interventional information is available, distribution shifts induced by interventions can further improve orientability and identifiability; this idea is formalized in methods such as IGSP~\citep{wang2017permutationbasedcausalinferencealgorithms}.
Score-based methods maximize decomposable scores via search; foundational greedy analyses appear in Chickering~\citep{journals/jmlr/Chickering02a}, with GIES~\citep{hauser2012characterizationgreedylearninginterventional} extending score-based search to interventional settings; without intervention targets, GIES reduces to the observational GES procedure.
To reduce explicit search, another line of work formulates structure learning as differentiable or continuous optimization and learns an adjacency matrix directly with gradient-based methods.
NOTEARS~\citep{zheng2018dagstearscontinuousoptimization} is a landmark example, introducing a differentiable acyclicity constraint combined with sparsity regularization, and it inspired variants with alternative objectives or constraints such as NoDAGS~\citep{sethuraman2023nodagsflownonlinearcycliccausal}.
DAGMA~\citep{NEURIPS2022_36e2967f} uses an M-matrix/log-determinant acyclicity characterization for continuous optimization of DAG structure, while SDCD~\citep{nazaret2024stabledifferentiablecausaldiscovery} introduces a more stable spectral acyclicity constraint and a sparse-graph-oriented two-stage training procedure that can scale to larger graphs.
While these methods can be effective in low-to-moderate dimensions when assumptions are well matched, they often struggle in high-dimensional and finite-sample regimes due to the heavy computational burden of repeated tests and the statistical fragility of CI or score estimation~\citep{colombo2013orderindependentconstraintbasedcausalstructure,journals/uhler2013geometry}.
Functional-based methods, such as LiNGAM~\citep{lingam}, can identify directions under linear non-Gaussian independent noise, but may be sensitive to assumption mismatch, optimization details, and hyperparameter choices such as penalties and sparsification thresholds.
Varsortability can also make simulated DAG benchmarks easier than intended~\citep{NEURIPS2021_e987eff4}, motivating ordering/regression control baselines; we therefore include RandomRegress in our evaluation as such a control.\looseness=-1

\noindent{\bf Tabular foundation models and causal discovery foundation models}.
Tabular foundation models, exemplified by TabPFN~\citep{Hollmann2022TabPFN,hollmann2025tabpfn}, show that large-scale pretraining over synthetic tabular tasks yields strong in-context predictors for supervised learning.
They are built to predict labels or targets from feature tables. Tabular causal discovery outputs a directed graph over the variables themselves, so a strong predictor of a target column does not by itself recover edge orientation or use intervention information.

Recent amortized causal discovery maps observational and/or interventional datasets directly to edge probabilities or adjacencies in one forward pass~\citep{Montagna2024Demystifying}.
AVICI~\citep{lorch2022amortizedinferencecausalstructure} trains such a predictor on synthetic graphs and mechanisms.
SEA (Sample, Estimate, Aggregate)~\citep{wu2025sampleestimateaggregaterecipe} trains a supervised aggregator on synthetic causal tasks to combine summary statistics and estimates from classical discovery algorithms run on sampled variable subsets, yielding a pretrained causal discovery model that can predict graphs on new datasets.
Recent pretrained or foundation-model-style causal discovery methods have also begun to appear, including BCNP~\citep{ICLR2025_24faedc5}, Arrow~\citep{thompson2026arrowfoundationmodelcausal}, and CauScale~\citep{peng2026causcaleneuralcausaldiscovery}.
At the time of our evaluation, however, these methods did not provide publicly usable checkpoints/weights. We therefore report AVICI and SEA among pretrained baselines because their released implementation could be integrated into our evaluation harness.
We follow this CDFM paradigm, with the emphasis on pretraining task construction: broader environments, composed tasks, and evaluation on anonymous synthetic benchmarks, semantic causal environments, public real datasets, scaling analyses, and held-out synthetic OOD tests.

%% file: preliminary.tex
\section{Preliminaries}
\label{sec:prelim}

\subsection{Causal discovery: inputs, outputs, and evaluation}
Causal discovery aims to recover the direct causal relations among a set of features from data, typically represented as a directed acyclic graph (DAG). Let the feature index set be $V \coloneqq \{1,\ldots,d\}$ and the causal graph be $G=(V,E)$ with $E \subseteq V \times V$. We use an adjacency matrix $A \in \{0,1\}^{d \times d}$ to represent $G$, where $A_{ij}=1$ indicates a directed edge $i \to j$ (and $A_{ii}=0$).

At the mechanism level, we consider a structural causal model (SCM)~\citep{Pearl2009}: each feature $X_j$ is generated from its parents $\mathrm{pa}(j)\coloneqq \{i \in V : i \to j \in E\}$ via
\begin{equation}
    X_j \;=\; f_j\!\big(X_{\mathrm{pa}(j)},\, \epsilon_j\big), \qquad j=1,\ldots,d,
    \label{eq:scm}
\end{equation}
where $\epsilon_j$ is a noise term. Observational data are given as a table of $N$ rows and $d$ columns, where each row is one sample:
\begin{equation}
    \mathcal{D}^{\mathrm{obs}} \;=\; \{\mathbf{x}^{(n)}\}_{n=1}^{N}, \qquad \mathbf{x}^{(n)} \in \mathbb{R}^{d}.
    \label{eq:obs_data}
\end{equation}
We also consider mixed observational--interventional inputs, where each sample is accompanied by an intervention indicator vector. Concretely, for sample $n$ and feature $j$, let $m^{(n)}_j\in\{0,1\}$ denote whether $X_j$ is intervened on when generating $\mathbf{x}^{(n)}$. The input can be written as
\begin{equation}
    \mathcal{D} \;=\; \{(\mathbf{x}^{(n)},\, \mathbf{m}^{(n)})\}_{n=1}^{N}, \qquad \mathbf{m}^{(n)} \in \{0,1\}^{d},
    \label{eq:mixed_data}
\end{equation}
where $\mathbf{m}^{(n)}=\mathbf{0}$ for purely observational samples. The causal discovery task is to estimate the underlying structure (e.g., $A$ or edge-direction probabilities) from $\mathcal{D}$. We evaluate the predicted directed edge set $\widehat{E}$ against the ground-truth $E^\star$ using directed edge-level $F_1$ as well as structural distances, specifically Structural Hamming Distance (SHD) and Structural Intervention Distance (SID)~\citep{peters2014structuralinterventiondistancesid}.\looseness=-1

\subsection{Identifiability scope}
From purely observational data, a DAG is generally identifiable only up to its Markov equivalence class under standard Markov and faithfulness assumptions~\citep{books/spirtes2000causation,CPDAG,Peters2017}.
Recovering a unique direction therefore requires additional assumptions, such as linear non-Gaussian noise in LiNGAM~\citep{lingam} or specific nonlinear function/noise restrictions in post-nonlinear models; even post-nonlinear causal models contain non-identifiable cases~\citep{zhang2009identifiabilitypnl}.
\method{} does not impose a specific functional-form assumption. It is pretrained over a broad SCM prior, so we do not claim a unique DAG for arbitrary observational mechanisms.
In observation-only settings it uses structural cues induced by that prior; in mixed-interventional settings the masks and perturbed samples add distributional information that can shrink equivalence-class ambiguity.

\subsection{Foundation-model viewpoint: amortized dataset-to-graph inference}
A causal discovery foundation model reframes structure learning as a shared, reusable mapping from tabular datasets to directed graphs.
For inference, we encode each dataset $\mathcal{D}$ as a value--mask tensor $\mathbf{X}\in\mathbb{R}^{N\times d\times 2}$ whose channels stack observed values and per-cell intervention indicators (Section~\ref{sec:method_arch}; equivalently $(\mathbf{X}^{\mathrm{val}},\mathbf{M})$ with $\mathbf{X}^{\mathrm{val}}\in\mathbb{R}^{N\times d}$ and $\mathbf{M}\in\{0,1\}^{N\times d}$).
A foundation model is a parameterized predictor $f_\theta$ that maps $\mathbf{X}$ to directed edge probabilities in one forward pass:
\begin{equation}
    \widehat{\mathbf{P}} \;=\; f_{\theta}(\mathbf{X}), \qquad \widehat{\mathbf{P}} \in (0,1)^{d\times d},
    \label{eq:fm_map}
\end{equation}
with an optional binary adjacency obtained by thresholding off-diagonal entries of $\widehat{\mathbf{P}}$.

The same $f_\theta$ is used for different sample sizes $N$, feature counts $d$, and data-generating mechanisms, and can read intervention channels when they are present. Inference is one forward pass on each new dataset.
The training distribution is part of the model: a CDFM learns structural cues from many pretraining tasks, so the range of graphs, mechanisms, noise, dimensions, sample sizes, and intervention regimes affects what transfers to unseen tables.

%% file: method.tex
\section{Method}
\label{sec:method}

\subsection{Model architecture}
\label{sec:method_arch}

\begin{figure}[!htbp]
\centering
\includegraphics[width=0.94\linewidth]{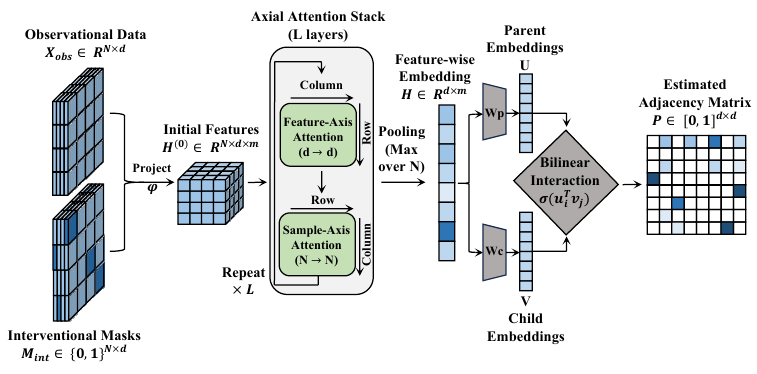}
\vspace{-3mm}
\caption{\textbf{Model architecture overview.}
Input embedding, axial attention encoder (alternating over features and samples), and a directed edge head that maps pooled feature tokens to edge probabilities.}
\vspace{-3mm}
\label{fig:architecture}
\end{figure}

\method{} performs amortized dataset-to-graph inference: it maps a tabular dataset with optional per-cell intervention indicators to directed edge scores in one forward pass. The input is a tensor $\mathbf{X}\in\mathbb{R}^{N\times d\times 2}$ whose last channel stacks observed values and binary intervention indicators (equivalently $(\mathbf{X}^{\mathrm{val}},\mathbf{M})$ with $\mathbf{X}^{\mathrm{val}}\in\mathbb{R}^{N\times d}$, $\mathbf{M}\in\{0,1\}^{N\times d}$). The network outputs $\widehat{\mathbf{P}}\in(0,1)^{d\times d}$; an optional binary adjacency follows from thresholding off-diagonal entries. The same backbone accepts purely observational data (all-zero masks) and mixed observational--interventional batches. Figure~\ref{fig:architecture} summarizes the layout; Appendix~\ref{sec:appendix-architecture} gives additional detail.

A shared linear map $\phi$ embeds each value--indicator pair into width $m$,
\begin{equation}
    \mathbf{H}^{(0)}=\phi(\mathbf{X})\in\mathbb{R}^{N\times d\times m}.
    \label{eq:embed}
\end{equation}
The encoder stacks $L$ blocks that alternate multi-head self-attention over the feature axis and over the sample axis (with residuals and feed-forward layers), capturing cross-feature structure within rows and cross-sample regularities within columns~\citep{vaswani2023attentionneed,ho2019axialattentionmultidimensionaltransformers}. Schematically,
\begin{equation}
    \mathbf{H}^{(\ell)}=\mathrm{Attn}_{\text{samp}}\!\Big(\mathrm{Attn}_{\text{var}}\big(\mathbf{H}^{(\ell-1)}\big)\Big),\quad \ell=1,\ldots,L.
    \label{eq:axial_stack}
\end{equation}
We max-pool over samples to obtain feature tokens $\widetilde{\mathbf{H}}\in\mathbb{R}^{d\times m}$~\citep{zaheer2018deepsets}, apply learnable parent and child projections $\mathbf{W}_{\mathrm{p}},\mathbf{W}_{\mathrm{c}}$,
\begin{equation}
    \mathbf{U}=\widetilde{\mathbf{H}}\mathbf{W}_{\mathrm{p}},\qquad
    \mathbf{V}=\widetilde{\mathbf{H}}\mathbf{W}_{\mathrm{c}},
    \label{eq:parent_child}
\end{equation}
and score each ordered pair $(i,j)$ with normalized cosine similarity, learnable temperature $\tau$, and bias $b$~\citep{dozat2017deepbiaffineattentionneural,lorch2022amortizedinferencecausalstructure},\looseness=-1
\begin{equation}
    \widehat{S}_{ij}= e^{\tau}\Big\langle \tfrac{\mathbf{u}_i}{\|\mathbf{u}_i\|_2},\tfrac{\mathbf{v}_j}{\|\mathbf{v}_j\|_2}\Big\rangle + b,\qquad
    \widehat{P}_{ij}=\sigma(\widehat{S}_{ij}),\qquad
    \widehat{A}_{ij}=\mathbbm{1}[\widehat{P}_{ij}\ge \tau_{\mathrm{thr}}]\ (i\neq j),\ \widehat{A}_{ii}=0.
    \label{eq:edge_prob}
\end{equation}
Here $\tau_{\mathrm{thr}}$ is a decoding threshold (distinct from $\tau$).

\subsection{Broad causal pretraining}
\label{sec:method_pretrain}

\begin{figure}[H]
\centering
\includegraphics[width=0.85\linewidth]{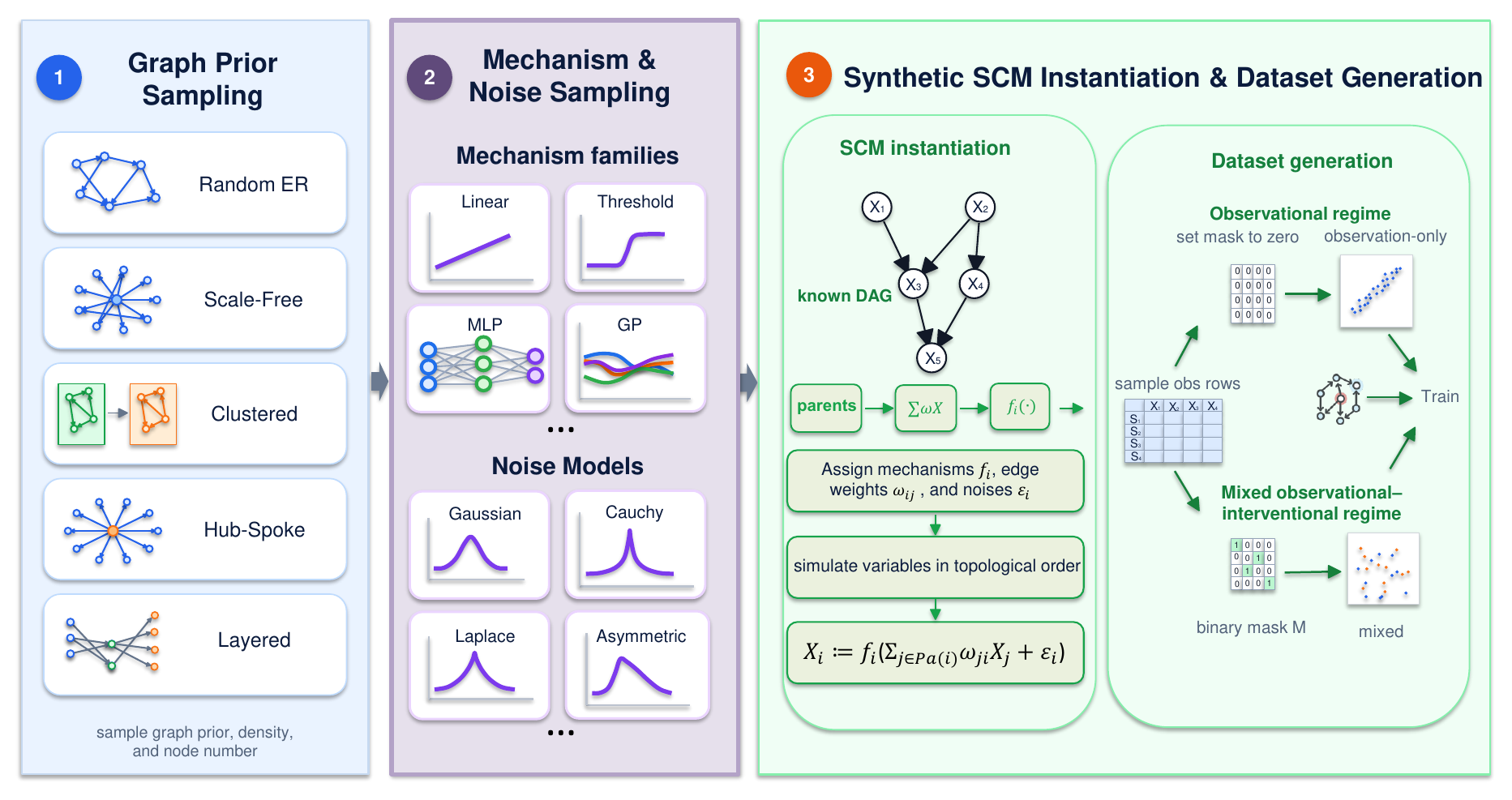}
\vspace{-2mm}
\caption{\textbf{Causal pretraining engine.}
The engine samples graph priors, mechanisms, noise models, dimensions, sample sizes, and regimes; builds SCMs with known DAGs; and emits observational (all-zero masks) or mixed-interventional (binary masks) value--mask tensors with simulator-known adjacencies for supervised edge prediction.}
\label{fig:pretrain-pipeline}
\end{figure}

Figure~\ref{fig:pretrain-pipeline} summarizes the engine formalized below.
\vspace{-6pt}
\vspace{4pt}

Let $\mathcal{G}$, $\mathcal{F}$, and $\mathcal{P}_{\epsilon}$ denote catalogs of graph priors, structural-equation mechanism families, and exogenous noise models, respectively; let $\mathcal{D}_{\mathrm{dim}}$ and $\mathcal{N}_{\mathrm{samp}}$ denote admissible feature dimensions~$d$ and dataset sizes~$N$; and let $\mathcal{R}=\{\mathrm{obs},\mathrm{mix}\}$ encode observational versus mixed observational--interventional regimes.
Each pretraining episode draws a configuration
\begin{equation}
  z=(g,f,\rho,d,N,r)\sim \mathcal{S},
  \qquad
  g\in\mathcal{G},\;
  f\in\mathcal{F},\;
  \rho\in\mathcal{P}_{\epsilon},\;
  d\in\mathcal{D}_{\mathrm{dim}},\;
  N\in\mathcal{N}_{\mathrm{samp}},\;
  r\in\mathcal{R},
  \label{eq:pretrain_episode}
\end{equation}
where $\mathcal{S}$ denotes the engine's joint resampling rule over these factors (we leave marginals abstract here).
Given $z$, the simulator instantiates a structural causal model $\mathcal{M}_z$ with simulator-known adjacency $A^\star$,
\begin{equation}
  (A^\star,\mathcal{M}_z)=\mathrm{SCM}(z),
\end{equation}
where $\mathcal{M}_z$ denotes the graph and structural equations for configuration $z$.
It then generates observations and a binary intervention mask $\mathbf{M}$ consistent with regime~$r$ and sample budget~$N$,
\begin{equation}
  (\mathbf{X}^{\mathrm{val}},\mathbf{M})\sim \mathrm{Gen}(\mathcal{M}_z,r,N),
  \qquad
  \mathbf{X}=[\mathbf{X}^{\mathrm{val}},\mathbf{M}]\in\mathbb{R}^{N\times d\times 2}.
\end{equation}
The supervised pretraining tuple is $\upsilon=(\mathbf{X},A^\star)$.
Concrete instantiations of $\mathcal{G},\mathcal{F},\mathcal{P}_{\epsilon},\mathcal{D}_{\mathrm{dim}},\mathcal{N}_{\mathrm{samp}}$, and $\mathcal{S}$ are listed in Appendix~\ref{appendix:pretrain_data}.

When $r=\mathrm{obs}$, $\mathbf{M}$ is all-zero; when $r=\mathrm{mix}$, $\mathbf{M}$ marks intervened coordinates on interventional rows while observational rows remain zero, following the layout in Section~\ref{sec:method_arch}.
Each episode jointly resamples $(g,f,\rho,d,N,r)$, which widens the support of the pretraining task distribution.
Interventional counts and compositions vary across episodes so that context size and perturbation structure are not memorized surface cues.

Pretraining minimizes the expected directed edge-wise BCE over tasks sampled by this engine: letting $\mathcal{P}_{\mathrm{train}}$ denote the distribution of $(\mathbf{X},A^\star)$ induced by $z\sim\mathcal{S}$ followed by $\mathrm{Gen}$, we minimize the average loss between $A^\star$ and $\widehat{\mathbf{P}}=f_\theta(\mathbf{X})$ over off-diagonal ordered pairs,
\begin{equation*}
\min_{\theta}\mathbb{E}_{(\mathbf{X},A^\star)\sim\mathcal{P}_{\mathrm{train}}}\!\big[\mathcal{L}_{\mathrm{BCE}}(\theta)\big],\quad
\mathcal{L}_{\mathrm{BCE}}(\theta)=\frac{1}{d(d-1)}\sum_{i\ne j}\!\left[-A^\star_{ij}\log \widehat{P}_{ij}-(1-A^\star_{ij})\log(1-\widehat{P}_{ij})\right].
\label{eq:pretrain_obj}
\end{equation*}
At test time, we apply the same amortized map without dataset-specific fine-tuning and threshold $\widehat{\mathbf{P}}$ when a discrete graph is needed (Eq.~\eqref{eq:edge_prob}).
Training uses no explicit differentiable acyclicity constraint; when strict DAG outputs are required at evaluation time, we optionally apply lightweight cycle pruning as inference-only decoding post-processing.\looseness=-1

%% file: semantic_benchmark.tex
\section{Semantic Causal Environment Benchmark}
\label{sec:semantic-benchmark}

\subsection{Benchmark design}
Anonymous synthetic benchmarks give simulator-known graphs with little variable-level meaning, which makes failure cases hard to read.
We introduce a domain-grounded semantic causal environment benchmark: the graphs are still SCM-generated, and the variables and mechanisms have explicit domain names.
Human-specified domain blueprints, mechanism lists, graph and intervention constraints, and validation rules set the authoring space; LLM-assisted generation then fills scenario specifications with named variables, directed edges, edge rationales, and intervention handles.
Those specifications are checked against the human rules, reviewed in a separate LLM consistency audit, and compiled into SCMs with known DAGs.
The generator emits observational and mixed-interventional tables.
The suite has 100 scenarios across 10 domains under \texttt{obs} and \texttt{int}, hence 200 dataset--regime tasks, and is evaluated alongside the anonymous synthetic benchmark in Section~\ref{sec:exp}.
Appendix~\ref{app:semantic-benchmark} documents the inventory, schema, sampling and generation rules, and the LLM-aided audit protocol (\cref{app:semantic-llm-audit}).

\subsection{Aggregate semantic benchmark results}
Table~\ref{tab:semantic-summary} aggregates semantic benchmark performance by regime; domain- and motif-level breakdowns appear in Appendix~\ref{app:semantic-benchmark} (Figure~\ref{fig:semantic-domain-heatmap} and Table~\ref{tab:semantic-slice-breakdown}).
In observation-only settings \method{} is second in F1 and best in SHD among applicable methods; GIES has the highest observation-only F1.
With mixed interventions, \method{} is best in F1, SHD, and SID among applicable methods.
Mixed-interventional rankings follow the anonymous synthetic tables, with a clearer gap for \method{}.

\input{figs/tables/semantic/table_semantic_summary}

%% file: figs/tables/semantic/table_semantic_summary.tex
\begin{table}[!htbp]
\centering
\footnotesize
\setlength{\tabcolsep}{3.5pt}
\renewcommand{\arraystretch}{1.0}
\begin{threeparttable}
\caption{\textbf{Semantic causal environment benchmark by regime.} Mean F1, SHD, and SID are reported separately for observation-only and mixed-interventional semantic tasks. Values are mean (standard deviation).}
\label{tab:semantic-summary}
\begin{tabular*}{\textwidth}{@{\extracolsep{\fill}}lccc@{\hspace{8pt}}ccc@{}}
\toprule
& \multicolumn{3}{c}{\textbf{Observation-only}} & \multicolumn{3}{c}{\textbf{Mixed-interventional}} \\
\cmidrule(lr){2-4}\cmidrule(lr){5-7}
\textbf{Method} & \textbf{F1}$\uparrow$ & \textbf{SHD}$\downarrow$ & \textbf{SID}$\downarrow$ & \textbf{F1}$\uparrow$ & \textbf{SHD}$\downarrow$ & \textbf{SID}$\downarrow$ \\
\midrule
RandomRegress & 0.206\scorestd{0.126} & 24.23\scorestd{18.38} & 48.19\scorestd{16.10} & \NA & \NA & \NA \\
DAS & 0.242\scorestd{0.098} & 17.54\scorestd{6.43} & 62.47\scorestd{28.50} & \NA & \NA & \NA \\
LiNGAM & 0.303\scorestd{0.175} & 11.28\scorestd{4.40} & 41.06\scorestd{18.54} & \NA & \NA & \NA \\
PC & 0.649\scorestd{0.171} & 6.71\scorestd{4.06} & 35.23\scorestd{25.14} & \NA & \NA & \NA \\
CDIS & 0.595\scorestd{0.164} & 7.78\scorestd{3.85} & 38.09\scorestd{24.14} & 0.604\scorestd{0.166} & 7.52\scorestd{4.04} & 37.02\scorestd{22.48} \\
GIES & \textbf{0.698}\scorestd{0.209} & 7.34\scorestd{5.80} & \underline{27.63}\scorestd{32.54} & 0.882\scorestd{0.079} & 3.86\scorestd{3.09} & \underline{3.95}\scorestd{6.76} \\
IGSP & 0.664\scorestd{0.184} & \underline{6.27}\scorestd{4.26} & 30.70\scorestd{22.72} & 0.611\scorestd{0.231} & 7.36\scorestd{5.33} & 34.57\scorestd{26.94} \\
DAGMA & 0.276\scorestd{0.139} & 11.33\scorestd{4.07} & 51.57\scorestd{23.33} & \NA & \NA & \NA \\
NOTEARS & 0.203\scorestd{0.129} & 11.92\scorestd{4.11} & 43.58\scorestd{18.41} & \NA & \NA & \NA \\
NOTEARS-MLP & 0.270\scorestd{0.130} & 11.55\scorestd{4.20} & 53.92\scorestd{23.25} & \NA & \NA & \NA \\
NoDAGS & \NA & \NA & \NA & 0.864\scorestd{0.124} & 3.19\scorestd{3.10} & 13.97\scorestd{13.87} \\
SEA & 0.471\scorestd{0.137} & 8.51\scorestd{3.55} & 45.27\scorestd{21.82} & 0.321\scorestd{0.122} & 16.02\scorestd{13.32} & 57.89\scorestd{28.37} \\
SDCD & 0.479\scorestd{0.151} & 11.13\scorestd{6.35} & 45.53\scorestd{24.39} & 0.780\scorestd{0.186} & 6.58\scorestd{7.81} & 19.78\scorestd{22.02} \\
DCDI & 0.338\scorestd{0.121} & 32.38\scorestd{25.33} & 69.10\scorestd{43.40} & 0.435\scorestd{0.181} & 32.97\scorestd{26.78} & 70.13\scorestd{49.09} \\
AVICI & 0.560\scorestd{0.187} & 8.30\scorestd{4.43} & 34.54\scorestd{19.02} & \underline{0.943}\scorestd{0.078} & \underline{1.70}\scorestd{2.56} & 6.39\scorestd{10.61} \\
\midrule
\method{} & \underline{0.681}\scorestd{0.173} & \textbf{6.07}\scorestd{3.34} & \textbf{27.45}\scorestd{19.51} & \textbf{0.971}\scorestd{0.057} & \textbf{0.88}\scorestd{1.55} & \textbf{1.89}\scorestd{6.92} \\
\bottomrule
\end{tabular*}
\end{threeparttable}
\end{table}

%% file: experiments.tex
\section{Experiments}
\label{sec:exp}

\subsection{Experimental setup}
\label{sec:exp_setup}

We evaluate seven synthetic families---\texttt{gp\_hard}, \texttt{gp\_simple}, \texttt{linear\_gauss}, \texttt{linear\_graph}, \texttt{linear\_nongauss}, \texttt{mul\_noise}, and \texttt{pfn}---under Observation-only (1000 observational samples) and Mixed-interventional (800 observational + 200 interventional samples), for graph sizes $d\in\{5,10,20\}$.
These are seven data-generation engines: two Gaussian-process nonlinear engines (\texttt{gp\_hard}, \texttt{gp\_simple}), three linear engines with different noise/graph assumptions (\texttt{linear\_gauss}, \texttt{linear\_graph}, \texttt{linear\_nongauss}), a multiplicative-noise engine (\texttt{mul\_noise}), and a PFN-style pretraining engine (\texttt{pfn}) that samples tasks from a prior-fitting-network-like synthetic task distribution.
In later tables and figures, tags such as \texttt{gp\_hard\_obs} or \texttt{pfn\_int} combine a family/engine prefix with a regime suffix: \texttt{\_obs} denotes observation-only data and \texttt{\_int} denotes mixed observational--interventional data.
Appendix~\ref{app:benchmark} specifies benchmark datasets; Section~\ref{sec:semantic-benchmark} summarizes the semantic benchmark.
We also evaluate public tabular datasets with complete directed ground truth (Section~\ref{sec:exp_real_data}) and fully held-out synthetic OOD tests whose graph, mechanism, root, and noise families are absent from pretraining, including alternative intervention protocols (Appendix~\ref{app:heldout-ood}).

Baselines include AVICI~\citep{lorch2022amortizedinferencecausalstructure}, NOTEARS, NOTEARS-MLP~\citep{zheng2018dagstearscontinuousoptimization}, IGSP~\citep{wang2017permutationbasedcausalinferencealgorithms}, CDIS~\citep{dai2025selectionmeetsinterventionadditional}, GIES~\citep{hauser2012characterizationgreedylearninginterventional}, LiNGAM~\citep{lingam}, DCDI~\citep{brouillard2020differentiablecausaldiscoveryinterventional}, NoDAGS~\citep{sethuraman2023nodagsflownonlinearcycliccausal}, PC~\citep{books/spirtes2000causation}, DAS~\citep{pmlr-v213-montagna23b}, SEA~\citep{wu2025sampleestimateaggregaterecipe}, DAGMA~\citep{NEURIPS2022_36e2967f}, SDCD~\citep{nazaret2024stabledifferentiablecausaldiscovery}, and RandomRegress~\citep{NEURIPS2021_e987eff4}.
We apply the same GIES implementation in both regimes: on observation-only datasets it receives no intervention targets and therefore reduces to observational GES, whereas on mixed-interventional datasets we pass targets derived from the intervention masks (details in Appendix~\ref{sec:appendix_hyperparams}).
We report edge-level F1 (higher is better), SHD (lower is better), and SID (lower is better).
A uniform five-minute per-graph wall-clock budget is shared across methods. The suite has thousands of graphs per method, and unbounded per-graph search or optimization is not workable at that scale.
Similar caps appear elsewhere; GLIP uses 300 seconds against PC and NOTEARS~\citep{kook2026exactgraph}.
On the main synthetic and semantic evaluations, almost all methods finish within this budget.
DCDI is an exception: its per-graph neural optimization typically takes several hours, so when the five-minute limit is reached we report the graph at an early-stopped iterate.

For each dataset setting $(\text{family},\text{regime},d)$, we evaluate 50 independently generated graph instances and report mean and standard deviation under fixed random seeds.
Orientation conversions for PC, GIES, and CDIS; thresholding/decoding for probabilistic outputs; and other wrapper defaults are unified as in Appendix~\ref{sec:appendix_hyperparams}.

\subsection{Synthetic data benchmark}
\label{sec:exp_main}

On the synthetic benchmark, \method{} has the strongest overall average rank in both observational and mixed-interventional regimes.
The mixed-interventional gap is the largest: \method{} has the best aggregate F1, SHD, and SID among applicable methods, consistent with using the intervention masks after broad causal pretraining.
On observational data it stays competitive across mechanism families; a specialized baseline can still win when its assumptions match the generator closely.
Table~\ref{tab:main-benchmark} reports macro-averaged scores; missing entries indicate regime-inapplicable methods.
\input{figs/tables/main/table_main_benchmark}
Figure~\ref{fig:f1-heatmap} in the appendix visualizes per-configuration F1 across dataset-regime columns for the full baseline pool.
Sample-size and high-dimensional scaling analyses appear in Section~\ref{sec:analysis}.
A fully held-out synthetic OOD split and alternative intervention protocols are reported in Appendix~\ref{app:heldout-ood}.

\subsection{Real-data causal-discovery check}
\label{sec:exp_real_data}

We also score public tabular causal-discovery datasets that have complete directed ground-truth graphs.
We use all datasets from the selected sources that match our static tabular DAG-discovery setting: six CD-CSG observation-only datasets~\citep{zhao2023causalstar}, Sachs flow-cytometry data~\citep{sachs}, four Amazon PetShop operational-telemetry scenarios~\citep{hardt2024petshop}, and two Causal Chambers physical-system datasets~\citep{gamella2024causalchambers}.
For observation-only evaluation, the seven interventional datasets are also converted by dropping interventional rows, yielding 13 observation-only datasets; interventional results average over the seven datasets with intervention records.
Table~\ref{tab:real-data-full} reports the completed results for this real-data check, including the completed PC observation-only run over all 13 datasets.

\begin{table}[!htbp]
\centering
\footnotesize
\setlength{\tabcolsep}{3.5pt}
\renewcommand{\arraystretch}{0.95}
\vspace{-2pt}
\begin{threeparttable}
\caption{\textbf{Real-data benchmark.} Mean F1, SHD, and SID are reported separately for observation-only and mixed-interventional real datasets. Values are mean (standard deviation).}
\label{tab:real-data-full}
\begin{tabular*}{\textwidth}{@{\extracolsep{\fill}}lccc@{\hspace{8pt}}ccc@{}}
\toprule
& \multicolumn{3}{c}{\textbf{Observation-only}} & \multicolumn{3}{c}{\textbf{Mixed-interventional}} \\
\cmidrule(lr){2-4}\cmidrule(lr){5-7}
\textbf{Method} & \textbf{F1}$\uparrow$ & \textbf{SHD}$\downarrow$ & \textbf{SID}$\downarrow$ & \textbf{F1}$\uparrow$ & \textbf{SHD}$\downarrow$ & \textbf{SID}$\downarrow$ \\
\midrule
RandomRegress & 0.24\scorestd{0.14} & 60.46\scorestd{56.77} & 145.85\scorestd{179.18} & \NA & \NA & \NA \\
DAS & 0.18\scorestd{0.24} & 56.08\scorestd{48.58} & 119.62\scorestd{122.27} & \NA & \NA & \NA \\
LiNGAM & 0.13\scorestd{0.17} & \textbf{24.31}\scorestd{17.78} & \textbf{50.46}\scorestd{49.02} & \NA & \NA & \NA \\
PC & 0.18\scorestd{0.14} & 28.85\scorestd{21.05} & 63.85\scorestd{55.24} & \NA & \NA & \NA \\
CDIS & 0.25\scorestd{0.25} & 27.85\scorestd{21.24} & 55.62\scorestd{55.39} & 0.14\scorestd{0.18} & 45.43\scorestd{10.53} & 94.86\scorestd{50.53} \\
GIES & \underline{0.29}\scorestd{0.07} & 47.92\scorestd{43.41} & 164.31\scorestd{209.73} & \underline{0.29}\scorestd{0.12} & 118.57\scorestd{53.34} & 399.00\scorestd{212.40} \\
IGSP & 0.17\scorestd{0.22} & 31.00\scorestd{23.42} & 67.62\scorestd{73.48} & 0.11\scorestd{0.13} & 59.14\scorestd{23.80} & 154.71\scorestd{125.13} \\
NOTEARS & 0.16\scorestd{0.15} & 29.08\scorestd{24.02} & 57.62\scorestd{49.50} & \NA & \NA & \NA \\
NOTEARS-MLP & 0.21\scorestd{0.19} & 55.00\scorestd{106.03} & 146.77\scorestd{300.84} & \NA & \NA & \NA \\
NoDAGS & \NA & \NA & \NA & 0.02\scorestd{0.02} & \underline{41.00}\scorestd{10.60} & \textbf{85.14}\scorestd{34.49} \\
DCDI & 0.16\scorestd{0.20} & 122.08\scorestd{237.49} & 232.46\scorestd{457.02} & 0.13\scorestd{0.12} & 540.86\scorestd{273.27} & 1116.43\scorestd{525.04} \\
AVICI & 0.22\scorestd{0.18} & \underline{26.00}\scorestd{20.71} & \underline{55.38}\scorestd{52.43} & 0.28\scorestd{0.12} & 42.57\scorestd{12.15} & \underline{88.14}\scorestd{48.95} \\
\midrule
\method{} & \textbf{0.32}\scorestd{0.20} & 29.08\scorestd{22.25} & 84.31\scorestd{104.86} & \textbf{0.42}\scorestd{0.12} & \textbf{40.29}\scorestd{10.14} & 147.14\scorestd{125.36} \\
\bottomrule
\end{tabular*}
\end{threeparttable}
\end{table}

Averaged over all applicable real-data settings, \method{} also has the best overall F1.
\method{} has the best observation-only F1, mixed-interventional F1, and mixed-interventional SHD.
Observation-only SHD remains competitive for sparse or strongly constrained estimators: LiNGAM is best, AVICI is second, and CDIS and PC sit near \method{} after completing all 13 observation-only datasets.
The mixed-interventional comparison is clearer: \method{} is best in both F1 and SHD; NoDAGS and AVICI have nearby SHD with lower F1.

%% file: figs/tables/main/table_main_benchmark.tex
\begin{table}[!htbp]
\centering
\scriptsize
\setlength{\tabcolsep}{1.8pt}
\renewcommand{\arraystretch}{1.05}
\begin{threeparttable}
\caption{\textbf{Synthetic data benchmark.} Overall performance macro-averaged over all (dataset family $\times$ $d$) settings with equal weight. Obs (observation-only) uses 1000 observational samples; Mixed (mixed-interventional) uses 800 observational + 200 interventional samples. Rank is the average rank across all configurations (lower is better). Values are mean (standard deviation).}
\label{tab:main-benchmark}
\begin{tabular*}{\textwidth}{@{\extracolsep{\fill}}lcccc@{\hspace{4pt}}cccc}
\toprule
& \multicolumn{4}{c}{\textbf{Observation-only}} & \multicolumn{4}{c}{\textbf{Mixed-interventional}} \\
\cmidrule(lr){2-5}\cmidrule(lr){6-9}
\textbf{Method} & \textbf{Rank}$\downarrow$ & \textbf{F1}$\uparrow$ & \textbf{SHD}$\downarrow$ & \textbf{SID}$\downarrow$ & \textbf{Rank}$\downarrow$ & \textbf{F1}$\uparrow$ & \textbf{SHD}$\downarrow$ & \textbf{SID}$\downarrow$ \\
\midrule
RandomRegress & 14.3 & 0.240\scorestd{0.112} & 63.3\scorestd{47.9} & 76.1\scorestd{54.0} & \NA & \NA & \NA & \NA \\
DAS & 4.4 & 0.699\scorestd{0.185} & 10.7\scorestd{11.0} & 20.6\scorestd{23.5} & \NA & \NA & \NA & \NA \\
LiNGAM & 11.1 & 0.420\scorestd{0.237} & 15.7\scorestd{11.8} & 30.4\scorestd{25.4} & \NA & \NA & \NA & \NA \\
PC & 8.6 & 0.496\scorestd{0.163} & 14.1\scorestd{10.0} & 39.6\scorestd{30.3} & \NA & \NA & \NA & \NA \\
CDIS & 10.2 & 0.429\scorestd{0.175} & 15.0\scorestd{10.2} & 41.6\scorestd{31.9} & 6.8 & 0.500\scorestd{0.177} & 14.0\scorestd{10.3} & 38.7\scorestd{31.0} \\
GIES & 6.2 & 0.599\scorestd{0.187} & 11.6\scorestd{12.0} & 28.2\scorestd{28.7} & 3.0 & 0.780\scorestd{0.183} & 8.1\scorestd{10.8} & 16.8\scorestd{23.8} \\
IGSP & 7.2 & 0.554\scorestd{0.184} & 15.5\scorestd{13.4} & 41.8\scorestd{35.2} & 6.6 & 0.540\scorestd{0.192} & 13.6\scorestd{12.0} & 37.0\scorestd{33.1} \\
DAGMA & \underline{4.1} & \underline{0.717}\scorestd{0.207} & \underline{9.2}\scorestd{9.4} & 19.3\scorestd{21.9} & \NA & \NA & \NA & \NA \\
NOTEARS & 7.8 & 0.560\scorestd{0.249} & 9.3\scorestd{8.0} & \underline{17.8}\scorestd{18.8} & \NA & \NA & \NA & \NA \\
NOTEARS-MLP & 7.3 & 0.585\scorestd{0.210} & 12.0\scorestd{9.9} & 24.7\scorestd{22.6} & \NA & \NA & \NA & \NA \\
NoDAGS & \NA & \NA & \NA & \NA & 6.1 & 0.573\scorestd{0.232} & 12.7\scorestd{10.8} & 27.0\scorestd{26.1} \\
SEA & 5.2 & 0.647\scorestd{0.181} & 10.8\scorestd{8.6} & 24.4\scorestd{22.3} & 5.0 & 0.611\scorestd{0.182} & 15.2\scorestd{12.1} & 27.8\scorestd{24.6} \\
SDCD & 10.1 & 0.445\scorestd{0.168} & 18.6\scorestd{14.0} & 49.3\scorestd{38.9} & 4.9 & 0.635\scorestd{0.171} & 14.2\scorestd{11.3} & 32.7\scorestd{29.2} \\
DCDI & 13.0 & 0.276\scorestd{0.080} & 79.9\scorestd{54.3} & 124.2\scorestd{86.5} & 9.0 & 0.274\scorestd{0.078} & 81.3\scorestd{54.9} & 124.6\scorestd{86.7} \\
AVICI & 7.6 & 0.551\scorestd{0.197} & 11.6\scorestd{9.1} & 24.8\scorestd{22.7} & \underline{2.5} & \underline{0.844}\scorestd{0.142} & \underline{6.3}\scorestd{6.4} & \underline{9.4}\scorestd{15.9} \\
\midrule
\method{} & \textbf{2.7} & \textbf{0.811}\scorestd{0.156} & \textbf{5.6}\scorestd{5.1} & \textbf{11.0}\scorestd{13.9} & \textbf{1.0} & \textbf{0.952}\scorestd{0.074} & \textbf{2.2}\scorestd{2.9} & \textbf{2.9}\scorestd{7.2} \\
\bottomrule
\end{tabular*}
\end{threeparttable}
\end{table}

%% file: analysis.tex
\section{Analysis}
\label{sec:analysis}

\subsection{Sample-size scaling}
\label{sec:exp_n_scaling}

\begin{figure}[H]
  \centering
  \vspace{-6pt}
  \setlength{\abovecaptionskip}{6pt}
  \setlength{\belowcaptionskip}{0pt}
  \includegraphics[width=0.78\linewidth]{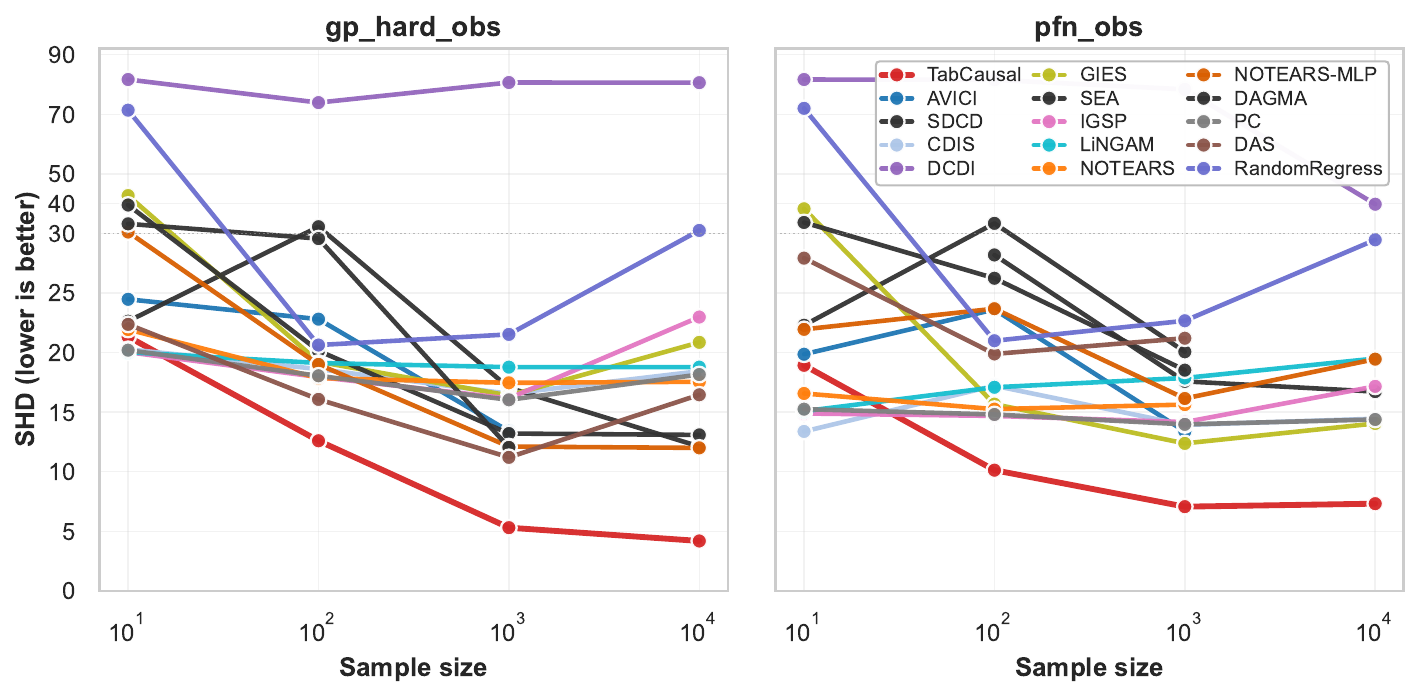}
  \caption{\textbf{Sample-size scaling on observational data.}
  SHD (lower is better) for \texttt{gp\_hard\_obs} and \texttt{pfn\_obs}, macro-averaged over $d\in\{5,10,20\}$ at each sample size $N\in\{10,100,1000,10000\}$.}
  \label{fig:n-scaling}
\end{figure}

Figure~\ref{fig:n-scaling} plots SHD versus $N\in\{10,100,1000,10000\}$ on observational \texttt{gp\_hard\_obs} and \texttt{pfn\_obs}; some methods are omitted at larger $N$ because they exceed the shared five-minute per-graph budget or fail to return a valid graph.
We use SHD here because it directly counts structural errors as sample size changes (lower is better).
As $N$ grows, \method{} uses the extra samples and is among the strongest methods at moderate and large $N$.
The drop in SHD is largest on \texttt{gp\_hard\_obs}, where more observations make nonlinear dependencies easier to read.
On \texttt{pfn\_obs} the field is tighter: several optimization-based baselines can use particular observational regularities, especially at $N=10000$.
With enough samples, \method{} stays strong in both families.
DCDI remains weak in this comparison because, under the shared five-minute budget of Section~\ref{sec:exp_setup}, its reported scores are early-stopped iterates.
\vspace{-6pt}
\subsection{Scalability to high dimensions}
\label{sec:exp_d_scaling}

\begin{figure}[!htbp]
\centering
\includegraphics[width=\linewidth]{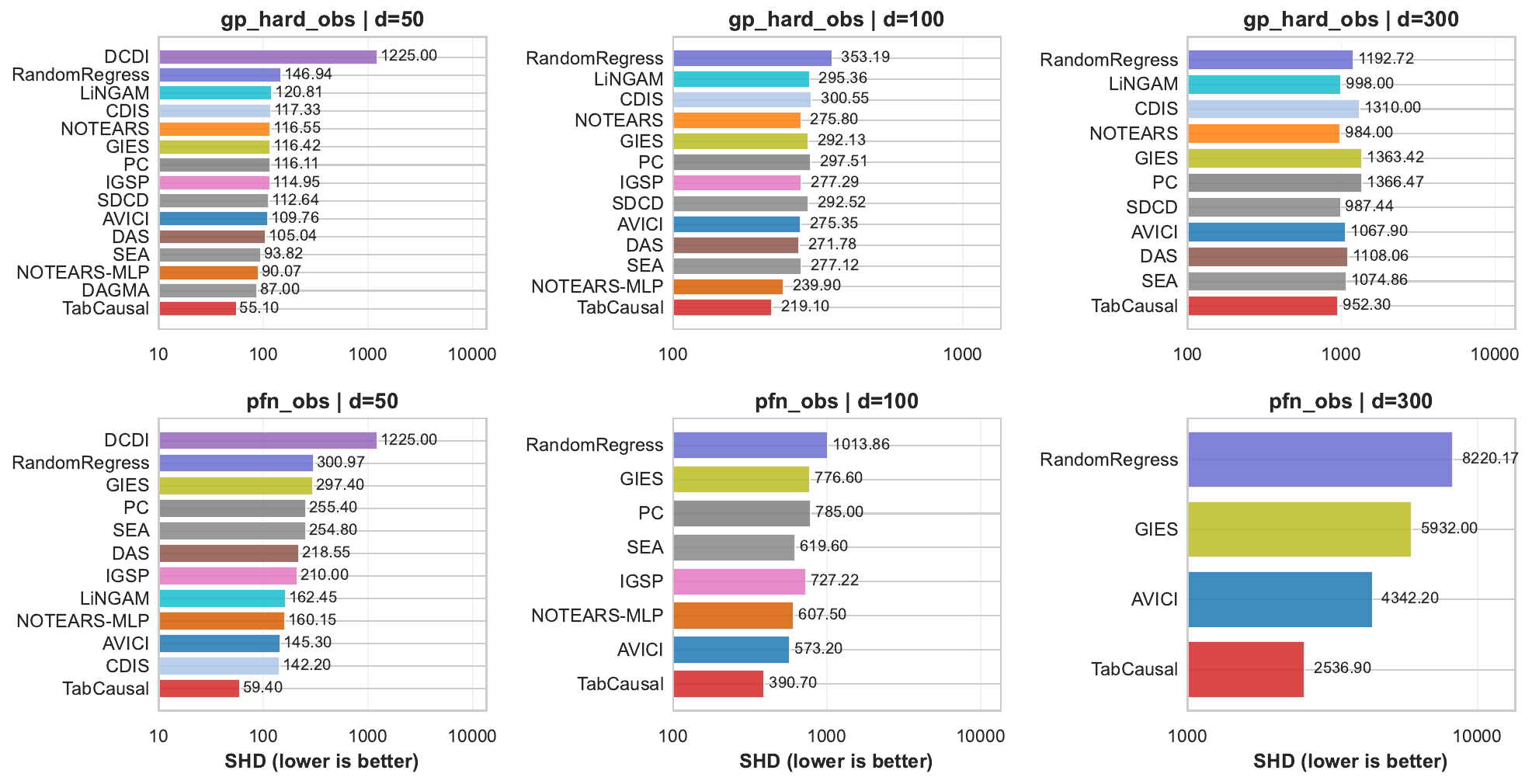}
\caption{%
\textbf{High-dimensional observational scalability.}
SHD (lower is better; log-scale axis) on \texttt{gp\_hard\_obs} and \texttt{pfn\_obs} for $d\in\{50,100,300\}$.
Only methods that completed under the five-minute per-graph timeout are shown.}
\label{fig:d-scaling}
\end{figure}

Figure~\ref{fig:d-scaling} reports SHD for \texttt{gp\_hard\_obs} and \texttt{pfn\_obs} at $d\in\{50,100,300\}$ among methods that complete under the shared five-minute per-graph budget; growing $d$ expands the directed-edge candidate set quadratically, so several baselines time out or return incomplete results.
Across displayed settings, \method{} remains competitive, while optimization-based baselines can be strong in some high-dimensional observational settings; RandomRegress remains an ordering/regression control with substantially worse SHD.

\FloatBarrier
\subsection{Embedding visualization on semantic causal environments}
\label{sec:analysis-embed}

\begin{figure}[!htbp]
  \centering
  \includegraphics[width=0.90\linewidth]{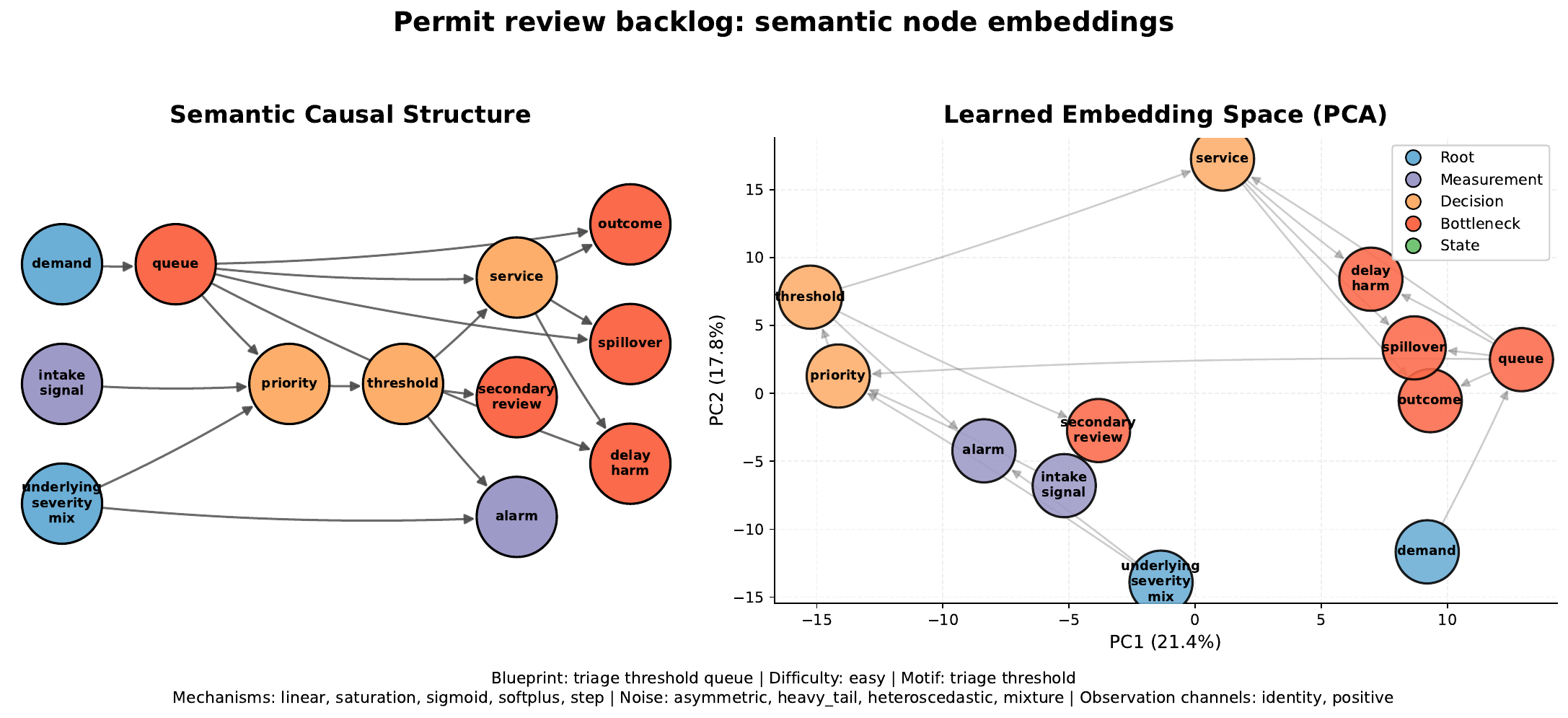}
  \caption{\textbf{Embedding visualization on Permit review backlog.}
  Left: true SCM. Right: PCA of last-layer node embeddings \(\widetilde{\mathbf{h}}_i\) from one observational table.}
  \label{fig:embeddings}
\end{figure}

We inspect whether last-layer node embeddings \(\widetilde{\mathbf{h}}_i\) group variables by their roles in a causal workflow.
The example is Permit review backlog, a government-services scenario from the semantic benchmark (Section~\ref{sec:semantic-benchmark}; \texttt{triage\_threshold\_queue} blueprint).
The graph is SCM-generated, so the directed edges are known, and the nodes have domain names such as queue load, priority score, and service start.
Figure~\ref{fig:embeddings} shows the true graph together with a 2D PCA of the embeddings computed from one observational table.

On the left, incoming demand load drives queue load.
Queue load, intake signal, and underlying severity mix inform the priority score; that score sets the triage threshold, which together with queue load affects when service starts.
Service start and queue load continue to a stabilized outcome, backlog spillover, and delay harm; the triage threshold feeds secondary review, and the triage threshold and underlying severity mix also feed an alarm record.

On the right, nearby points have similar learned representations.
Priority score and triage threshold sit together as decision variables.
Queue load, backlog spillover, stabilized outcome, and delay harm form a neighboring group of queue-related outcomes; service start is slightly apart, between the decision variables and those later outcomes.
Intake signal, alarm record, and underlying severity mix sit farther away, matching their earlier intake, measurement, and escalation roles in the graph.
The colored role labels (root, measurement, decision, bottleneck, state) follow this split: upstream and escalation variables separate from queue-centered outcomes, with decision and service variables in between.

Additional quantitative analyses of embeddings, graph-statistic prediction probes, and wall-clock runtime comparisons on the synthetic benchmark are provided in Appendix~\ref{sec:analysis-graphstats} and Appendix~\ref{sec:efficiency}.

%% file: conclusion.tex
\section{Conclusion}
\label{sec:conclusion}
We introduced \method{}, an amortized tabular causal discovery model that depends on broad causal pretraining: diverse graph priors, mechanisms, noise models, scales, and observational versus mixed-interventional regimes, composed into dynamic tasks so the learned cues still work when the generator changes.
On the macro-averaged synthetic and semantic benchmarks, a scored real-data check, held-out synthetic OOD tests (Appendix~\ref{app:heldout-ood}), and scaling analyses, \method{} recovers mixed-interventional structure well and keeps inference to a single forward pass, which is cheaper than optimizing a graph on each new dataset.

\paragraph{Limitations.}
Most of the evidence is still controlled and simulator-grounded. The public-table check is scored, but it is not a field deployment study.
Observational non-identifiability, Markov-equivalence ambiguity, and growing graph dimension (quadratic candidate edges) remain challenging, and the formulation does not address hidden confounding, cyclic feedback, temporal structure, strong selection effects, or calibrated uncertainty-aware decoding.

%% file: appendix.tex
\appendix
\section{Model Architecture and Training Details}
\label{app:training}

\subsection{Model Architecture}\label{sec:appendix-architecture}

\method{} employs a transformer-based encoder with a graph prediction head for tabular causal structure learning.

\paragraph{Feature encoder.}
The encoder processes input data through an axial attention mechanism that alternates between the observation axis and variable axis.
Our architecture consists of 8 transformer blocks, with each block applying attention along one axis before transposing to the other axis, yielding 16 effective layers.
Key architectural specifications:
\begin{itemize}[nosep,leftmargin=12pt]
    \item Hidden dimension: 128
    \item Multi-head attention: 8 heads with 32-dimensional keys/queries per head
    \item Feed-forward network: Expansion factor of 4 (hidden size 512)
    \item Regularization: 0.1 dropout rate
    \item Aggregation: Max pooling along the observation dimension
\end{itemize}

The axial design allows the model to capture both cross-variable dependencies and cross-sample distributional patterns efficiently.
Given input $\mathbf{X} \in \mathbb{R}^{B \times N \times d \times 2}$ (batch, observations, variables, channels), 
the encoder alternates attention operations along dimensions $N$ and $d$, and can run on graphs with hundreds of variables.

\paragraph{Graph prediction head.}
We score each directed pair $(i,j)$ with normalized similarity between asymmetric projections of node representations, then map scores to edge probabilities with a sigmoid (Eq.~\eqref{eq:edge_prob}).
Specifically, for node embeddings $\mathbf{h}_i$, we learn two projection functions $f_u(\cdot)$ and $f_v(\cdot)$ and compute:
\begin{equation}
\widehat{S}_{ij} = e^{\tau} \cdot \frac{f_u(\mathbf{h}_i)^\top f_v(\mathbf{h}_j)}{\|f_u(\mathbf{h}_i)\| \|f_v(\mathbf{h}_j)\|} + b,\qquad
\widehat{P}_{ij}=\sigma(\widehat{S}_{ij}),
\end{equation}
where $\tau$ is a learnable temperature (initialized at 2.0) and $b$ is a learnable bias (initialized at -1.0).
The L2 normalization bounds the similarity scores and improves training stability.

\subsection{Training Config}
\label{appendix:training_config}
\paragraph{Optimizer configuration.}
We employ the LAMB optimizer (Layer-wise Adaptive Moments optimizer for Batch training) with adaptive learning rate scheduling.
The base learning rate $\eta_{\text{base}} = 3 \times 10^{-5}$ is scaled according to the effective batch size:
\begin{equation}
\eta = \eta_{\text{base}} \cdot \sqrt{B_{\text{eff}}},
\end{equation}
where $B_{\text{eff}}$ denotes the maximum effective batch size observed during training.
This square-root scaling maintains optimization dynamics consistent across different batch sizes.
We use no weight decay, apply gradient clipping at magnitude 1.0, and maintain a constant base learning rate schedule for the duration of pretraining (scaled by effective batch size as above).

\paragraph{Mixed-precision training.}
Training is conducted entirely in BFloat16 arithmetic without loss scaling.
BFloat16 provides greater dynamic range than FP16, eliminating the need for gradient scaling and reducing numerical instability.
Training uses BFloat16 autocast in PyTorch.

\paragraph{Adaptive batch scheduling.}
To balance memory constraints and sample efficiency across varying graph dimensions, we use a dimension-dependent per-device micro-batch size.
For graph dimension $d$, the micro-batch size on each accelerator follows
\begin{equation}
B_d = \left\lfloor 0.1 + \frac{1000 \cdot (1 - e^{-0.1d})}{d} \right\rfloor.
\end{equation}
This damped schedule assigns larger micro-batches to small graphs (memory-efficient) and smaller micro-batches to large graphs (memory-intensive).
Each optimizer update accumulates gradients over $10$ consecutive micro-batches before stepping.

\paragraph{Dimension-balanced sampling.}
Dimensions are sampled with probability weights inversely related to their effective throughput under the micro-batch schedule above, accounting for the number of accelerator devices and the fixed gradient-accumulation depth (10 micro-batches per optimizer step).
Concretely, we use weights of the form
\begin{equation}
p(d) \propto \frac{d}{B_d \cdot N_{\text{GPU}}},
\end{equation}
which encourages comparable exposure across dimensions despite heterogeneous per-step costs.

\paragraph{Data augmentation via intervention mixing.}
Training alternates between observational-only and mixed observational--interventional episodes drawn from the same environment distribution.
With probability $0.5$, an episode is observational-only ($n_{\mathrm{int}}=0$); otherwise it is a mixed-regime episode.
In mixed episodes, the interventional sample count $n_{\mathrm{int}}$ is drawn approximately log-uniformly from an interval with lower endpoint $\max(d,1)$ and upper endpoint $200$, where $d$ is the graph dimension.
Conditional on $n_{\mathrm{int}}$, the observational sample count $n_{\mathrm{obs}}$ is drawn approximately log-uniformly from an interval with lower endpoint $\max(100,\,2n_{\mathrm{int}})$ and upper endpoint $1000$.
Sampled counts are then clipped by configured episode caps; for the final training setup these caps correspond to at most $600$ observational and $200$ interventional samples per episode.

\paragraph{Computational requirements.}
Pretraining was run on $2\times$ NVIDIA RTX~4090 GPUs using BFloat16 mixed precision as described above.
The full training run required on the order of 20 days of wall-clock time on this hardware.

\subsection{Pretraining Data Configuration}
\label{appendix:pretrain_data}

A critical part of \method{} is pretraining: the model learns dataset-to-graph inference on a wide mix of causal environments before it is used.
AVICI trains on a narrower simulator, mainly linear and smooth GP mechanisms.
We vary how much of the environment catalog is included, and measure transfer when that catalog grows.
The prior-strength ablation (Section~\ref{app:ablations}) measures transfer when that catalog grows; Appendix~\ref{app:heldout-ood} reports a fully held-out synthetic OOD check.

We consider two pretraining configurations that differ substantially in their coverage of causal mechanisms and graph structures:
the weak-prior variant focuses on well-studied mechanism families similar to prior work,
while the strong-prior variant significantly expands the distribution to include challenging nonlinear, non-additive, and heteroscedastic mechanisms.

\paragraph{Weak-prior pretraining.}
The weak-prior catalog covers:
\begin{itemize}[nosep,leftmargin=12pt]
    \item \textbf{Graph structures}: Erd\H{o}s-R\'enyi (edges per variable: 1.0, 2.0, 3.0), Scale-Free and its transpose (edges: 1--3, power: 0.7/1.0/1.2/1.5), Watts-Strogatz (dimension: 2/3, neighbors: 1/2, rewiring: 0.2/0.4), Stochastic Block Models (edges: 1--3, blocks: 2/5/10, damping: 0.1), and Geometric Random Graphs (radius: 0.08/0.10/0.15)
    \item \textbf{Mechanisms}: 
    \begin{itemize}[nosep,leftmargin=12pt]
        \item Linear additive with homoscedastic noise (weights: 0.25--2.0, 2.0--4.0, and 0.25--4.0; biases: $[-3,3]$)
        \item Linear additive with heteroscedastic noise (same weight and bias ranges, with RFF-based heteroscedastic scale modulation: length scale 10.0, output scale 2.0)
        \item RFF additive with RBF kernels (length scale: 5.0--8.0, 8.0--12.0, and 5.0--12.0; output scale: 8.0--15.0, 15.0--22.0, and 8.0--22.0)
        \item Heteroscedastic RFF additive with the same kernel ranges and RFF-based heteroscedastic scale modulation (length scale 10.0, output scale 2.0)
    \end{itemize}
    \item \textbf{Noise distributions}: Gaussian, Laplace, Cauchy
    \item \textbf{Interventions}: one intervened variable per interventional row, with signed uniform intervention magnitudes in $[1,5]$
    \item \textbf{Training dimensions}: $d \in \{2, 5, 10, 20, 30, 40, 60, 80, 100\}$
\end{itemize}

\paragraph{Strong-prior pretraining.}
To test whether broader pretraining distributions improve generalization, we substantially expand the simulator coverage:
\begin{itemize}[nosep,leftmargin=12pt]
    \item \textbf{Graph structures}: Erd\H{o}s-R\'enyi (edges per variable: 0.5, 1.0, 2.0, 4.0), Scale-Free and its transpose (edges: 1--2, power: 0.5/1.0/1.5), Stochastic Block Models (edges: 1--2, blocks: 2/5/10, damping: 0.05/0.2), and Geometric Random Graphs (radius: 0.08/0.12/0.16), together with additional structured graph families including staged workflow graphs (stages: 4/5/6, edge probability: 0.12/0.18/0.24, skip probability: 0.02/0.05/0.08, edge cap: 1.1/1.4 per variable), risk-root fork graphs (roots: 1/2/3, branch probability: 0.40/0.55/0.70, local-chain probability: 0.30/0.45/0.60, overlap probability: 0.03/0.08/0.12, edge cap: 1.2/1.6), and an additional sparse scale-free-transpose prior (edges: 1--2, power: 0.8--1.2)
    \item \textbf{Mechanisms}: 
    \begin{itemize}[nosep,leftmargin=12pt]
        \item Linear additive with stratified signal strengths (weights: 0.05--0.5, 0.5--2.0, and 2.0--5.0; biases: $[-2,2]$; noise scales: 0.05--0.2, 0.2--1.0, and 1.0--2.0)
        \item Heteroscedastic linear additive (weights: 0.25--3.0; biases: $[-3,3]$; RFF-based heteroscedastic scale modulation with length scale 10.0 and output scale 2.0)
        \item RFF additive covering both smooth regimes (length scale: 7.0--12.0, output scale: 8.0--20.0) and high-frequency regimes (length scale: 0.5--1.5 and 1.5--3.0, output scale: 0.5--2.0)
        \item MLP additive mechanisms in a simple regime (1 hidden layer, hidden dimension: 16/32, activations: Tanh/LeakyReLU, noise scale: 0.1--0.5) and a harder regime (2--3 hidden layers, hidden dimension: 32/64, activation: ReLU, noise scale: 0.2--1.0)
        \item Multiplicative linear mechanisms (weights: 0.5--2.0, biases: $[-1,1]$) with multiplicative noise factors in $[0.8,1.2]$ or $[0.5,1.5]$
        \item Polynomial additive mechanisms (degree 2 with interactions, weights: 0.1--0.8, biases: $[-0.5,0.5]$, noise scale: 0.1--0.8)
        \item Post-nonlinear models with a linear base (weights: 0.5--2.0, biases: $[-1,1]$, Gaussian noise scale: 0.2--0.8) followed by Tanh or Sigmoid distortions
        \item Additional neural structural mechanisms with randomized meta-parameters: number of layers sampled from a Gamma-family meta-prior with lower bound 2, hidden dimension sampled from a Gamma-family meta-prior with lower bound 4 and scale up to 100, dropout drawn from a scaled Beta-family meta-prior, structural noise scale drawn from a log-scaled truncated-normal meta-prior with means ranging from $10^{-4}$ to 0.3, initialization scale drawn from a log-scaled truncated-normal meta-prior with means ranging from $10^{-2}$ to 10.0, number of root causes sampled from a Gamma-family meta-prior with lower bound 2, and activations chosen from \{Tanh, Identity, ReLU\}; additional binary variations include pre-sampled weights, target-as-effect mode, block-wise dropout, feature sorting, clique-style structure, and random feature rotation
        \item Additional structured monotone and heavy-tail mechanisms including heavy-tail linear additive (weights: 0.20--2.50, biases: $[-1,1]$, Student-$t$ noise with $\mathrm{df}=4$ and Laplace noise, noise scale: 0.10--0.60 and 0.60--1.20), softplus-additive monotone mechanisms (weights: 0.25--2.00, biases: $[-0.8,0.8]$, noise scale: 0.08--0.50), sigmoid-additive monotone mechanisms (weights: 0.25--1.80, biases: $[-0.8,0.8]$, noise scale: 0.05--0.35), and threshold-additive mechanisms (weights: 0.25--1.50, biases: $[-0.4,0.4]$, thresholds: 0.10--0.45, noise scale: 0.05--0.30 and 0.30--0.70)
    \end{itemize}
    \item \textbf{Noise distributions}: Gaussian, Laplace, Uniform, Exponential, Gumbel, Student-$t$ ($\mathrm{df}=4$), together with multiplicative uniform noise factors for non-additive mechanisms
    \item \textbf{Interventions}: one intervened variable per interventional row, with signed uniform intervention magnitudes typically in $[1,5]$ for the broad base families, $[1,4]$ for heavy-tail linear mechanisms, $[1,3.5]$ for softplus and threshold mechanisms, and $[1,3]$ for sigmoid, high-frequency RFF, multiplicative, polynomial, and simple/hard MLP mechanisms
    \item \textbf{Training dimensions}: $d \in \{2, 5, 10, 20, 30, 40, 50, 60, 80, 100\}$
\end{itemize}

The strong-prior configuration expands mechanism diversity and task mix relative to the weak prior.
The ablation in Table~\ref{tab:ablation} shows substantial gains from that richer catalog.

\section{Benchmark Dataset Specifications}
\label{app:benchmark}

We evaluate \method{} on seven synthetic benchmark families spanning diverse causal mechanisms and graph structures.
Unless otherwise specified by the dataset family, graphs are Erd\H{o}s-R\'enyi (ER) random graphs.
Sparsity is controlled by the ER parameter \texttt{edges\_per\_var} (expected edges per variable), which is sampled per graph instance from a dimension-dependent discrete set.
In the paper evaluation suite, we use the following triples:
\begin{itemize}[nosep,leftmargin=12pt]
    \item $d=5$: $\{0.8, 1.0, 1.2\}$
    \item $d=10$: $\{1.2, 1.5, 1.8\}$
    \item $d=20$: $\{1.6, 2.0, 2.4\}$
    \item $d=50$: $\{2.0, 2.5, 3.0\}$
    \item $d=100$: $\{2.4, 3.0, 3.6\}$
    \item $d=300$: $\{3.0, 3.6, 4.2\}$
\end{itemize}
Graphs remain sparse across these settings, with \texttt{edges\_per\_var} increasing only modestly as $d$ grows.

\subsection{Intervention Method}

Each benchmark family is instantiated under observational-only and mixed-interventional regimes as described in the main experiments.
Observational regimes contain no interventional rows ($N_{\text{int}}=0$).
When interventional data are present (mixed regime), we apply structural single-variable interventions that propagate to descendants under the structural causal model.
Interventions are allocated so that each variable is intervened upon at least once, counts are split as evenly as possible across variables ($\lfloor N_{\text{int}}/d\rfloor$ each with remainder spread across variables), and mixed regimes typically use one intervened variable per interventional sample row.
Intervention strengths are drawn from signed uniform ranges (typically $[1.0, 3.0]$ or $[1.0, 5.0]$), representing moderate atomic perturbations.

\subsection{Dataset Specifications}

\paragraph{Linear mechanisms.}
Three datasets use linear additive noise models $X_j = \sum_{i \in \text{PA}(j)} w_{ij} X_i + b_j + \varepsilon_j$:

\begin{itemize}[nosep,leftmargin=12pt]
    \item \textbf{linear\_gauss}: Edge weights $w_{ij} \sim \text{Uniform}(0.5, 2.0)$, bias $b_j \sim \text{Uniform}(-2, 2)$, Gaussian noise $\varepsilon_j$ with scale $\sim \text{Uniform}(0.5, 1.5)$. Erdős-Rényi graphs.
    
    \item \textbf{linear\_nongauss}: Same weight and bias distributions as above, but noise follows either $\text{Uniform}(-1.5, 1.5)$ or $\text{Exponential}(1.0)$. Erdős-Rényi graphs.
    
    \item \textbf{linear\_graph}: Identical mechanism to \texttt{linear\_gauss}, but tests alternative graph topologies: Scale-Free graphs (power = 1.0, average degree = 4.0) and Stochastic Block Models (2/5/10 blocks, edge density 1-3 per variable, damping = 0.1). This tests generalization to hub-and-spoke and modular structures.
\end{itemize}

\paragraph{Gaussian Process mechanisms.}
Two datasets use nonlinear functions via Random Fourier Features approximating GP priors with RBF kernels.
The key distinction lies in the length scale parameter $\ell$:

\begin{itemize}[nosep,leftmargin=12pt]
    \item \textbf{gp\_simple}: Length scale $\ell \sim \text{Uniform}(7, 10)$, output scale $c \sim \text{Uniform}(10, 20)$. Large length scales produce smooth, slowly-varying functions. Bias $b \sim \text{Uniform}(-2, 2)$, Gaussian noise with scale $\sim \text{Uniform}(0.5, 1.5)$.
    
    \item \textbf{gp\_hard}: Length scale $\ell \sim \text{Uniform}(0.5, 1.5)$, output scale $c \sim \text{Uniform}(0.5, 2.0)$. Short length scales induce high-frequency oscillations and sharp transitions, significantly increasing approximation difficulty. Bias $b \sim \text{Uniform}(-1, 1)$, Gaussian noise with scale $\sim \text{Uniform}(0.2, 1.0)$.
\end{itemize}

Both GP datasets use Erdős-Rényi graphs.

\paragraph{Multiplicative noise.}
\texttt{mul\_noise} tests robustness to heteroscedasticity via multiplicative noise models:
$X_j = \left(\sum_{i \in \text{PA}(j)} w_{ij} X_i + b_j\right) \cdot \varepsilon_j$,
where weights $w_{ij} \sim \text{Uniform}(0.5, 2.0)$, bias $b_j \sim \text{Uniform}(-1, 1)$, and noise multiplier $\varepsilon_j \sim \text{Uniform}(0.5, 1.5)$.
This violates the additive noise assumption. Erdős-Rényi graphs.

\paragraph{PFN-style mechanisms.}
\texttt{pfn} is a synthetic causal family generated with a PFN-style prior over structural equations.
It uses heterogeneous nonlinear mechanisms, including MLP-based structural equations, and is evaluated under the same observational-only and mixed-interventional regimes as the other families.
This family provides an additional nonlinear benchmark with a generator distinct from the AVICI-style mechanism families above.

\section{Semantic Causal Environment Benchmark}
\label{app:semantic-benchmark}

\begin{figure}[ht]
\centering
\includegraphics[width=\textwidth]{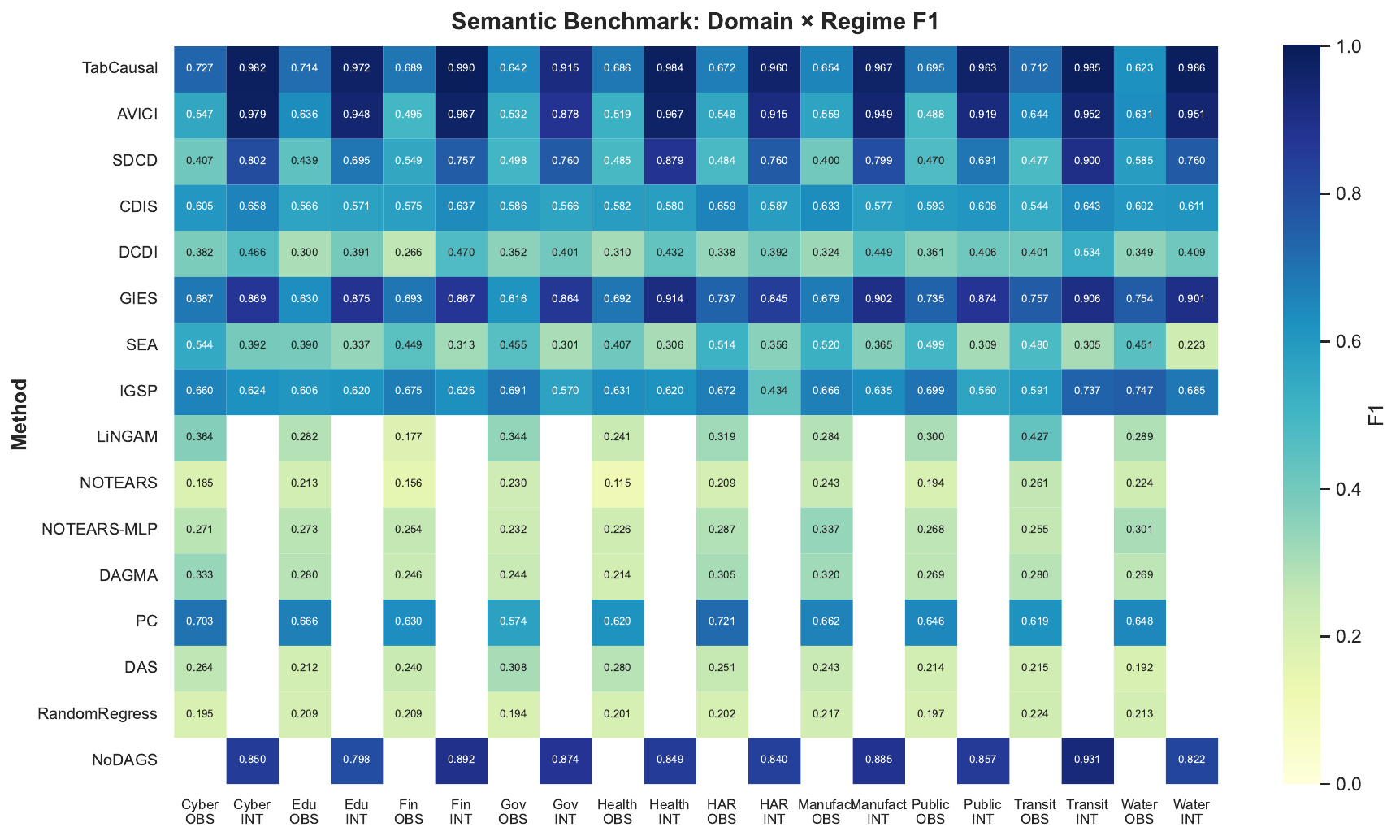}
\caption{\textbf{Semantic benchmark domain/regime heatmap.} F1 varies across domains and observational vs.\ mixed-interventional regimes. Blank cells indicate regimes not applicable or not run for the corresponding method. \method{} tends to gain most from mixed-interventional evidence, whereas observational panels remain more competitive across baselines.}
\label{fig:semantic-domain-heatmap}
\end{figure}

The semantic causal environment benchmark is synthetic and simulator-grounded.
Human blueprints, mechanism lists, graph and intervention constraints, and validation rules set what may be authored.
An LLM then fills scenario specifications, which are checked and compiled into SCMs with named variables and directed mechanisms, and into observational and interventional tables with known adjacencies.
Figure~\ref{fig:semantic-domain-heatmap} and Table~\ref{tab:semantic-slice-breakdown} summarize empirical scores by domain/regime and by recurring failure-motif tags; the main text reports regime-split aggregates.

\begin{table}[ht]
\centering
\small
\caption{Semantic causal environment benchmark inventory summary. Synthetic, simulator-grounded SCMs with known adjacencies; detailed schema, sampling, generation, validation, and audit specifications appear in the remainder of this appendix.}
\label{tab:semantic-inventory}
\begin{tabularx}{\linewidth}{@{}>{\raggedright\arraybackslash}p{0.22\linewidth}X@{}}
\toprule
\textbf{Aspect} & \textbf{Summary} \\
\midrule
Suite character & Semantically grounded synthetic SCMs compiled from validated scenario specifications (domain blueprints plus LLM-assisted instantiation); simulator-known DAGs. \\
Scale & 10 thematic domains; 100 scenarios ($10$ per domain); 200 dataset--regime evaluation tasks (\texttt{obs} and \texttt{int} per scenario). \\
Graph size & $|\mathcal{V}|\in[8,17]$. \\
Regimes & \texttt{obs}: observation-only tensors; \texttt{int}: mixed observational and structural-intervention rows with coordinate masks (aligned with Section~\ref{sec:method_arch}). \\
Sampling scale & Sample sizes drawn uniformly from $\{100,\ldots,2000\}$; mixed regimes randomize an intervention fraction in $[0.10,\,0.30]$ (detailed below). \\
Interventions (mixed) & Balanced variable-level coverage: each variable intervened at least once with near-equal counts. \\
\bottomrule
\end{tabularx}
\end{table}

\subsection{Package scope and domains}
\begin{itemize}[nosep,leftmargin=12pt]
    \item \textbf{Inventory:} $10$ thematic domains $\times$ $10$ scenarios each $=$ 100 scenario graphs; each scenario yields two datasets: an observation-only regime and a mixed observational--interventional regime (structural interventions), for 200 dataset--regime evaluation instances in total.
    \item \textbf{Domains:} healthcare delivery; finance and credit; education; manufacturing; public health; urban transportation; water and sanitation; cybersecurity and IT operations; housing and real estate; government services and policy implementation.
\end{itemize}
Each scenario specification attaches blueprint metadata, per-variable roles, per-edge mechanism assignments and rationales, difficulty tags, challenge motifs, and semantic intervention handles for interpretation.
Specifications are generated under human-defined blueprint and validation constraints.

\subsection{Graph scale and difficulty}
\begin{itemize}[nosep,leftmargin=12pt]
    \item \textbf{Graph size:} every graph has $|\mathcal{V}| \in [8,17]$ nodes; most scenarios are small-to-medium graphs.
    \item \textbf{Per-domain difficulty mix:} within each domain, scenarios follow a fixed composition of 2 easy, 5 medium, and 3 hard graphs.
\end{itemize}
\textbf{Semantic rationale.}
Most graphs are small to medium because the intended difficulty is semantic: noisy or delayed reports, screening and detection pathways, queues and thresholds, and how interventions should be read.
That size still includes nonlinear mechanisms and mixed observation families, and it keeps specification, validation, and repeated motifs manageable.

\subsection{Regime sampling and intervention construction}
\paragraph{Sample sizes.}
Let $d=|\mathcal{V}|$ denote the number of variables in the scenario graph.
\begin{itemize}[nosep,leftmargin=12pt]
    \item \textbf{Observation-only regime:} draw $n_{\mathrm{obs}} \sim \mathrm{Unif}\{100,\ldots,2000\}$ and set $n_{\mathrm{int}}=0$.
    \item \textbf{Mixed regime:} draw total sample size $n \sim \mathrm{Unif}\{100,\ldots,2000\}$ and an intervention ratio $\rho \sim \mathrm{Unif}[0.10,0.30]$.
    Set $n_{\mathrm{int}}=\max\!\big(d,\,\mathrm{round}(n\rho)\big)$ and $n_{\mathrm{obs}}=n-n_{\mathrm{int}}$.
    If this allocation would violate feasibility (e.g., interventional demand exceeds the sampled total), the generator increases $n$ so that the mixed dataset still contains at least one observational row while respecting the intended intervention coverage.
\end{itemize}

\paragraph{Intervention scheduling and values.}
Interventions are structural: perturbations propagate to descendants in the SCM.
When $n_{\mathrm{int}}>0$, generation uses a balanced full-variable intervention schedule:
every variable is an eligible target; each variable is intervened at least once; per-variable intervention counts differ by at most one; and most interventional rows intervene on a single variable.

Intervention values are sampled as signed structural shifts on the standardized latent scale, typically lying roughly within $[-2,2]$, with magnitudes chosen to make interventions informative while avoiding a deterministic readout of the graph.

\subsection{Structural mechanisms}
Each directed edge is assigned a named mechanism that maps a weighted combination of parent variables to the child.
The name (for example, linear or sigmoid) fixes the shape of that mapping.
Edge weights, node biases, structural noise, and observation noise still vary across edges and scenarios, so two edges with the same name need not match numerically.

\paragraph{Primary mechanisms.}
\begin{itemize}[nosep,leftmargin=12pt]
    \item \textbf{linear:} identity response on weighted parent aggregation.
    \item \textbf{softplus:} monotone softplus-type activation, $\approx \mathrm{log1p}(\exp(\mathrm{clip}(x,-8,8)))-\log 2$.
    \item \textbf{sigmoid:} bounded monotone response, $\approx 2/(1+e^{-x})-1$.
    \item \textbf{threshold:} positive-part thresholding, $\approx \max(x-0.25,\,0)$.
    \item \textbf{saturation:} saturating response, $\approx \tanh(x/1.5)$.
    \item \textbf{step:} binary thresholding around $0$.
    \item \textbf{reciprocal\_like:} smooth sign-preserving compression, $\approx \tanh(x)$.
    \item \textbf{u\_shape:} U-shaped response, $\approx x^2-1$.
    \item \textbf{heavy\_tail\_driver:} heavy-tail-sensitive mapping, $\approx \mathrm{sign}(x)\sqrt{|x|+10^{-6}}$.
\end{itemize}

\paragraph{Rare optional mechanisms.}
\texttt{piecewise}, \texttt{interaction}, and \texttt{accumulation} appear only as negligible exceptions in the current package.

\paragraph{Package proportions.}
Across the packaged benchmark, mechanism assignments follow these aggregate proportions:
\begin{itemize}[nosep,leftmargin=12pt]
    \item \textbf{linear:} about $50\%$--$60\%$ of edges.
    \item \textbf{softplus + saturation:} together about $30\%$--$40\%$.
    \item \textbf{sigmoid:} about $5\%$--$10\%$.
    \item \textbf{threshold + step:} together about $5\%$--$10\%$.
    \item \textbf{reciprocal\_like + u\_shape + heavy\_tail\_driver:} together at most $\approx 5\%$.
    \item \textbf{optional rare mechanisms:} together at most $\approx 5\%$.
\end{itemize}

\textbf{Semantic rationale.}
Linear mechanisms dominate because many domain narratives admit approximately local, monotone causal influence over the modeled operating range.
Softplus and saturation capture activation and ceiling effects common in queues, throughput limits, adherence, and screening uptake.
Sigmoid-like shapes suit bounded middle-range outcomes (rates, probabilities, completion fractions).
Threshold and step shapes are used sparingly for gating, escalation, screening cutoffs, and policy triggers.
Reciprocal-like, U-shaped, and heavy-tail-driver mechanisms are intentionally rare, and they encode specific semantic motifs.

\subsection{Latent structural noise}
Each variable picks a noise type from the list below.
A scenario does not need to use every type.
The numbers in parentheses are defaults; a compiled SCM may use other values from its own specification.
\begin{itemize}[nosep,leftmargin=12pt]
    \item \textbf{gaussian:} zero-mean Gaussian structural noise.
    \item \textbf{heteroscedastic:} Gaussian noise multiplied by a random scale factor (e.g., $\approx 0.7 + U(0,1)$).
    \item \textbf{heavy\_tail:} Student-$t$--style heavy-tail noise (e.g., $\mathrm{df}=3$, scale $\approx 0.6$).
    \item \textbf{asymmetric:} Gaussian component plus a one-sided exponential component.
    \item \textbf{mixture:} Gaussian baseline with an additional ``burst'' component (e.g., activation probability $\approx 0.14$ at $\approx 2\times$ baseline scale).
\end{itemize}
\textbf{Typical scales.}
Noise is added in the structural equation and kept moderate relative to the standardized variables.
We also limit how often each type appears in the full benchmark, so one family does not take over.

\textbf{Semantic rationale.}
Gaussian noise is ordinary variation.
Heteroscedastic noise grows or shrinks with a random scale.
Asymmetric noise is one-sided, as with delays or shocks.
Mixture noise adds occasional bursts.
Heavy tails allow rare large deviations.

\subsection{Observation channels}
After the SCM produces a latent value, that value is written into a table column through an observation channel.
Each variable chooses one channel and an observation-noise scale.
The families below are options; a single graph does not use all of them.

\paragraph{Main observation families.}
\texttt{identity}, \texttt{bounded}, \texttt{positive}, \texttt{ordinal}, \texttt{contaminated\_identity}, \texttt{missing\_to\_zero\_identity}, \texttt{missing\_to\_zero\_positive}, \texttt{zero\_inflated\_positive}, \texttt{censored\_positive}.

\paragraph{Rare optional observation families.}
\texttt{count\_like}, \texttt{rate\_like}, \texttt{proxy\_noisy}, \texttt{missingness\_corrupted} (almost unused).

\paragraph{Package proportions.}
The percentages below describe the packaged benchmark as a whole:
\begin{itemize}[nosep,leftmargin=12pt]
    \item \textbf{bounded:} about $45\%$--$55\%$ of variable observations.
    \item \textbf{positive:} about $30\%$--$40\%$.
    \item \textbf{identity:} about $5\%$--$15\%$.
    \item \textbf{ordinal:} at most $\approx 5\%$.
    \item \textbf{all harder families together:} about $2\%$--$10\%$.
    \item Most graphs mainly use identity, bounded, positive, or ordinal. A few graphs include one harder channel; only a small number mix several harder channels.
\end{itemize}

\paragraph{Rendering.}
Corruption rates and noise levels are set per variable in the scenario file; the numbers below are defaults.
\begin{itemize}[nosep,leftmargin=12pt]
    \item \textbf{identity:} additive Gaussian observation noise on centered latent channels.
    \item \textbf{bounded:} logistic-style squashing toward $[-1,1]$.
    \item \textbf{positive:} softplus-style strictly positive rendering.
    \item \textbf{ordinal:} thresholded discretization with a small fixed cut set.
    \item \textbf{censored\_positive:} positive rendering with upper censoring.
    \item \textbf{zero\_inflated\_positive:} positive rendering with zero inflation (e.g., $\approx 10\%$).
    \item \textbf{contaminated\_identity:} identity rendering with contamination spikes (e.g., $\approx 5\%$).
    \item \textbf{missing\_to\_zero\_identity / missing\_to\_zero\_positive:} missingness mapped to zero (e.g., $\approx 10\%$).
\end{itemize}
After rendering, observation noise is usually about $0.04$--$0.16$ (median $\approx 0.08$).

\textbf{Semantic rationale.}
Bounded channels fit scores, rates, and fractions.
Positive channels fit counts, queues, delays, and costs.
Identity fits centered process variables.
Ordinal fits coarse grades.
Harder channels sit mainly on measurement, alarm, screening, and follow-up nodes: missing values written as zeros, rare spikes, and top-coding.
Each scenario picks from this list as needed.

\subsection{Edge weights}
Edges are assigned per scenario under the blueprints, with roles such as main process versus reporting.
We aim for absolute weights in about $[0.20,\,1.00]$, with operational drivers stronger than reporting or proxy edges.
In the packaged set, absolute weights range about $[0.30,\,1.18]$, with median near $0.72$.
This leaves some edges weak and others strong, which matches the story of each scenario.

\subsection{Motifs and slice tags}
\paragraph{Blueprint motifs.}
The same story patterns appear across domains: triage--threshold--queue; forecast--buffer--imbalance; calibration--drift--alarm; screening--detection--followup; contamination--spread--control; engagement--fatigue--mastery; pricing--friction--conversion; throughput--reciprocal--process.

\paragraph{Slice tags.}
For scoring we also tag graphs with labels such as \texttt{triage\_threshold}, \texttt{measurement\_failure}, \texttt{detection\_failure}, \texttt{followup\_breakdown}, \texttt{alarm\_drift}, \texttt{proxy\_interpretation}, \texttt{queue\_pressure}, \texttt{reciprocal\_process}, \texttt{heavy\_tail\_instability}, and \texttt{mixed\_regime\_ambiguity}.
Table~\ref{tab:semantic-slice-breakdown} shortens these names (e.g., \texttt{triage\_thr.}, \texttt{meas\_fail.}, \texttt{mix\_amb.}).
A graph can carry several tags, and the same tag can appear in more than one domain.
The tags name the difficulty itself, such as wrong measurement, missed detection, or an overloaded queue.

\subsection{Results by failure motif}
Table~\ref{tab:semantic-slice-breakdown} reports mean (standard deviation) F1 by slice tag.

\input{figs/tables/semantic/table_semantic_slices}

A graph can belong to more than one slice, so the rows are overlapping groups.
On mixed-interventional slices, \method{} keeps a high F1 across tags, and the spread across tags is small.
Observation-only slices are closer among methods, because directions are harder to recover; GIES is often strong there.
The mixed-interventional slices show a clearer gap for \method{}.

\subsection{LLM generation and audit}
\label{app:semantic-llm-audit}

We use an LLM to fill scene specifications from the human blueprints, mechanism lists, graph and intervention rules, and validation checks; the resulting JSON is then compiled into SCMs.
With 100 scenes, free-text fields can drift (repeated phrasing, uneven difficulty notes).
We therefore run a separate review with Claude that does not look at \method{} scores or any other method ranking.

The review checks (i) the packaging protocol, (ii) whether the generator matches the documented sampling and intervention rules, (iii) whether the overall mix matches the summary tables, and (iv) whether sampled JSON files still match their blueprints, across domains and difficulty levels.
It reads the protocol, generator logic, packaged metadata, summary statistics, and a stratified sample of scene files.

On the inspected files, the generator mostly follows the protocol, and the graphs still match the blueprints.
We did find small metadata issues, such as repetitive free-text or uneven extra labels in the narrative fields.

\subsection{Two example scenarios}
Each packaged scene is one SCM: named columns, a directed graph the generator uses, and tables of numbers sampled from that graph.
Mixed-regime tables use the balanced variable-level intervention schedule above; the stories name the operator-facing handles.

\paragraph{Healthcare delivery (screening--detection--follow-up).}
The scene is a diabetes clinic. Each row is one week.
How many patients ought to be screened, together with how hard it is to get in (travel, coverage, wait), sets how many actually arrive; the lab reading is taken only on those who showed up.
A high reading marks the patient for further tests, and a mark should lead the clinic to order those tests.
How many follow-up appointments are still open then decides whether that order becomes a finished visit, and how long marked patients wait.
If the calendar is full, marked patients wait and finished visits stay low even when the mark was right.
A week with few finished visits can mean two different things: high-risk patients were never marked, or they were marked and there were no appointments.
Those two weeks look the same if one only looks at finished visits.
Weeks where the clinic adds appointments, or forces the test order, raise finished visits when marks were already there.
If the marks were already missing, extra appointments do little.
Those two changes are what a coordinator would actually make.

\paragraph{Urban transportation (triage--threshold--queue).}
The scene is one intersection. Each row is one cycle of the traffic lights.
Cars arrive: how many want to get through, how many the roadside sensor counts on this side, and how many are actually waiting.
A score says which side should get green first (a main road scores higher than a side street); how long the queue is, or how high that score is, decides whether this side actually gets green.
Once green is given, cars still stuck when green ends, cars pushed into the next cycle, and extra waiting time all go up if the real queue was already long or green was already hard to get.
Buses and trucks in the mix make the same queue move more slowly.
The written shift log on the operator screen, and a later hand count of the same cycle, are two further looks; both can disagree with the sensor.
A cycle with long extra waiting time can mean the sensor under-counted cars, the score never let this side get green, or cars had already piled into the next cycle.
Forcing green, or raising this side's score, cuts waiting time when cars were actually there.
If the sensor was wrong and this side was empty, forcing green does not move waiting time.

\section{Additional Experimental Results}
\label{app:additional-results}

This appendix collects extra experimental details and results.

\subsection{Prior strength ablation}
\label{app:ablations}

We conduct an ablation study isolating pretraining prior strength.

\paragraph{Prior strength.}
We compare two variants that differ only in causal environment coverage during pretraining.
Both variants use the same backbone, trainable parameter count, inference/evaluation protocol, optimizer, adaptive batch schedule, and training budget; only the simulator support is changed.
The weak-prior variant focuses on linear additive and GP/RFF additive mechanisms (including heteroscedastic noise)
over multiple graph distributions (e.g., ER, SF, SBM, GRG).
The strong-prior variant substantially expands both (i) graph-structure coverage
(wider sparsity/density and broader structural parameters) and (ii) mechanism/noise diversity
(e.g., MLP mechanisms, multiplicative noise, polynomial interactions, post-nonlinear models, and richer non-Gaussian/asymmetric noises).

\paragraph{Results.}
Table~\ref{tab:ablation} summarizes the findings.
The strong-prior full model substantially improves over the weak-prior variant across SHD, SID, and F1.
Wider environment coverage in pretraining therefore shows up as better transfer on the evaluation mix.

\begin{table}[ht]
\centering
\small
\setlength{\tabcolsep}{4pt}
\renewcommand{\arraystretch}{1.12}
\caption{\textbf{Prior-strength ablation.} Effects of weak vs.\ strong pretraining priors under the same evaluation protocol.}
\label{tab:ablation}
\begin{tabular}{lccc}
\toprule
\textbf{Variant} & \textbf{SHD}$\downarrow$ & \textbf{SID}$\downarrow$ & \textbf{F1}$\uparrow$ \\
\midrule
Weak prior                      &  26.0 &  40.4 & 0.659 \\
Strong prior (full)             &  3.9 &  7.0 & 0.882 \\
\bottomrule
\end{tabular}
\end{table}

\subsection{Held-out synthetic OOD and intervention-protocol checks}
\label{app:heldout-ood}

The seven-family synthetic benchmark varies graph topology, mechanism family, noise, sample size, dimension, and observational versus mixed-interventional regimes under one protocol.
It measures transfer across those axes. Some families still overlap the pretraining catalog.

We add a held-out synthetic OOD benchmark.
Its graph families, mechanism families, root distributions, and noise families are absent from the \method{} pretraining prior.
The final three-family subset uses small-world, dense-local-module, and block-sparse-layered graphs; rational-ratio, trigonometric-periodic, tree-based, discretization-assignment, and mixture-regime mechanisms; Poisson, beta-scaled, gamma, lognormal, and Bernoulli roots; and beta, quantized, censored, mixture-outlier, and multiplicative-lognormal noise families.
In addition, pretraining uses one graph-level SCM configuration per graph, whereas this benchmark assigns mechanism, root, and noise families across nodes in a balanced way, so one graph can contain heterogeneous local mechanisms.

We follow the paper's synthetic protocol: 1000 observational samples for the observation-only regime, 800 observational plus 200 interventional samples for the mixed-interventional regime, $d\in\{5,10,20\}$, and 100 SCMs per graph-family/dimension/regime.
The final three-family subset therefore contains 900 graphs per regime.
Table~\ref{tab:heldout-ood-full} reports the completed baseline set for this OOD run, with methods shown under each regime where their wrappers are applicable.
``--'' means that the method is not applicable to that regime in our protocol.

\begin{table}[!htbp]
\centering
\footnotesize
\setlength{\tabcolsep}{3.5pt}
\renewcommand{\arraystretch}{1.0}
\begin{threeparttable}
\caption{\textbf{Held-out synthetic OOD benchmark.} Mean F1, SHD, and SID over the final three held-out graph families, reported separately for observation-only and mixed-interventional regimes. Values are mean (standard deviation).}
\label{tab:heldout-ood-full}
\begin{tabular*}{\textwidth}{@{\extracolsep{\fill}}lccc@{\hspace{8pt}}ccc@{}}
\toprule
& \multicolumn{3}{c}{\textbf{Observation-only}} & \multicolumn{3}{c}{\textbf{Mixed-interventional}} \\
\cmidrule(lr){2-4}\cmidrule(lr){5-7}
\textbf{Method} & \textbf{F1}$\uparrow$ & \textbf{SHD}$\downarrow$ & \textbf{SID}$\downarrow$ & \textbf{F1}$\uparrow$ & \textbf{SHD}$\downarrow$ & \textbf{SID}$\downarrow$ \\
\midrule
RandomRegress & 0.34\scorestd{0.19} & 17.61\scorestd{20.48} & 48.67\scorestd{57.01} & \NA & \NA & \NA \\
DAS & 0.38\scorestd{0.18} & 16.25\scorestd{17.00} & 46.12\scorestd{52.32} & \NA & \NA & \NA \\
LiNGAM & 0.34\scorestd{0.21} & 12.44\scorestd{13.80} & 43.06\scorestd{53.46} & \NA & \NA & \NA \\
PC & 0.38\scorestd{0.20} & 12.75\scorestd{13.90} & 47.24\scorestd{58.78} & \NA & \NA & \NA \\
CDIS & 0.24\scorestd{0.20} & 13.27\scorestd{12.87} & 44.91\scorestd{54.29} & 0.39\scorestd{0.23} & 12.17\scorestd{12.97} & 42.39\scorestd{54.10} \\
GIES & \underline{0.41}\scorestd{0.20} & 13.12\scorestd{14.28} & 43.58\scorestd{49.72} & \underline{0.58}\scorestd{0.18} & 12.52\scorestd{13.52} & 38.83\scorestd{44.06} \\
IGSP & 0.41\scorestd{0.22} & 12.58\scorestd{14.82} & 43.66\scorestd{54.26} & 0.38\scorestd{0.21} & 12.74\scorestd{15.01} & 43.67\scorestd{54.49} \\
NOTEARS & 0.31\scorestd{0.20} & 12.42\scorestd{13.44} & 43.96\scorestd{52.74} & \NA & \NA & \NA \\
NOTEARS-MLP & 0.37\scorestd{0.20} & 13.24\scorestd{13.73} & 45.63\scorestd{53.39} & \NA & \NA & \NA \\
NoDAGS & \NA & \NA & \NA & 0.42\scorestd{0.20} & 10.56\scorestd{12.14} & \textbf{31.98}\scorestd{44.66} \\
DCDI & 0.26\scorestd{0.05} & 83.50\scorestd{63.39} & 103.88\scorestd{81.75} & 0.28\scorestd{0.14} & 66.30\scorestd{68.29} & 104.08\scorestd{113.18} \\
AVICI & 0.38\scorestd{0.21} & \underline{11.51}\scorestd{12.34} & \textbf{41.83}\scorestd{50.92} & 0.55\scorestd{0.21} & \underline{10.08}\scorestd{11.70} & \underline{37.22}\scorestd{47.70} \\
\midrule
\method{} & \textbf{0.42}\scorestd{0.22} & \textbf{11.37}\scorestd{12.25} & \underline{42.31}\scorestd{51.31} & \textbf{0.60}\scorestd{0.20} & \textbf{9.58}\scorestd{10.99} & \underline{35.95}\scorestd{45.16} \\
\bottomrule
\end{tabular*}
\end{threeparttable}
\end{table}

\method{} obtains the best F1 and SHD in both regimes.
The margin is clearest when interventional evidence is available: \method{} improves over the strongest non-\method{} F1 baseline, GIES, while also reducing SHD relative to both GIES and AVICI.
In the observation-only regime, GIES and IGSP remain close in F1, and \method{} has the lowest SHD. AVICI has the lowest SID.

We also test interventions whose form differs from the default resampling-style protocol used in pretraining.
Table~\ref{tab:heldout-intervention-protocol} compares the default OOD mixed-interventional protocol with fixed-setpoint interventions and multi-target-per-sample interventions.
\method{} remains strongest in both F1 and SHD.
In the fixed-setpoint setting, its F1 is close to GIES, but its SHD is substantially lower; in the multi-target setting, it is best on both metrics.
The mixed-regime gain still shows up under these other intervention forms.

\begin{table}[!htbp]
\centering
\scriptsize
\setlength{\tabcolsep}{1.8pt}
\renewcommand{\arraystretch}{1.05}
\begin{threeparttable}
\caption{\textbf{Held-out intervention protocols on the synthetic OOD benchmark.} Mean F1, SHD, and SID for observation-only data and three mixed-interventional protocols.}
\label{tab:heldout-intervention-protocol}
\begin{tabular*}{\textwidth}{@{\extracolsep{\fill}}lccc@{\hspace{4pt}}ccc@{\hspace{4pt}}ccc@{\hspace{4pt}}ccc}
\toprule
& \multicolumn{3}{c}{\textbf{Observation-only}} & \multicolumn{3}{c}{\textbf{Default int.}} & \multicolumn{3}{c}{\textbf{Fixed-setpoint}} & \multicolumn{3}{c}{\textbf{Multi-target}} \\
\cmidrule(lr){2-4}\cmidrule(lr){5-7}\cmidrule(lr){8-10}\cmidrule(lr){11-13}
\textbf{Method} & \textbf{F1}$\uparrow$ & \textbf{SHD}$\downarrow$ & \textbf{SID}$\downarrow$ & \textbf{F1}$\uparrow$ & \textbf{SHD}$\downarrow$ & \textbf{SID}$\downarrow$ & \textbf{F1}$\uparrow$ & \textbf{SHD}$\downarrow$ & \textbf{SID}$\downarrow$ & \textbf{F1}$\uparrow$ & \textbf{SHD}$\downarrow$ & \textbf{SID}$\downarrow$ \\
\midrule
RandomRegress & 0.34 & 17.61 & 48.67 & \NA & \NA & \NA & \NA & \NA & \NA & \NA & \NA & \NA \\
DAS & 0.38 & 16.25 & 46.12 & \NA & \NA & \NA & \NA & \NA & \NA & \NA & \NA & \NA \\
LiNGAM & 0.34 & 12.44 & 43.06 & \NA & \NA & \NA & \NA & \NA & \NA & \NA & \NA & \NA \\
PC & 0.38 & 12.75 & 47.24 & \NA & \NA & \NA & \NA & \NA & \NA & \NA & \NA & \NA \\
CDIS & 0.24 & 13.27 & 44.91 & 0.39 & 12.17 & 42.39 & 0.393 & 12.09 & 42.38 & 0.306 & 13.09 & 43.84 \\
GIES & \underline{0.41} & 13.12 & 43.58 & \underline{0.58} & 12.52 & 38.83 & \underline{0.590} & 12.56 & 38.85 & \underline{0.597} & 11.85 & 37.83 \\
IGSP & \underline{0.41} & 12.58 & 43.66 & 0.38 & 12.74 & 43.67 & 0.384 & 12.79 & 43.87 & 0.178 & 14.13 & 43.93 \\
NOTEARS & 0.31 & 12.42 & 43.96 & \NA & \NA & \NA & \NA & \NA & \NA & \NA & \NA & \NA \\
NOTEARS-MLP & 0.37 & 13.24 & 45.63 & \NA & \NA & \NA & \NA & \NA & \NA & \NA & \NA & \NA \\
NoDAGS & \NA & \NA & \NA & 0.42 & 10.56 & \textbf{31.98} & 0.273 & 13.18 & 41.30 & 0.269 & 13.27 & 42.36 \\
DCDI & 0.26 & 83.50 & 103.88 & 0.28 & 66.30 & 104.08 & 0.301 & 67.84 & 107.15 & 0.294 & 66.61 & 104.39 \\
AVICI & 0.38 & \underline{11.51} & \textbf{41.83} & 0.55 & \underline{10.08} & 37.22 & 0.563 & \underline{9.90} & \underline{36.60} & 0.566 & \underline{9.78} & \underline{36.83} \\
\midrule
\method{} & \textbf{0.42} & \textbf{11.37} & \underline{42.31} & \textbf{0.60} & \textbf{9.58} & \underline{35.95} & \textbf{0.594} & \textbf{9.77} & \textbf{36.19} & \textbf{0.618} & \textbf{9.03} & \textbf{34.65} \\
\bottomrule
\end{tabular*}
\end{threeparttable}
\end{table}

\subsection{Predicting Graph Statistics from Embeddings}
\label{sec:analysis-graphstats}

To quantify how much structural information is captured by embeddings,
we first examine embedding distances between different causal substructures,
then test whether pooled embeddings can predict global graph statistics.
\begin{figure}[t]
  \centering
  \includegraphics[width=0.8\linewidth]{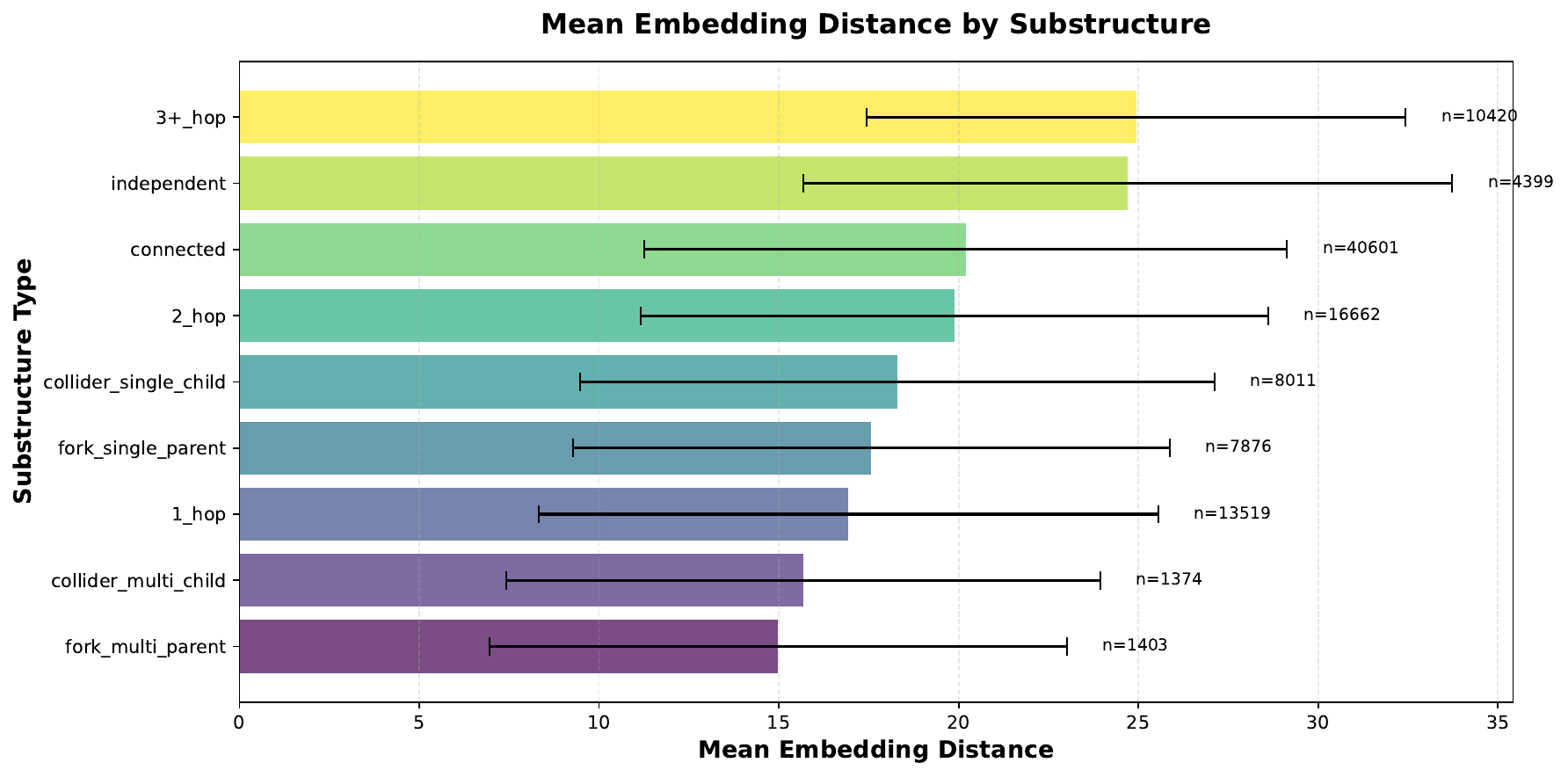}
  \caption{\textbf{Embedding distances vs.\ causal substructures.}
  Distances increase monotonically with path length, 
  structures with shared parents/children have smallest distances.}
  \label{fig:embed-distances}
\end{figure}
Figure~\ref{fig:embed-distances} shows average embedding distances for different causal relationships.
Distances increase monotonically with causal path length (1-hop: 17.2 < 2-hop: 20.1 < 3+hop: 26.3),
Forks (14.6) and colliders (15.8) stay closest; independent pairs (24.8) sit farther apart.
Nodes that are closer on the graph therefore have more similar embeddings.

We further test whether pooled embeddings can predict global graph properties.
We train lightweight Ridge regression~\citep{ridge} probes to predict four statistics from pooled embeddings 
$\widetilde{\mathbf{h}}_{\mathrm{set}} = \mathrm{mean}(\{\widetilde{\mathbf{h}}_i\}_{i=1}^{d})$:
edge count $|E^\star|$, average degree $\bar{d}$, maximum in-degree, and DAG depth.
We compare against raw data features (correlation matrix statistics, means, standard deviations) and random projections as baselines.

\begin{figure}[t]
  \centering
  \includegraphics[width=0.92\linewidth]{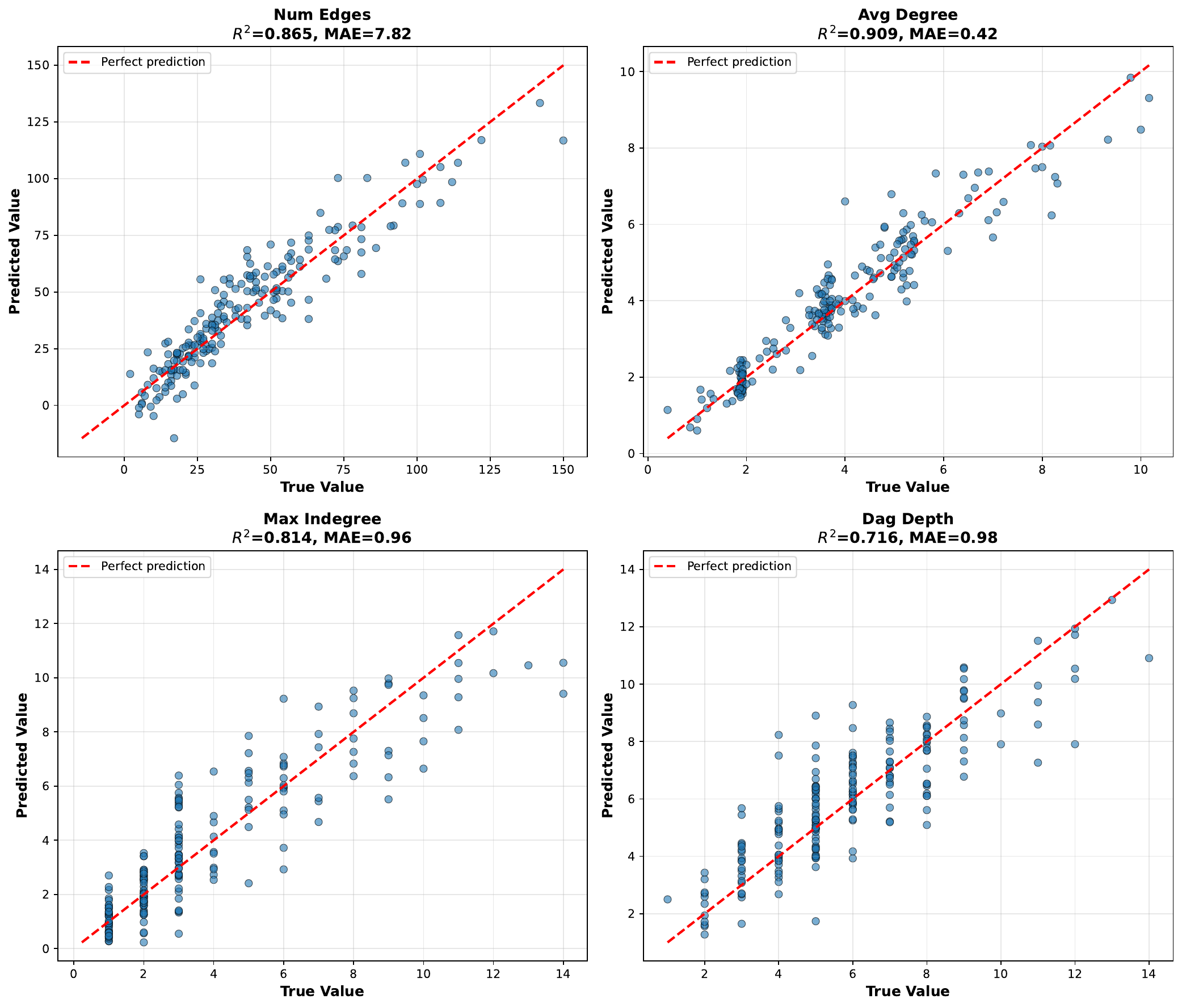}
  \caption{
  \textbf{Predicting graph statistics from embeddings.}
True vs.\ predicted values for four graph properties (edge count, average degree, max in-degree, and DAG depth).
Simple probes on our learned embeddings achieve strong fits ($R^2 \ge 0.71$), substantially outperforming raw-feature baselines.
  }
  \label{fig:graph-stat-pred}
\end{figure}

Figure~\ref{fig:graph-stat-pred} shows that embedding-based probes achieve strong prediction accuracy across all targets ($R^2 \ge 0.71$).
Table~\ref{tab:graph-stat} quantifies the performance: embeddings reduce MAE by 30-62\% compared to raw feature baselines,
with the largest gain for average degree (61.7\% reduction).
Maximum in-degree and DAG depth are also predicted well, so the embeddings carry directed topology as well as undirected connectivity.
The probes suggest the representations can be reused for simple graph-level predictions.

\begin{table}[t]
\centering
\setlength{\tabcolsep}{6pt}
\renewcommand{\arraystretch}{1.12}
\caption{Graph statistic prediction performance. 
Embedding probes reduce MAE by 30-62\% compared to raw features.}
\label{tab:graph-stat}
\begin{tabular}{lcccc}
\toprule
\textbf{Target} & \textbf{Embed MAE}$\downarrow$ & \textbf{Raw MAE} & \textbf{Embed $R^2$}$\uparrow$ & \textbf{MAE Gain} \\
\midrule
Edge count $|E^\star|$  & 7.82  & 12.67 & 0.865 & $-38.3\%$ \\
Average degree $\bar{d}$ & 0.42  & 1.09  & 0.909 & $-61.7\%$ \\
Max in-degree            & 0.96  & 1.79  & 0.814 & $-46.7\%$ \\
DAG depth                & 0.98  & 1.41  & 0.716 & $-30.3\%$ \\
\bottomrule
\end{tabular}
\end{table}

\begin{figure}[t]
  \centering
  \includegraphics[width=0.9\linewidth]{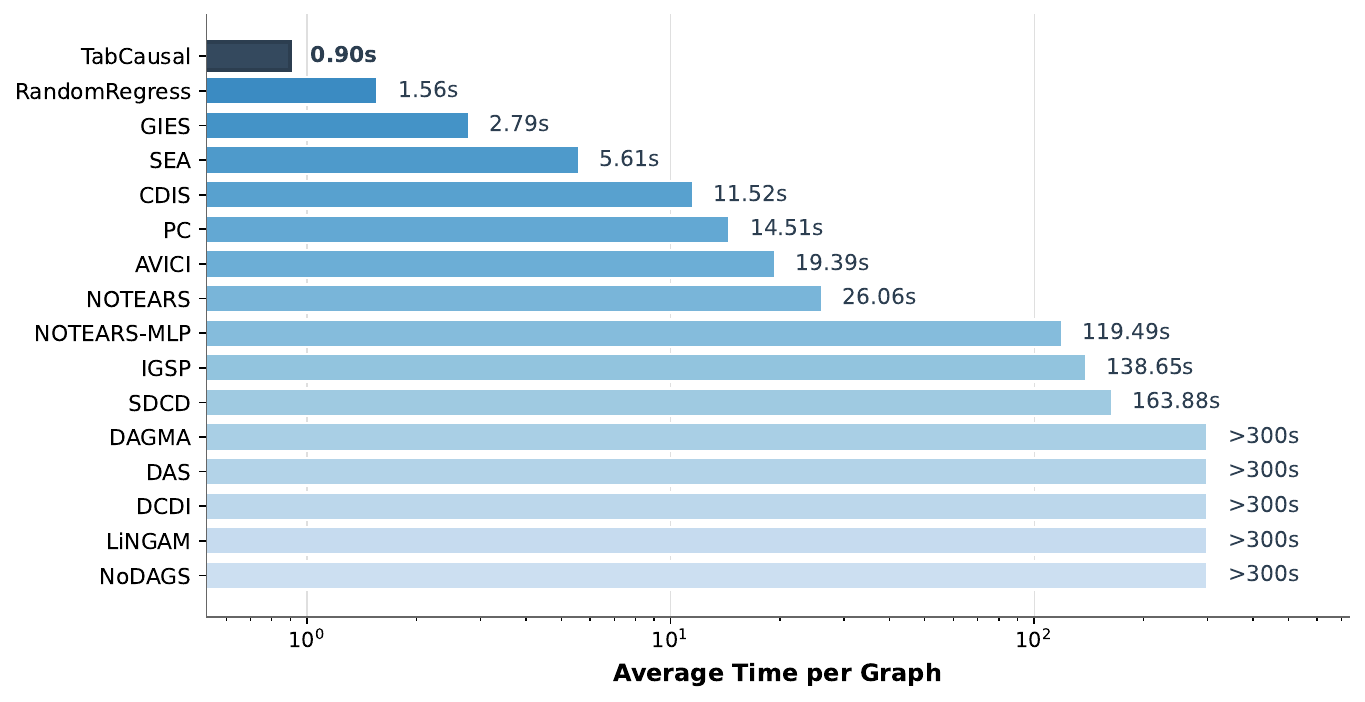}
  \caption{\textbf{Computational efficiency.}
  Mean wall-clock time per graph (seconds) on observational \texttt{gp\_hard\_obs} with $d{=}100$ variables, averaged over 10 graphs per method under the shared five-minute per-graph budget.
  \method{} performs amortized inference in a single forward pass, whereas several baselines rely on per-instance testing, search, or iterative optimization and exhibit much larger per-graph means in this view.
  Reported seconds are produced by our benchmarking harness and can vary with hardware and implementation choices.
  }
  \label{fig:runtime}
\end{figure}

\subsection{Computational Efficiency}
\label{sec:efficiency}
For runtime, we evaluate each method on observational \texttt{gp\_hard\_obs} with $d{=}100$ variables and report the average wall-clock time per graph over 10 graphs under the shared five-minute per-graph budget.
Figure~\ref{fig:runtime} visualizes these per-method means in seconds.
\method{} runs in one forward pass. Many classical and neural baselines test, search, or fit on each graph, so their average time is much higher under this protocol.
Absolute seconds and rank order can move with hardware, batching, solvers, and engineering; the difference between one-pass prediction and per-graph search is still visible here.

\subsection{Embedding Quality Analysis}
\label{app:embedding-quality}

To understand when \method{} succeeds or fails, we analyze embedding quality across three difficulty factors:
sample size ($N \in \{10, 100, 1000\}$), dimension ($d \in \{5, 10, 30, 100\}$), and graph complexity (edge density $p \in \{0.05, 0.1, 0.2, 0.4\}$).
For each condition, we generate 50 random DAGs and evaluate three metrics:
(1) Separability: mean distance between non-edges minus mean distance between true edges,
(2) k-NN Accuracy: proportion of k-nearest neighbors that are true graph neighbors,
(3) Effective Dimensionality: number of principal components explaining 90\% variance.

Table~\ref{tab:embedding-quality} shows the results across all conditions.
Dimension is the largest of the three factors.
From $d=5$ to $d=100$, k-NN accuracy drops from 1.000 to 0.383 ($-62\%$) and separability from 7.44 to 1.20 ($-84\%$),
while effective dimensionality inflates from 3 to 23 ($+667\%$).
Performance degrades sharply between $d=10$ (k-NN = 0.853) and $d=30$ (k-NN = 0.489),
suggesting a critical threshold around $d \approx 15$--$20$.

Sample size shows diminishing returns beyond $N=100$.
Increasing $N$ from 10 to 100 helps a lot: separability increases by 58\% and k-NN accuracy by 35\%.
A further increase to $N=1000$ leaves k-NN accuracy essentially unchanged; separability changes only modestly and differs across the two $N=1000$ runs,
suggesting an effective sample size of roughly $N \approx 5d$ suffices---a sample efficiency advantage over classical methods requiring $N \gg d^2$.

Graph complexity causes linear degradation.
As edge density increases from $p=0.05$ to $p=0.4$, separability decreases from 7.33 to 1.76 ($-76\%$) and k-NN accuracy from 0.923 to 0.597 ($-35\%$).
Relative to dimension, density changes more gradually.

These results suggest \method{} performs reliably when $d \lesssim 20$, $N \gtrsim 5d$, and graphs are moderately sparse ($p \lesssim 0.2$).
The sharp drop past $d \approx 20$ means going much further in dimension may need a different encoder.

\begin{table}[ht]
\centering
\small
\setlength{\tabcolsep}{5pt}
\caption{Embedding quality metrics across difficulty levels. 
Sep: Separability. k-NN: k-Nearest neighbor accuracy. Eff.Dim: Effective dimensionality.}
\label{tab:embedding-quality}
\begin{tabular}{llccc}
\toprule
\textbf{Factor} & \textbf{Level} & \textbf{Sep}$\uparrow$ & \textbf{k-NN}$\uparrow$ & \textbf{Eff.Dim}$\downarrow$ \\
\midrule
Sample Size & $N=10$        & 2.02 & 0.441 & 9 \\
            & $N=100$       & 3.19 & 0.594 & 12 \\
            & $N=1000$ (r1) & 3.63 & 0.596 & 11 \\
            & $N=1000$ (r2) & 3.22 & 0.599 & 11 \\
\midrule
Dimension   & $d=5$         & 7.44 & 1.000 & 3 \\
            & $d=10$        & 4.94 & 0.853 & 5 \\
            & $d=30$        & 2.73 & 0.489 & 15 \\
            & $d=100$       & 1.20 & 0.383 & 23 \\
\midrule
Graph       & Very sparse ($p=0.05$) & 7.33 & 0.923 & 7 \\
Complexity  & Sparse ($p=0.1$)       & 5.07 & 0.765 & 9 \\
            & Medium ($p=0.2$)       & 3.50 & 0.618 & 11 \\
            & Dense ($p=0.4$)        & 1.76 & 0.597 & 11 \\
\bottomrule
\end{tabular}
\end{table}

\section{Hyperparameter Configuration}
\label{sec:appendix_hyperparams}

Unless noted otherwise, the methods below are evaluated using the benchmark wrapper's current direct/default configuration. The appendix therefore reports the fixed wrapper settings used in the current benchmark pipeline.

\subsection{Orientation and decoding for metric computation}
\label{sec:appendix-orientation}

Reported F1, SHD, and SID compare each method's prediction to a simulator-known DAG.
Methods that output partially directed structures are therefore converted to a directed adjacency matrix before scoring.
For PC and GIES, any remaining undirected CPDAG edge between variables $i$ and $j$ is oriented as $i \rightarrow j$ when $i<j$ and as $j \rightarrow i$ otherwise, using the dataset's variable index order.
For CDIS, orientation-uncertain PAG adjacencies are resolved with the same index rule.
This conversion is fixed and deterministic: it is neither random nor informed by ground-truth orientations.
Continuous-output and probabilistic methods (e.g., \method{}, AVICI, SEA, SDCD) are binarized or thresholded using each wrapper's default decoding rule in the same evaluation pipeline.

\subsection{Method-Specific Hyperparameters}

\paragraph{\method{}.}
Pretrained transformer-based amortized dataset-to-graph predictor; reported numbers use the final pretrained checkpoint from this paper, without per-instance retraining or test-time optimization.

\paragraph{AVICI.}
Official pretrained weights released by the authors; no per-instance optimization.

\paragraph{PC.}
\begin{itemize}[leftmargin=*,nosep]
    \item Significance level $\alpha = 0.05$
    \item Conditional independence testing: Fisher-$Z$ / Gaussian tests
    \item CPDAG-to-DAG conversion for metrics: index tie-break in \cref{sec:appendix-orientation}
\end{itemize}

\paragraph{GIES.}
We use the same \texttt{pcalg::gies} wrapper in both evaluation regimes.
In observation-only settings, all intervention masks are zero, so no intervention targets are passed; the call therefore reduces to the observational GES special case.
In mixed-interventional settings, nonzero intervention masks are encoded as intervention targets and passed to the interventional GIES variant.
\begin{itemize}[leftmargin=*,nosep]
    \item Penalty: BIC-style default ($\lambda = -1.0$ in the wrapper)
    \item CPDAG-to-DAG conversion for metrics: index tie-break in \cref{sec:appendix-orientation}
\end{itemize}

\paragraph{IGSP.}
\begin{itemize}[leftmargin=*,nosep]
    \item Gaussian conditional independence testing
    \item Observational significance level $\alpha = 10^{-3}$
    \item Invariance-test significance $\alpha_{\mathrm{inv}} = 10^{-3}$
\end{itemize}

\paragraph{LiNGAM.}
DirectLiNGAM through the current wrapper with standard decoding/default handling.

\paragraph{CDIS.}
Library default discovery procedure in the current direct wrapper; PAG-to-DAG conversion for metrics follows the index tie-break in \cref{sec:appendix-orientation}.

\paragraph{NOTEARS.}
Linear NOTEARS with $\ell_1$ coefficient $0.1$.

\paragraph{NOTEARS-MLP.}
Hidden width $10$, $\lambda_1 = 0.01$, $\lambda_2 = 0.01$.

\paragraph{NoDAGS.}
\begin{itemize}[leftmargin=*,nosep]
    \item Acyclicity coefficient: $10^{-3}$
    \item Learning rate: $10^{-2}$
    \item Depth: $0$ hidden layers
    \item Training: 200 epochs, batch size 1024
    \item Minimum interventional samples per regime: 10
\end{itemize}

\paragraph{DCDI.}
\begin{itemize}[leftmargin=*,nosep]
    \item Encoder: 2 layers, hidden width 16
    \item Flow: 2 layers, hidden width 16
    \item Activation: Leaky ReLU
    \item Optimizer: RMSprop, learning rate $10^{-3}$
    \item Regularization coefficient: $0.1$
    \item Inputs normalized in the wrapper
    \item Train/test split: 80/20 within each sampled dataset
    \item Mini-batch size: $\min(64, N_{\text{train}})$
    \item Augmented Lagrangian optimization with small initial multipliers and patience-based early stopping
    \item Wall-clock budget: five minutes per graph, consistent with the shared benchmark budget; if the budget is reached, the graph predicted at the current optimizer iteration is reported
\end{itemize}

\paragraph{DAS.}
\begin{itemize}[leftmargin=*,nosep]
    \item $\eta_G = 10^{-3}$, $\eta_H = 10^{-3}$, $\alpha = 0.05$
    \item Pruning enabled
    \item Splines: degree 3 with $10$ basis functions
    \item Modest parent-set constraints per wrapper defaults
\end{itemize}

\paragraph{SEA.}
Pretrained Sample-Estimate-Aggregate causal discovery model with a released supervised aggregator; evaluated via the benchmark wrapper defaults~\citep{wu2025sampleestimateaggregaterecipe}.

\paragraph{DAGMA.}
Continuous optimization with an M-matrix/log-determinant acyclicity characterization; evaluated with the benchmark wrapper defaults~\citep{NEURIPS2022_36e2967f}.

\paragraph{SDCD.}
Stable Differentiable Causal Discovery with a spectral acyclicity constraint and sparse-graph-oriented two-stage training; evaluated with the benchmark wrapper defaults~\citep{nazaret2024stabledifferentiablecausaldiscovery}.

\paragraph{RandomRegress.}
Random-order regression sanity baseline; deliberately weak structural control.

\section{Full Synthetic Benchmark Results}

Macro-averaged synthetic scores are reported in Table~\ref{tab:main-benchmark} of the main text; this appendix gives per-configuration breakdowns and heatmaps.

\subsection{Performance Heatmap Across All Benchmarks}
\label{app:heatmap}

\begin{figure}[ht]
\centering
\includegraphics[width=0.9\textwidth]{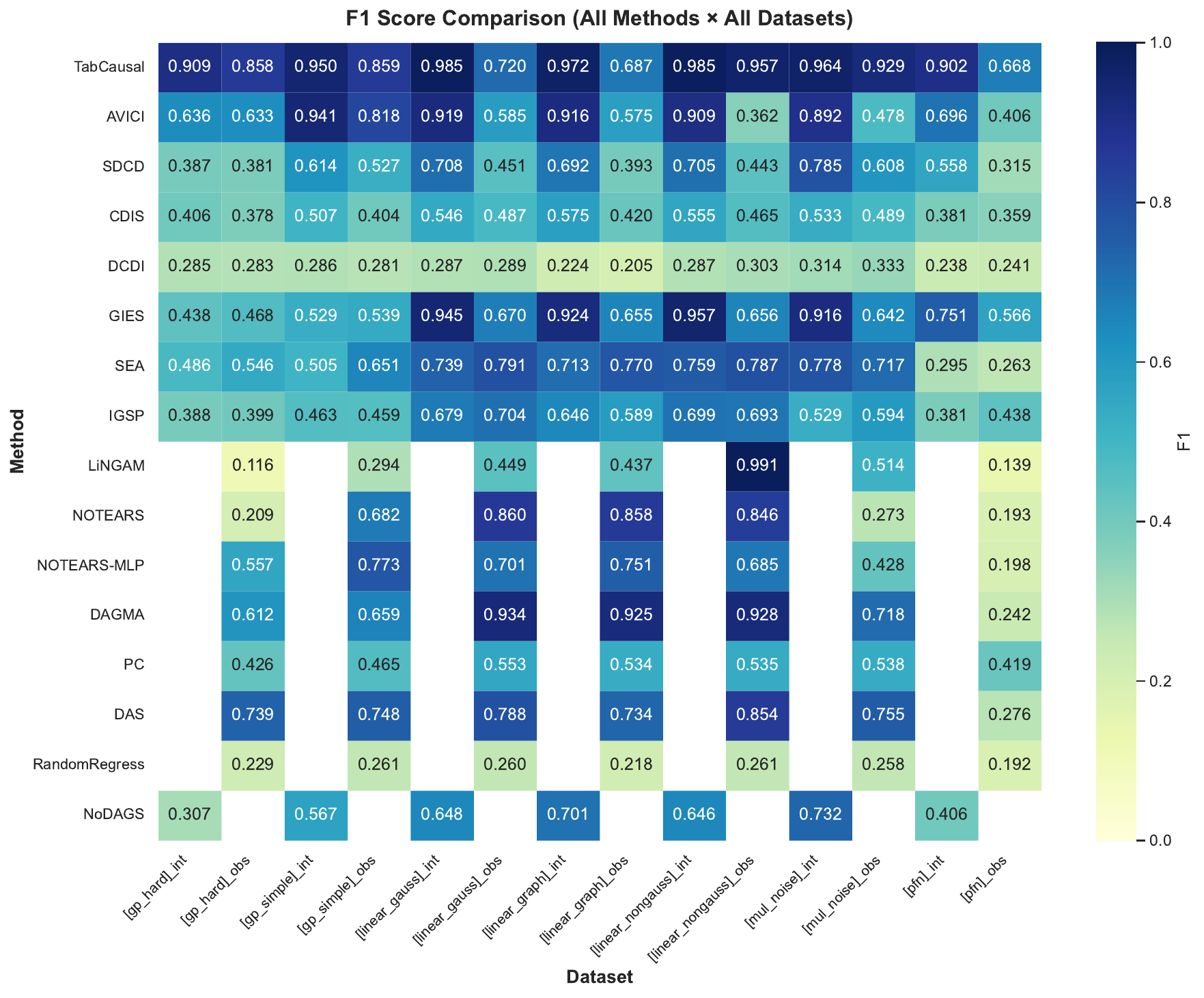}
\caption{\textbf{Detailed F1 heatmap across dataset-regime settings.}
Each cell reports F1 for a method on one observational or mixed-interventional dataset setting; blank cells denote methods not applicable or not run under that regime.}
\label{fig:f1-heatmap}
\end{figure}

Figure~\ref{fig:f1-heatmap} presents a heatmap of F1 for each method on each synthetic dataset-regime column.

Note that some baselines are not applicable to both regimes: NOTEARS, NOTEARS-MLP, LiNGAM, PC, DAS, DAGMA, and RandomRegress are evaluated only in the observational setting, while NoDAGS is evaluated only in the mixed-interventional setting; SEA and SDCD are evaluated in both regimes.

\subsection{Detailed Per-Dataset Results}
\label{app:detailed-results}

This section provides complete numerical results for all benchmarks across different dimensions and evaluation metrics.
Each table reports F1 score, Structural Hamming Distance (SHD), and Structural Intervention Distance (SID) for both observational and mixed-interventional settings; values are mean (standard deviation) over 50 benchmark graph instances per dataset setting.

Each table is organized as follows:
\begin{itemize}[nosep]
    \item Columns are grouped by evaluation setting (Observation-only vs.\ Mixed-interventional)
    \item Within each setting, results are shown for three dimensions ($d \in \{5, 10, 20\}$)
    \item For each dimension, we report three metrics: F1 (higher is better), SHD (lower is better), and SID (lower is better)
    \item "--" indicates the method is not applicable to that setting
\end{itemize}

\clearpage

\input{figs/tables/main/appendix/table_detailed_gp-hard}
\input{figs/tables/main/appendix/table_detailed_gp-simple}
\input{figs/tables/main/appendix/table_detailed_linear-gauss}
\input{figs/tables/main/appendix/table_detailed_linear-graph}
\input{figs/tables/main/appendix/table_detailed_linear-nongauss}
\input{figs/tables/main/appendix/table_detailed_mul-noise}
\input{figs/tables/main/appendix/table_detailed_pfn}

%% file: figs/tables/semantic/table_semantic_slices.tex
\begin{table*}[t]
\centering
\footnotesize
\setlength{\tabcolsep}{2.5pt}
\renewcommand{\arraystretch}{1.05}
\caption{Semantic benchmark breakdown by analysis slice (F1 mean and standard deviation across semantic scenarios), reported separately for observation-only and mixed-interventional regimes.}
\label{tab:semantic-slice-breakdown}
\textbf{Observation-only}\\[3pt]
\begin{tabular}{@{}l@{\hspace{4pt}}c@{\hspace{3pt}}c@{\hspace{3pt}}c@{\hspace{3pt}}c@{\hspace{3pt}}c@{\hspace{3pt}}c@{\hspace{3pt}}c@{\hspace{3pt}}c@{}}
\toprule
Slice & \method{} & AVICI & SDCD & CDIS & DCDI & GIES & SEA & IGSP \\
\midrule
mix\_amb. & \underline{0.681}\scorestd{0.174} & 0.560\scorestd{0.188} & 0.479\scorestd{0.152} & 0.595\scorestd{0.165} & 0.338\scorestd{0.121} & \textbf{0.698}\scorestd{0.210} & 0.471\scorestd{0.137} & 0.664\scorestd{0.185} \\
meas\_fail. & \underline{0.691}\scorestd{0.163} & 0.560\scorestd{0.175} & 0.484\scorestd{0.156} & 0.608\scorestd{0.166} & 0.330\scorestd{0.125} & \textbf{0.720}\scorestd{0.210} & 0.485\scorestd{0.130} & 0.676\scorestd{0.182} \\
proxy\_interp. & \underline{0.688}\scorestd{0.173} & 0.566\scorestd{0.181} & 0.502\scorestd{0.147} & 0.589\scorestd{0.160} & 0.336\scorestd{0.123} & \textbf{0.706}\scorestd{0.211} & 0.479\scorestd{0.143} & 0.681\scorestd{0.170} \\
followup\_br. & 0.675\scorestd{0.155} & 0.553\scorestd{0.177} & 0.494\scorestd{0.150} & 0.566\scorestd{0.153} & 0.343\scorestd{0.113} & \textbf{0.704}\scorestd{0.213} & 0.468\scorestd{0.145} & \underline{0.702}\scorestd{0.166} \\
detect\_fail. & 0.674\scorestd{0.128} & 0.537\scorestd{0.169} & 0.459\scorestd{0.156} & 0.586\scorestd{0.179} & 0.328\scorestd{0.125} & \textbf{0.714}\scorestd{0.209} & 0.453\scorestd{0.132} & \underline{0.679}\scorestd{0.169} \\
queue\_press. & \textbf{0.650}\scorestd{0.178} & 0.548\scorestd{0.196} & 0.468\scorestd{0.145} & 0.606\scorestd{0.175} & 0.354\scorestd{0.124} & \underline{0.647}\scorestd{0.208} & 0.460\scorestd{0.153} & 0.624\scorestd{0.179} \\
triage\_thr. & \underline{0.701}\scorestd{0.207} & 0.594\scorestd{0.214} & 0.507\scorestd{0.140} & 0.611\scorestd{0.128} & 0.338\scorestd{0.123} & \textbf{0.706}\scorestd{0.217} & 0.480\scorestd{0.152} & 0.659\scorestd{0.181} \\
alarm\_drift & 0.640\scorestd{0.125} & 0.574\scorestd{0.210} & 0.489\scorestd{0.160} & \underline{0.737}\scorestd{0.145} & 0.377\scorestd{0.148} & \textbf{0.789}\scorestd{0.174} & 0.469\scorestd{0.121} & 0.705\scorestd{0.204} \\
heavy\_tail & \textbf{0.682}\scorestd{0.266} & 0.548\scorestd{0.251} & 0.382\scorestd{0.134} & 0.466\scorestd{0.144} & 0.332\scorestd{0.102} & 0.570\scorestd{0.159} & 0.393\scorestd{0.099} & 0.527\scorestd{0.215} \\
reciprocal & \textbf{0.682}\scorestd{0.266} & 0.548\scorestd{0.251} & 0.382\scorestd{0.134} & 0.466\scorestd{0.144} & 0.332\scorestd{0.102} & 0.570\scorestd{0.159} & 0.393\scorestd{0.099} & 0.527\scorestd{0.215} \\
\bottomrule
\end{tabular}

\vspace{4pt}
\begin{tabular}{@{}l@{\hspace{4pt}}c@{\hspace{3pt}}c@{\hspace{3pt}}c@{\hspace{3pt}}c@{\hspace{3pt}}c@{\hspace{3pt}}c@{\hspace{3pt}}c@{}}
\toprule
Slice & LiNGAM & NOTEARS & NOTEARS-MLP & DAGMA & PC & DAS & RandomRegress \\
\midrule
mix\_amb. & 0.303\scorestd{0.176} & 0.203\scorestd{0.130} & 0.270\scorestd{0.131} & 0.276\scorestd{0.140} & 0.649\scorestd{0.172} & 0.242\scorestd{0.099} & 0.206\scorestd{0.044} \\
meas\_fail. & 0.334\scorestd{0.172} & 0.229\scorestd{0.128} & 0.288\scorestd{0.128} & 0.294\scorestd{0.135} & 0.665\scorestd{0.157} & 0.247\scorestd{0.096} & 0.208\scorestd{0.048} \\
proxy\_interp. & 0.302\scorestd{0.168} & 0.201\scorestd{0.125} & 0.249\scorestd{0.129} & 0.270\scorestd{0.138} & 0.652\scorestd{0.162} & 0.236\scorestd{0.094} & 0.208\scorestd{0.041} \\
followup\_br. & 0.303\scorestd{0.176} & 0.204\scorestd{0.130} & 0.271\scorestd{0.133} & 0.275\scorestd{0.122} & 0.670\scorestd{0.164} & 0.243\scorestd{0.098} & 0.212\scorestd{0.036} \\
detect\_fail. & 0.310\scorestd{0.163} & 0.219\scorestd{0.126} & 0.287\scorestd{0.122} & 0.282\scorestd{0.114} & 0.665\scorestd{0.177} & 0.251\scorestd{0.101} & 0.207\scorestd{0.045} \\
queue\_press. & 0.257\scorestd{0.163} & 0.145\scorestd{0.107} & 0.238\scorestd{0.145} & 0.236\scorestd{0.152} & 0.608\scorestd{0.189} & 0.230\scorestd{0.102} & 0.203\scorestd{0.039} \\
triage\_thr. & 0.237\scorestd{0.153} & 0.198\scorestd{0.154} & 0.243\scorestd{0.141} & 0.250\scorestd{0.154} & 0.630\scorestd{0.187} & 0.242\scorestd{0.115} & 0.192\scorestd{0.039} \\
alarm\_drift & 0.246\scorestd{0.151} & 0.199\scorestd{0.133} & 0.315\scorestd{0.102} & 0.272\scorestd{0.141} & 0.627\scorestd{0.276} & 0.257\scorestd{0.128} & 0.188\scorestd{0.055} \\
heavy\_tail & 0.301\scorestd{0.231} & 0.142\scorestd{0.096} & 0.229\scorestd{0.097} & 0.277\scorestd{0.176} & \underline{0.593}\scorestd{0.161} & 0.217\scorestd{0.069} & 0.210\scorestd{0.047} \\
reciprocal & 0.301\scorestd{0.231} & 0.142\scorestd{0.096} & 0.229\scorestd{0.097} & 0.277\scorestd{0.176} & \underline{0.593}\scorestd{0.161} & 0.217\scorestd{0.069} & 0.210\scorestd{0.047} \\
\bottomrule
\end{tabular}

\vspace{8pt}
\textbf{Mixed-interventional}\\[3pt]
\begin{tabular}{@{}l@{\hspace{4pt}}c@{\hspace{3pt}}c@{\hspace{3pt}}c@{\hspace{3pt}}c@{\hspace{3pt}}c@{\hspace{3pt}}c@{\hspace{3pt}}c@{\hspace{3pt}}c@{\hspace{3pt}}c@{}}
\toprule
Slice & \method{} & AVICI & SDCD & CDIS & DCDI & GIES & SEA & IGSP & NoDAGS \\
\midrule
mix\_amb. & \textbf{0.971}\scorestd{0.057} & \underline{0.943}\scorestd{0.079} & 0.780\scorestd{0.187} & 0.604\scorestd{0.166} & 0.435\scorestd{0.181} & 0.882\scorestd{0.080} & 0.321\scorestd{0.122} & 0.611\scorestd{0.233} & 0.864\scorestd{0.125} \\
meas\_fail. & \textbf{0.973}\scorestd{0.042} & \underline{0.946}\scorestd{0.074} & 0.795\scorestd{0.187} & 0.623\scorestd{0.152} & 0.418\scorestd{0.168} & 0.900\scorestd{0.064} & 0.315\scorestd{0.116} & 0.584\scorestd{0.244} & 0.877\scorestd{0.127} \\
proxy\_interp. & \textbf{0.976}\scorestd{0.036} & \underline{0.948}\scorestd{0.068} & 0.794\scorestd{0.178} & 0.608\scorestd{0.139} & 0.445\scorestd{0.180} & 0.880\scorestd{0.078} & 0.326\scorestd{0.122} & 0.613\scorestd{0.255} & 0.853\scorestd{0.131} \\
followup\_br. & \textbf{0.978}\scorestd{0.031} & \underline{0.950}\scorestd{0.059} & 0.798\scorestd{0.187} & 0.591\scorestd{0.138} & 0.440\scorestd{0.176} & 0.886\scorestd{0.082} & 0.320\scorestd{0.130} & 0.637\scorestd{0.235} & 0.849\scorestd{0.140} \\
detect\_fail. & \textbf{0.972}\scorestd{0.040} & \underline{0.943}\scorestd{0.072} & 0.789\scorestd{0.195} & 0.600\scorestd{0.155} & 0.409\scorestd{0.158} & 0.907\scorestd{0.064} & 0.304\scorestd{0.122} & 0.564\scorestd{0.208} & 0.859\scorestd{0.160} \\
queue\_press. & \textbf{0.969}\scorestd{0.075} & \underline{0.935}\scorestd{0.084} & 0.756\scorestd{0.180} & 0.585\scorestd{0.183} & 0.477\scorestd{0.200} & 0.851\scorestd{0.088} & 0.341\scorestd{0.133} & 0.645\scorestd{0.211} & 0.841\scorestd{0.110} \\
triage\_thr. & \textbf{0.971}\scorestd{0.040} & \underline{0.936}\scorestd{0.076} & 0.751\scorestd{0.179} & 0.616\scorestd{0.127} & 0.466\scorestd{0.212} & 0.842\scorestd{0.089} & 0.363\scorestd{0.128} & 0.647\scorestd{0.238} & 0.858\scorestd{0.106} \\
alarm\_drift & \textbf{0.974}\scorestd{0.050} & 0.923\scorestd{0.094} & 0.758\scorestd{0.215} & 0.738\scorestd{0.158} & 0.442\scorestd{0.220} & \underline{0.926}\scorestd{0.076} & 0.293\scorestd{0.114} & 0.656\scorestd{0.166} & 0.915\scorestd{0.099} \\
heavy\_tail & \textbf{0.935}\scorestd{0.141} & \underline{0.932}\scorestd{0.139} & 0.731\scorestd{0.200} & 0.460\scorestd{0.265} & 0.403\scorestd{0.185} & 0.839\scorestd{0.092} & 0.275\scorestd{0.100} & 0.581\scorestd{0.195} & 0.849\scorestd{0.132} \\
reciprocal & \textbf{0.935}\scorestd{0.141} & \underline{0.932}\scorestd{0.139} & 0.731\scorestd{0.200} & 0.460\scorestd{0.265} & 0.403\scorestd{0.185} & 0.839\scorestd{0.092} & 0.275\scorestd{0.100} & 0.581\scorestd{0.195} & 0.849\scorestd{0.132} \\
\bottomrule
\end{tabular}
\end{table*}

%% file: figs/tables/main/appendix/table_detailed_gp-hard.tex
\begin{table*}[htbp]
\centering
\small
\caption{Detailed results for \texttt{gp\_hard} across different graph sizes ($d$) and metrics.}
\label{tab:detail-gp-hard}
\setlength{\tabcolsep}{4pt}
\resizebox{\textwidth}{!}{
\begin{tabular}{lccccccccc@{\hspace{6pt}}ccccccccc}
\toprule
& \multicolumn{9}{c}{\textbf{Observation-only}} & \multicolumn{9}{c}{\textbf{Mixed-interventional}} \\
\cmidrule(lr){2-10}\cmidrule(lr){11-19}
\textbf{Method} & \multicolumn{3}{c}{$d=5$} & \multicolumn{3}{c}{$d=10$} & \multicolumn{3}{c}{$d=20$} & \multicolumn{3}{c}{$d=5$} & \multicolumn{3}{c}{$d=10$} & \multicolumn{3}{c}{$d=20$} \\
\cmidrule(lr){2-10}\cmidrule(lr){11-19}
 & F1 & SHD & SID & F1 & SHD & SID & F1 & SHD & SID & F1 & SHD & SID & F1 & SHD & SID & F1 & SHD & SID \\
\midrule
RandomRegress & 0.30\scorestd{0.24} & 6.8\scorestd{2.4} & 9.3\scorestd{3.8} & 0.23\scorestd{0.12} & 32.6\scorestd{9.2} & 41.2\scorestd{9.8} & 0.16\scorestd{0.08} & 140.2\scorestd{56.8} & 168.7\scorestd{48.2} & -- & -- & -- & -- & -- & -- & -- & -- & -- \\
DAS & \underline{0.83}\scorestd{0.17} & \underline{1.3}\scorestd{1.2} & \underline{2.0}\scorestd{2.4} & \underline{0.71}\scorestd{0.12} & \underline{7.7}\scorestd{3.4} & \underline{12.9}\scorestd{6.7} & \underline{0.61}\scorestd{0.09} & 25.2\scorestd{7.8} & \underline{61.4}\scorestd{20.1} & -- & -- & -- & -- & -- & -- & -- & -- & -- \\
LiNGAM & 0.11\scorestd{0.20} & 4.6\scorestd{1.7} & 6.2\scorestd{2.6} & 0.13\scorestd{0.12} & 14.5\scorestd{3.8} & 24.3\scorestd{7.0} & 0.11\scorestd{0.07} & 37.1\scorestd{8.4} & 83.6\scorestd{25.0} & -- & -- & -- & -- & -- & -- & -- & -- & -- \\
PC & 0.47\scorestd{0.29} & 3.3\scorestd{1.8} & 5.7\scorestd{3.6} & 0.43\scorestd{0.15} & 12.0\scorestd{4.8} & 26.3\scorestd{10.1} & 0.38\scorestd{0.11} & 33.7\scorestd{8.8} & 99.2\scorestd{30.1} & -- & -- & -- & -- & -- & -- & -- & -- & -- \\
CDIS & 0.38\scorestd{0.29} & 3.7\scorestd{1.7} & 5.7\scorestd{3.5} & 0.40\scorestd{0.15} & 12.6\scorestd{4.8} & 28.4\scorestd{11.4} & 0.36\scorestd{0.09} & 34.7\scorestd{8.7} & 108.1\scorestd{33.1} & 0.45\scorestd{0.27} & 3.5\scorestd{2.0} & 5.9\scorestd{3.9} & 0.42\scorestd{0.16} & 11.6\scorestd{4.7} & 25.1\scorestd{10.6} & 0.35\scorestd{0.10} & 36.1\scorestd{10.2} & 105.8\scorestd{32.2} \\
GIES & 0.49\scorestd{0.31} & 3.3\scorestd{2.1} & 5.8\scorestd{4.4} & 0.46\scorestd{0.16} & 12.6\scorestd{5.1} & 28.0\scorestd{12.3} & 0.44\scorestd{0.11} & 34.4\scorestd{9.7} & 107.6\scorestd{33.3} & 0.48\scorestd{0.25} & 3.7\scorestd{2.2} & 6.5\scorestd{4.1} & 0.43\scorestd{0.14} & 13.0\scorestd{4.9} & 29.6\scorestd{11.7} & 0.38\scorestd{0.09} & 40.0\scorestd{10.6} & 121.5\scorestd{31.7} \\
IGSP & 0.44\scorestd{0.29} & 3.4\scorestd{1.9} & 6.2\scorestd{4.2} & 0.39\scorestd{0.15} & 12.6\scorestd{4.7} & 27.8\scorestd{10.1} & 0.37\scorestd{0.09} & 33.6\scorestd{9.2} & 101.0\scorestd{27.8} & 0.43\scorestd{0.29} & 3.5\scorestd{2.0} & 6.2\scorestd{3.9} & 0.40\scorestd{0.15} & 11.3\scorestd{4.2} & 26.5\scorestd{10.4} & 0.34\scorestd{0.09} & 35.6\scorestd{9.6} & 101.9\scorestd{28.7} \\
DAGMA & 0.66\scorestd{0.23} & 2.1\scorestd{1.4} & 3.4\scorestd{3.2} & 0.62\scorestd{0.15} & 9.4\scorestd{4.4} & 16.9\scorestd{8.3} & 0.55\scorestd{0.13} & \underline{24.4}\scorestd{8.7} & 64.4\scorestd{22.4} & -- & -- & -- & -- & -- & -- & -- & -- & -- \\
NOTEARS & 0.25\scorestd{0.25} & 4.0\scorestd{1.6} & 5.2\scorestd{2.3} & 0.21\scorestd{0.13} & 13.5\scorestd{4.0} & 22.6\scorestd{7.1} & 0.16\scorestd{0.09} & 35.2\scorestd{8.6} & 80.6\scorestd{25.3} & -- & -- & -- & -- & -- & -- & -- & -- & -- \\
NOTEARS-MLP & 0.56\scorestd{0.29} & 2.7\scorestd{1.5} & 3.7\scorestd{2.8} & 0.57\scorestd{0.17} & 9.1\scorestd{3.8} & 16.2\scorestd{7.8} & 0.53\scorestd{0.12} & 25.0\scorestd{7.7} & 64.9\scorestd{22.3} & -- & -- & -- & -- & -- & -- & -- & -- & -- \\
NoDAGS & -- & -- & -- & -- & -- & -- & -- & -- & -- & 0.38\scorestd{0.26} & 3.7\scorestd{1.8} & 4.8\scorestd{2.6} & 0.31\scorestd{0.16} & 12.6\scorestd{4.6} & 21.3\scorestd{7.9} & 0.20\scorestd{0.08} & 38.1\scorestd{9.0} & 89.2\scorestd{22.9} \\
SEA & 0.63\scorestd{0.22} & 2.5\scorestd{1.6} & 3.6\scorestd{2.6} & 0.55\scorestd{0.16} & 10.5\scorestd{4.8} & 18.2\scorestd{7.6} & 0.47\scorestd{0.12} & 27.5\scorestd{8.3} & 70.1\scorestd{23.5} & 0.48\scorestd{0.29} & 2.9\scorestd{1.8} & 4.1\scorestd{2.7} & 0.51\scorestd{0.14} & 10.4\scorestd{3.7} & 17.8\scorestd{7.1} & \underline{0.46}\scorestd{0.10} & 33.6\scorestd{9.5} & 72.8\scorestd{20.6} \\
SDCD & 0.42\scorestd{0.26} & 3.6\scorestd{1.9} & 7.0\scorestd{3.9} & 0.38\scorestd{0.14} & 13.8\scorestd{4.9} & 31.3\scorestd{9.8} & 0.35\scorestd{0.12} & 34.6\scorestd{9.6} & 101.2\scorestd{27.0} & 0.41\scorestd{0.26} & 3.6\scorestd{2.1} & 6.9\scorestd{4.2} & 0.39\scorestd{0.16} & 12.4\scorestd{4.5} & 30.9\scorestd{11.2} & 0.36\scorestd{0.11} & 37.8\scorestd{9.3} & 109.0\scorestd{24.2} \\
DCDI & 0.37\scorestd{0.11} & 10.0\scorestd{0.0} & 14.3\scorestd{2.2} & 0.30\scorestd{0.06} & 45.0\scorestd{0.0} & 64.3\scorestd{7.0} & 0.18\scorestd{0.03} & 189.8\scorestd{1.3} & 298.5\scorestd{22.2} & 0.38\scorestd{0.13} & 10.0\scorestd{0.0} & 14.2\scorestd{2.6} & 0.28\scorestd{0.07} & 45.0\scorestd{0.0} & 66.7\scorestd{6.5} & 0.20\scorestd{0.04} & 190.0\scorestd{0.0} & 288.3\scorestd{22.8} \\
AVICI & 0.72\scorestd{0.25} & 1.8\scorestd{1.4} & 2.7\scorestd{2.6} & 0.64\scorestd{0.16} & 8.2\scorestd{4.0} & 14.7\scorestd{7.7} & 0.46\scorestd{0.12} & 29.1\scorestd{9.9} & 69.8\scorestd{24.9} & \underline{0.79}\scorestd{0.23} & \underline{1.5}\scorestd{1.5} & \underline{1.9}\scorestd{2.1} & \underline{0.66}\scorestd{0.18} & \underline{7.4}\scorestd{4.3} & \underline{13.4}\scorestd{7.9} & \underline{0.46}\scorestd{0.14} & \underline{28.9}\scorestd{9.9} & \underline{70.4}\scorestd{24.5} \\
\method{} & \textbf{0.88}\scorestd{0.15} & \textbf{0.9}\scorestd{1.0} & \textbf{1.1}\scorestd{1.5} & \textbf{0.88}\scorestd{0.10} & \textbf{3.3}\scorestd{2.7} & \textbf{5.5}\scorestd{6.0} & \textbf{0.82}\scorestd{0.07} & \textbf{12.3}\scorestd{6.0} & \textbf{28.0}\scorestd{16.1} & \textbf{0.94}\scorestd{0.15} & \textbf{0.5}\scorestd{0.7} & \textbf{0.5}\scorestd{1.0} & \textbf{0.94}\scorestd{0.06} & \textbf{1.8}\scorestd{1.9} & \textbf{1.9}\scorestd{2.9} & \textbf{0.85}\scorestd{0.07} & \textbf{11.8}\scorestd{7.0} & \textbf{20.3}\scorestd{14.1} \\
\bottomrule
\end{tabular}
}
\end{table*}

%% file: figs/tables/main/appendix/table_detailed_gp-simple.tex
\begin{table*}[htbp]
\centering
\small
\caption{Detailed results for \texttt{gp\_simple} across different graph sizes ($d$) and metrics.}
\label{tab:detail-gp-simple}
\setlength{\tabcolsep}{4pt}
\resizebox{\textwidth}{!}{
\begin{tabular}{lccccccccc@{\hspace{6pt}}ccccccccc}
\toprule
& \multicolumn{9}{c}{\textbf{Observation-only}} & \multicolumn{9}{c}{\textbf{Mixed-interventional}} \\
\cmidrule(lr){2-10}\cmidrule(lr){11-19}
\textbf{Method} & \multicolumn{3}{c}{$d=5$} & \multicolumn{3}{c}{$d=10$} & \multicolumn{3}{c}{$d=20$} & \multicolumn{3}{c}{$d=5$} & \multicolumn{3}{c}{$d=10$} & \multicolumn{3}{c}{$d=20$} \\
\cmidrule(lr){2-10}\cmidrule(lr){11-19}
 & F1 & SHD & SID & F1 & SHD & SID & F1 & SHD & SID & F1 & SHD & SID & F1 & SHD & SID & F1 & SHD & SID \\
\midrule
RandomRegress & 0.34\scorestd{0.21} & 6.7\scorestd{2.4} & 8.7\scorestd{3.5} & 0.26\scorestd{0.10} & 33.7\scorestd{9.1} & 41.8\scorestd{10.3} & 0.18\scorestd{0.06} & 151.3\scorestd{44.3} & 178.8\scorestd{36.9} & -- & -- & -- & -- & -- & -- & -- & -- & -- \\
DAS & \textbf{0.84}\scorestd{0.19} & \underline{1.2}\scorestd{1.2} & 1.6\scorestd{2.1} & 0.72\scorestd{0.11} & 8.7\scorestd{4.5} & 10.4\scorestd{5.7} & 0.61\scorestd{0.09} & 34.5\scorestd{13.3} & 46.3\scorestd{16.0} & -- & -- & -- & -- & -- & -- & -- & -- & -- \\
LiNGAM & 0.26\scorestd{0.26} & 4.2\scorestd{2.1} & 5.8\scorestd{2.9} & 0.30\scorestd{0.15} & 14.2\scorestd{4.3} & 23.1\scorestd{6.9} & 0.32\scorestd{0.10} & 40.3\scorestd{9.6} & 78.2\scorestd{21.0} & -- & -- & -- & -- & -- & -- & -- & -- & -- \\
PC & 0.52\scorestd{0.29} & 3.0\scorestd{2.2} & 5.2\scorestd{3.9} & 0.46\scorestd{0.17} & 11.3\scorestd{5.1} & 23.6\scorestd{10.3} & 0.42\scorestd{0.13} & 33.4\scorestd{10.8} & 91.2\scorestd{31.9} & -- & -- & -- & -- & -- & -- & -- & -- & -- \\
CDIS & 0.40\scorestd{0.32} & 3.5\scorestd{2.0} & 5.3\scorestd{3.8} & 0.44\scorestd{0.15} & 11.9\scorestd{4.5} & 25.9\scorestd{11.6} & 0.38\scorestd{0.11} & 35.5\scorestd{10.5} & 105.3\scorestd{32.0} & 0.57\scorestd{0.28} & 2.8\scorestd{1.8} & 4.9\scorestd{3.9} & 0.50\scorestd{0.17} & 10.8\scorestd{5.0} & 22.0\scorestd{11.6} & 0.40\scorestd{0.14} & 34.4\scorestd{12.7} & 97.4\scorestd{33.6} \\
GIES & 0.57\scorestd{0.29} & 2.8\scorestd{2.2} & 4.7\scorestd{3.6} & 0.56\scorestd{0.18} & 12.6\scorestd{7.2} & 22.7\scorestd{14.2} & 0.45\scorestd{0.15} & 51.7\scorestd{21.8} & 107.4\scorestd{46.4} & 0.64\scorestd{0.23} & 3.4\scorestd{2.3} & 4.2\scorestd{3.7} & 0.48\scorestd{0.15} & 17.6\scorestd{6.7} & 25.4\scorestd{11.7} & 0.38\scorestd{0.10} & 64.4\scorestd{19.5} & 123.1\scorestd{36.5} \\
IGSP & 0.58\scorestd{0.30} & 2.6\scorestd{2.0} & 4.5\scorestd{3.8} & 0.48\scorestd{0.17} & 12.9\scorestd{6.5} & 27.0\scorestd{13.8} & 0.31\scorestd{0.10} & 55.7\scorestd{17.8} & 140.1\scorestd{38.5} & 0.55\scorestd{0.28} & 2.8\scorestd{1.9} & 5.3\scorestd{3.9} & 0.47\scorestd{0.20} & 12.9\scorestd{6.3} & 27.6\scorestd{14.7} & 0.32\scorestd{0.11} & 52.7\scorestd{16.9} & 135.1\scorestd{40.7} \\
DAGMA & 0.70\scorestd{0.24} & 2.3\scorestd{1.6} & 2.5\scorestd{2.4} & 0.67\scorestd{0.11} & 11.6\scorestd{4.8} & 12.5\scorestd{5.9} & 0.61\scorestd{0.07} & 39.7\scorestd{10.8} & 49.1\scorestd{15.2} & -- & -- & -- & -- & -- & -- & -- & -- & -- \\
NOTEARS & 0.72\scorestd{0.23} & 1.9\scorestd{1.3} & 2.1\scorestd{1.7} & 0.69\scorestd{0.12} & 8.3\scorestd{4.3} & 10.7\scorestd{5.6} & 0.60\scorestd{0.09} & 31.2\scorestd{11.9} & 43.2\scorestd{16.9} & -- & -- & -- & -- & -- & -- & -- & -- & -- \\
NOTEARS-MLP & \underline{0.83}\scorestd{0.18} & \underline{1.2}\scorestd{1.0} & \underline{1.5}\scorestd{1.7} & 0.79\scorestd{0.08} & 5.9\scorestd{3.1} & 7.6\scorestd{4.4} & 0.70\scorestd{0.08} & 30.6\scorestd{16.2} & 31.4\scorestd{11.9} & -- & -- & -- & -- & -- & -- & -- & -- & -- \\
NoDAGS & -- & -- & -- & -- & -- & -- & -- & -- & -- & \underline{0.72}\scorestd{0.19} & \underline{2.4}\scorestd{1.6} & \underline{2.5}\scorestd{2.2} & 0.58\scorestd{0.13} & 10.0\scorestd{4.2} & 15.3\scorestd{7.9} & 0.33\scorestd{0.14} & 34.2\scorestd{10.0} & 77.3\scorestd{25.1} \\
SEA & 0.72\scorestd{0.20} & 2.0\scorestd{1.6} & 2.5\scorestd{2.2} & 0.70\scorestd{0.13} & 8.2\scorestd{4.7} & 10.8\scorestd{6.0} & 0.53\scorestd{0.13} & 27.5\scorestd{8.8} & 59.9\scorestd{22.4} & 0.58\scorestd{0.21} & 3.4\scorestd{1.9} & 4.4\scorestd{2.8} & 0.53\scorestd{0.18} & 15.4\scorestd{6.6} & 15.4\scorestd{7.0} & 0.41\scorestd{0.12} & 39.8\scorestd{13.8} & 65.1\scorestd{24.6} \\
SDCD & 0.50\scorestd{0.32} & 2.9\scorestd{2.1} & 5.2\scorestd{4.0} & 0.55\scorestd{0.17} & 12.1\scorestd{5.3} & 22.4\scorestd{10.9} & 0.53\scorestd{0.09} & 42.9\scorestd{13.1} & 81.0\scorestd{22.0} & 0.70\scorestd{0.18} & \underline{2.4}\scorestd{1.5} & 3.2\scorestd{2.6} & 0.58\scorestd{0.16} & 12.9\scorestd{6.6} & 19.8\scorestd{11.7} & 0.56\scorestd{0.08} & 42.6\scorestd{14.0} & 61.7\scorestd{22.8} \\
DCDI & 0.36\scorestd{0.14} & 10.0\scorestd{0.0} & 14.7\scorestd{2.5} & 0.29\scorestd{0.06} & 45.0\scorestd{0.0} & 66.1\scorestd{7.0} & 0.19\scorestd{0.03} & 189.8\scorestd{1.4} & 295.3\scorestd{19.6} & 0.38\scorestd{0.10} & 10.0\scorestd{0.0} & 13.9\scorestd{2.0} & 0.29\scorestd{0.06} & 45.0\scorestd{0.0} & 66.6\scorestd{7.8} & 0.19\scorestd{0.04} & 190.0\scorestd{0.0} & 298.3\scorestd{24.8} \\
AVICI & 0.81\scorestd{0.24} & \textbf{1.1}\scorestd{1.1} & \textbf{1.4}\scorestd{2.0} & \underline{0.84}\scorestd{0.10} & \underline{4.2}\scorestd{2.5} & \underline{4.8}\scorestd{5.0} & \underline{0.80}\scorestd{0.05} & \underline{13.5}\scorestd{4.9} & \underline{18.8}\scorestd{9.7} & \textbf{0.97}\scorestd{0.05} & \textbf{0.3}\scorestd{0.5} & \textbf{0.2}\scorestd{0.6} & \underline{0.95}\scorestd{0.05} & \underline{1.5}\scorestd{1.7} & \underline{0.5}\scorestd{1.2} & \underline{0.90}\scorestd{0.05} & \underline{7.9}\scorestd{4.7} & \underline{4.5}\scorestd{4.5} \\
\method{} & \underline{0.83}\scorestd{0.19} & 1.3\scorestd{1.3} & 1.8\scorestd{2.5} & \textbf{0.87}\scorestd{0.10} & \textbf{3.3}\scorestd{2.4} & \textbf{4.5}\scorestd{5.8} & \textbf{0.87}\scorestd{0.05} & \textbf{9.4}\scorestd{4.5} & \textbf{13.9}\scorestd{10.4} & \textbf{0.97}\scorestd{0.05} & \textbf{0.3}\scorestd{0.5} & \textbf{0.2}\scorestd{0.6} & \textbf{0.96}\scorestd{0.04} & \textbf{1.4}\scorestd{1.3} & \textbf{0.4}\scorestd{1.1} & \textbf{0.92}\scorestd{0.05} & \textbf{6.6}\scorestd{4.9} & \textbf{2.8}\scorestd{3.6} \\
\bottomrule
\end{tabular}
}
\end{table*}

%% file: figs/tables/main/appendix/table_detailed_linear-gauss.tex
\begin{table*}[htbp]
\centering
\small
\caption{Detailed results for \texttt{linear\_gauss} across different graph sizes ($d$) and metrics.}
\label{tab:detail-linear-gauss}
\setlength{\tabcolsep}{4pt}
\resizebox{\textwidth}{!}{
\begin{tabular}{lccccccccc@{\hspace{6pt}}ccccccccc}
\toprule
& \multicolumn{9}{c}{\textbf{Observation-only}} & \multicolumn{9}{c}{\textbf{Mixed-interventional}} \\
\cmidrule(lr){2-10}\cmidrule(lr){11-19}
\textbf{Method} & \multicolumn{3}{c}{$d=5$} & \multicolumn{3}{c}{$d=10$} & \multicolumn{3}{c}{$d=20$} & \multicolumn{3}{c}{$d=5$} & \multicolumn{3}{c}{$d=10$} & \multicolumn{3}{c}{$d=20$} \\
\cmidrule(lr){2-10}\cmidrule(lr){11-19}
 & F1 & SHD & SID & F1 & SHD & SID & F1 & SHD & SID & F1 & SHD & SID & F1 & SHD & SID & F1 & SHD & SID \\
\midrule
RandomRegress & 0.35\scorestd{0.23} & 7.0\scorestd{2.2} & 9.3\scorestd{3.6} & 0.25\scorestd{0.12} & 34.6\scorestd{7.8} & 43.8\scorestd{10.6} & 0.18\scorestd{0.05} & 152.4\scorestd{41.5} & 182.0\scorestd{31.8} & -- & -- & -- & -- & -- & -- & -- & -- & -- \\
DAS & 0.86\scorestd{0.17} & 1.1\scorestd{1.4} & 2.1\scorestd{2.9} & 0.79\scorestd{0.14} & 6.1\scorestd{4.5} & 10.7\scorestd{8.7} & 0.64\scorestd{0.16} & 29.5\scorestd{18.6} & 52.8\scorestd{23.7} & -- & -- & -- & -- & -- & -- & -- & -- & -- \\
LiNGAM & 0.47\scorestd{0.30} & 3.6\scorestd{2.1} & 5.1\scorestd{3.3} & 0.46\scorestd{0.19} & 13.0\scorestd{5.6} & 19.6\scorestd{9.1} & 0.42\scorestd{0.12} & 40.7\scorestd{12.6} & 75.2\scorestd{22.9} & -- & -- & -- & -- & -- & -- & -- & -- & -- \\
PC & 0.67\scorestd{0.29} & 1.8\scorestd{1.6} & 4.0\scorestd{3.7} & 0.54\scorestd{0.18} & 9.5\scorestd{5.5} & 20.9\scorestd{9.8} & 0.45\scorestd{0.15} & 29.6\scorestd{12.2} & 87.8\scorestd{35.3} & -- & -- & -- & -- & -- & -- & -- & -- & -- \\
CDIS & 0.56\scorestd{0.34} & 2.5\scorestd{1.6} & 3.9\scorestd{3.0} & 0.49\scorestd{0.19} & 10.6\scorestd{5.6} & 23.3\scorestd{12.7} & 0.42\scorestd{0.14} & 32.2\scorestd{12.1} & 92.5\scorestd{32.4} & 0.71\scorestd{0.28} & 1.8\scorestd{1.6} & 3.3\scorestd{3.3} & 0.52\scorestd{0.18} & 10.2\scorestd{5.2} & 23.6\scorestd{12.3} & 0.41\scorestd{0.15} & 33.0\scorestd{12.5} & 93.7\scorestd{33.6} \\
GIES & 0.69\scorestd{0.30} & 1.7\scorestd{1.9} & 4.2\scorestd{4.2} & 0.68\scorestd{0.22} & 8.4\scorestd{7.9} & 18.6\scorestd{14.6} & 0.61\scorestd{0.21} & 35.7\scorestd{27.4} & 89.2\scorestd{53.3} & \underline{0.98}\scorestd{0.08} & \underline{0.2}\scorestd{0.7} & \underline{0.3}\scorestd{1.1} & \underline{0.94}\scorestd{0.11} & \underline{1.8}\scorestd{3.5} & 3.5\scorestd{7.1} & \underline{0.88}\scorestd{0.14} & \underline{10.6}\scorestd{15.2} & 27.3\scorestd{31.2} \\
IGSP & 0.75\scorestd{0.26} & 1.5\scorestd{1.7} & 3.8\scorestd{4.0} & 0.72\scorestd{0.21} & 6.4\scorestd{6.3} & 16.7\scorestd{14.0} & 0.64\scorestd{0.15} & 27.9\scorestd{17.3} & 78.4\scorestd{41.5} & 0.71\scorestd{0.26} & 1.6\scorestd{1.8} & 3.9\scorestd{3.8} & 0.71\scorestd{0.19} & 6.9\scorestd{6.0} & 17.3\scorestd{12.9} & 0.58\scorestd{0.14} & 33.4\scorestd{17.5} & 93.2\scorestd{40.4} \\
DAGMA & \textbf{0.94}\scorestd{0.13} & \textbf{0.4}\scorestd{1.0} & \textbf{0.8}\scorestd{1.9} & \textbf{0.92}\scorestd{0.10} & \textbf{1.9}\scorestd{2.7} & \textbf{3.4}\scorestd{4.2} & \textbf{0.93}\scorestd{0.07} & \textbf{4.2}\scorestd{4.4} & \textbf{10.9}\scorestd{13.4} & -- & -- & -- & -- & -- & -- & -- & -- & -- \\
NOTEARS & \underline{0.87}\scorestd{0.15} & \underline{1.0}\scorestd{1.2} & \underline{1.1}\scorestd{1.7} & \underline{0.85}\scorestd{0.10} & \underline{3.9}\scorestd{3.2} & \underline{5.3}\scorestd{4.8} & \underline{0.85}\scorestd{0.08} & \underline{10.6}\scorestd{6.2} & \underline{19.3}\scorestd{13.8} & -- & -- & -- & -- & -- & -- & -- & -- & -- \\
NOTEARS-MLP & 0.85\scorestd{0.17} & 1.3\scorestd{1.5} & 1.8\scorestd{2.2} & 0.67\scorestd{0.17} & 7.9\scorestd{4.5} & 13.1\scorestd{7.4} & 0.58\scorestd{0.14} & 25.3\scorestd{10.0} & 61.9\scorestd{25.9} & -- & -- & -- & -- & -- & -- & -- & -- & -- \\
NoDAGS & -- & -- & -- & -- & -- & -- & -- & -- & -- & 0.95\scorestd{0.09} & 0.6\scorestd{1.0} & 0.4\scorestd{1.0} & 0.69\scorestd{0.19} & 7.9\scorestd{4.6} & 10.2\scorestd{8.4} & 0.19\scorestd{0.22} & 37.3\scorestd{11.8} & 84.0\scorestd{29.5} \\
SEA & 0.77\scorestd{0.20} & 1.8\scorestd{1.8} & 3.7\scorestd{3.4} & 0.81\scorestd{0.12} & 4.8\scorestd{3.7} & 9.8\scorestd{8.1} & 0.80\scorestd{0.09} & 14.6\scorestd{9.5} & 37.7\scorestd{21.5} & 0.76\scorestd{0.20} & 1.4\scorestd{1.0} & 3.3\scorestd{2.7} & 0.76\scorestd{0.10} & 7.7\scorestd{4.9} & 9.9\scorestd{5.6} & 0.70\scorestd{0.09} & 27.3\scorestd{13.9} & 47.7\scorestd{19.6} \\
SDCD & 0.55\scorestd{0.29} & 2.9\scorestd{2.3} & 6.4\scorestd{4.6} & 0.44\scorestd{0.15} & 12.6\scorestd{5.5} & 31.4\scorestd{13.1} & 0.37\scorestd{0.11} & 42.9\scorestd{15.0} & 143.8\scorestd{47.0} & 0.76\scorestd{0.26} & 1.4\scorestd{1.7} & 3.2\scorestd{3.6} & 0.78\scorestd{0.20} & 5.5\scorestd{5.5} & 11.4\scorestd{13.2} & 0.58\scorestd{0.11} & 28.6\scorestd{9.2} & 88.4\scorestd{32.8} \\
DCDI & 0.39\scorestd{0.11} & 10.0\scorestd{0.0} & 13.7\scorestd{2.4} & 0.29\scorestd{0.08} & 45.0\scorestd{0.0} & 67.0\scorestd{7.8} & 0.19\scorestd{0.04} & 190.0\scorestd{0.1} & 294.7\scorestd{24.0} & 0.38\scorestd{0.11} & 10.0\scorestd{0.0} & 14.1\scorestd{2.1} & 0.29\scorestd{0.07} & 45.0\scorestd{0.0} & 65.5\scorestd{7.4} & 0.19\scorestd{0.04} & 190.0\scorestd{0.0} & 292.8\scorestd{24.0} \\
AVICI & 0.58\scorestd{0.28} & 2.8\scorestd{2.1} & 4.7\scorestd{3.5} & 0.62\scorestd{0.16} & 9.5\scorestd{5.4} & 14.0\scorestd{8.3} & 0.56\scorestd{0.15} & 27.7\scorestd{12.2} & 53.3\scorestd{25.6} & \underline{0.98}\scorestd{0.04} & \underline{0.2}\scorestd{0.5} & \textbf{0.0}\scorestd{0.1} & 0.93\scorestd{0.06} & 2.4\scorestd{2.3} & \underline{0.3}\scorestd{0.8} & 0.84\scorestd{0.06} & 12.8\scorestd{7.0} & \underline{4.5}\scorestd{3.9} \\
\method{} & 0.69\scorestd{0.29} & 2.1\scorestd{2.1} & 3.9\scorestd{3.5} & 0.70\scorestd{0.18} & 7.7\scorestd{5.7} & 13.0\scorestd{9.0} & 0.78\scorestd{0.12} & 16.2\scorestd{11.2} & 29.9\scorestd{19.7} & \textbf{1.00}\scorestd{0.01} & \textbf{0.0}\scorestd{0.1} & \textbf{0.0}\scorestd{0.0} & \textbf{0.99}\scorestd{0.02} & \textbf{0.4}\scorestd{0.8} & \textbf{0.0}\scorestd{0.3} & \textbf{0.97}\scorestd{0.03} & \textbf{3.0}\scorestd{2.7} & \textbf{0.5}\scorestd{1.8} \\
\bottomrule
\end{tabular}
}
\end{table*}

%% file: figs/tables/main/appendix/table_detailed_linear-graph.tex
\begin{table*}[htbp]
\centering
\small
\caption{Detailed results for \texttt{linear\_graph} across different graph sizes ($d$) and metrics.}
\label{tab:detail-linear-graph}
\setlength{\tabcolsep}{4pt}
\resizebox{\textwidth}{!}{
\begin{tabular}{lccccccccc@{\hspace{6pt}}ccccccccc}
\toprule
& \multicolumn{9}{c}{\textbf{Observation-only}} & \multicolumn{9}{c}{\textbf{Mixed-interventional}} \\
\cmidrule(lr){2-10}\cmidrule(lr){11-19}
\textbf{Method} & \multicolumn{3}{c}{$d=5$} & \multicolumn{3}{c}{$d=10$} & \multicolumn{3}{c}{$d=20$} & \multicolumn{3}{c}{$d=5$} & \multicolumn{3}{c}{$d=10$} & \multicolumn{3}{c}{$d=20$} \\
\cmidrule(lr){2-10}\cmidrule(lr){11-19}
 & F1 & SHD & SID & F1 & SHD & SID & F1 & SHD & SID & F1 & SHD & SID & F1 & SHD & SID & F1 & SHD & SID \\
\midrule
RandomRegress & 0.26\scorestd{0.26} & 7.2\scorestd{3.2} & 8.4\scorestd{4.0} & 0.22\scorestd{0.12} & 34.8\scorestd{10.7} & 41.4\scorestd{11.2} & 0.17\scorestd{0.07} & 153.2\scorestd{41.9} & 178.8\scorestd{36.1} & -- & -- & -- & -- & -- & -- & -- & -- & -- \\
DAS & 0.76\scorestd{0.28} & 1.0\scorestd{1.3} & 1.8\scorestd{2.7} & 0.78\scorestd{0.15} & 4.8\scorestd{4.1} & 8.2\scorestd{7.3} & 0.64\scorestd{0.14} & 26.4\scorestd{13.0} & 47.6\scorestd{21.8} & -- & -- & -- & -- & -- & -- & -- & -- & -- \\
LiNGAM & 0.44\scorestd{0.39} & 2.0\scorestd{1.9} & 2.9\scorestd{3.1} & 0.46\scorestd{0.22} & 9.8\scorestd{6.2} & 14.1\scorestd{8.5} & 0.42\scorestd{0.12} & 38.8\scorestd{10.0} & 67.1\scorestd{20.7} & -- & -- & -- & -- & -- & -- & -- & -- & -- \\
PC & 0.56\scorestd{0.39} & 1.2\scorestd{1.1} & 2.5\scorestd{2.4} & 0.60\scorestd{0.21} & 5.9\scorestd{4.1} & 13.2\scorestd{9.4} & 0.45\scorestd{0.13} & 27.6\scorestd{8.3} & 73.9\scorestd{26.2} & -- & -- & -- & -- & -- & -- & -- & -- & -- \\
CDIS & 0.30\scorestd{0.41} & 1.7\scorestd{1.4} & 2.3\scorestd{2.2} & 0.52\scorestd{0.26} & 6.8\scorestd{3.9} & 12.8\scorestd{9.0} & 0.44\scorestd{0.13} & 29.1\scorestd{8.8} & 72.9\scorestd{28.6} & 0.65\scorestd{0.42} & 1.0\scorestd{1.2} & 1.3\scorestd{2.0} & 0.62\scorestd{0.25} & 6.2\scorestd{4.7} & 11.3\scorestd{9.5} & 0.45\scorestd{0.13} & 28.5\scorestd{9.5} & 71.0\scorestd{26.2} \\
GIES & 0.59\scorestd{0.40} & 1.0\scorestd{1.2} & 2.2\scorestd{2.7} & 0.73\scorestd{0.21} & 4.5\scorestd{5.3} & 10.0\scorestd{9.9} & 0.67\scorestd{0.19} & 25.2\scorestd{19.8} & 59.3\scorestd{45.3} & 0.91\scorestd{0.23} & 0.2\scorestd{0.4} & \underline{0.2}\scorestd{0.5} & \underline{0.96}\scorestd{0.09} & \underline{0.8}\scorestd{1.6} & 2.1\scorestd{5.2} & \underline{0.89}\scorestd{0.13} & \underline{8.7}\scorestd{12.0} & 24.2\scorestd{29.9} \\
IGSP & 0.57\scorestd{0.39} & 1.0\scorestd{1.2} & 2.4\scorestd{2.7} & 0.68\scorestd{0.23} & 5.2\scorestd{5.5} & 11.5\scorestd{11.0} & 0.52\scorestd{0.20} & 38.3\scorestd{23.2} & 98.1\scorestd{46.8} & 0.65\scorestd{0.40} & 0.8\scorestd{1.1} & 2.0\scorestd{2.7} & 0.70\scorestd{0.21} & 5.1\scorestd{5.5} & 11.5\scorestd{10.7} & 0.53\scorestd{0.19} & 35.6\scorestd{21.1} & 91.9\scorestd{43.1} \\
DAGMA & \textbf{0.89}\scorestd{0.27} & \textbf{0.2}\scorestd{0.8} & \textbf{0.3}\scorestd{0.9} & \textbf{0.94}\scorestd{0.10} & \textbf{1.1}\scorestd{2.2} & \textbf{1.8}\scorestd{3.8} & \textbf{0.95}\scorestd{0.07} & \textbf{3.4}\scorestd{5.4} & \textbf{7.9}\scorestd{11.4} & -- & -- & -- & -- & -- & -- & -- & -- & -- \\
NOTEARS & 0.85\scorestd{0.30} & \underline{0.3}\scorestd{0.6} & \underline{0.5}\scorestd{1.2} & \underline{0.87}\scorestd{0.12} & \underline{2.8}\scorestd{3.2} & \underline{3.0}\scorestd{3.7} & \underline{0.86}\scorestd{0.08} & \underline{9.4}\scorestd{6.0} & \underline{15.1}\scorestd{12.1} & -- & -- & -- & -- & -- & -- & -- & -- & -- \\
NOTEARS-MLP & \underline{0.86}\scorestd{0.29} & \underline{0.3}\scorestd{0.8} & 0.6\scorestd{1.3} & 0.73\scorestd{0.18} & 5.7\scorestd{4.6} & 9.2\scorestd{7.5} & 0.55\scorestd{0.12} & 25.1\scorestd{7.4} & 54.1\scorestd{18.6} & -- & -- & -- & -- & -- & -- & -- & -- & -- \\
NoDAGS & -- & -- & -- & -- & -- & -- & -- & -- & -- & 0.91\scorestd{0.22} & 0.3\scorestd{0.6} & 0.3\scorestd{0.6} & 0.79\scorestd{0.17} & 4.8\scorestd{4.4} & 5.1\scorestd{5.7} & 0.26\scorestd{0.22} & 33.0\scorestd{9.2} & 62.9\scorestd{24.8} \\
SEA & 0.71\scorestd{0.35} & 1.2\scorestd{1.6} & 2.8\scorestd{3.6} & 0.84\scorestd{0.13} & 3.0\scorestd{3.1} & 4.9\scorestd{5.0} & 0.76\scorestd{0.09} & 15.9\scorestd{7.0} & 31.7\scorestd{16.9} & 0.69\scorestd{0.32} & 1.6\scorestd{1.9} & 3.4\scorestd{3.7} & 0.79\scorestd{0.13} & 4.7\scorestd{4.3} & 8.2\scorestd{7.4} & 0.65\scorestd{0.13} & 28.7\scorestd{15.1} & 51.8\scorestd{24.9} \\
SDCD & 0.41\scorestd{0.40} & 1.6\scorestd{1.6} & 3.6\scorestd{3.7} & 0.48\scorestd{0.25} & 8.8\scorestd{7.1} & 19.8\scorestd{15.4} & 0.29\scorestd{0.11} & 46.6\scorestd{14.1} & 133.5\scorestd{36.7} & 0.79\scorestd{0.29} & 1.0\scorestd{1.6} & 2.1\scorestd{3.1} & 0.79\scorestd{0.24} & 2.9\scorestd{3.3} & 6.2\scorestd{7.6} & 0.49\scorestd{0.12} & 33.0\scorestd{10.4} & 93.1\scorestd{31.1} \\
DCDI & 0.23\scorestd{0.13} & 10.0\scorestd{0.0} & 16.6\scorestd{2.6} & 0.21\scorestd{0.09} & 45.0\scorestd{0.0} & 74.8\scorestd{8.3} & 0.18\scorestd{0.03} & 189.9\scorestd{0.6} & 307.6\scorestd{18.4} & 0.27\scorestd{0.12} & 10.0\scorestd{0.0} & 16.1\scorestd{2.3} & 0.22\scorestd{0.09} & 45.0\scorestd{0.0} & 74.0\scorestd{8.3} & 0.18\scorestd{0.03} & 190.0\scorestd{0.0} & 307.2\scorestd{17.0} \\
AVICI & 0.49\scorestd{0.42} & 1.1\scorestd{1.1} & 1.8\scorestd{2.3} & 0.64\scorestd{0.24} & 6.0\scorestd{4.4} & 8.0\scorestd{7.1} & 0.59\scorestd{0.13} & 25.4\scorestd{8.5} & 40.1\scorestd{23.6} & \underline{0.94}\scorestd{0.22} & \underline{0.1}\scorestd{0.4} & \textbf{0.0}\scorestd{0.2} & \underline{0.96}\scorestd{0.05} & 1.2\scorestd{1.8} & \underline{0.1}\scorestd{0.5} & 0.85\scorestd{0.06} & 11.6\scorestd{6.1} & \underline{3.2}\scorestd{3.6} \\
\method{} & 0.58\scorestd{0.40} & 1.1\scorestd{1.3} & 2.0\scorestd{2.4} & 0.71\scorestd{0.23} & 4.7\scorestd{4.4} & 7.5\scorestd{7.1} & 0.77\scorestd{0.11} & 15.1\scorestd{8.4} & 22.8\scorestd{17.2} & \textbf{0.95}\scorestd{0.21} & \textbf{0.0}\scorestd{0.1} & \textbf{0.0}\scorestd{0.0} & \textbf{0.99}\scorestd{0.02} & \textbf{0.2}\scorestd{0.5} & \textbf{0.0}\scorestd{0.1} & \textbf{0.97}\scorestd{0.03} & \textbf{2.5}\scorestd{2.6} & \textbf{0.6}\scorestd{2.3} \\
\bottomrule
\end{tabular}
}
\end{table*}

%% file: figs/tables/main/appendix/table_detailed_linear-nongauss.tex
\begin{table*}[htbp]
\centering
\small
\caption{Detailed results for \texttt{linear\_nongauss} across different graph sizes ($d$) and metrics.}
\label{tab:detail-linear-nongauss}
\setlength{\tabcolsep}{4pt}
\resizebox{\textwidth}{!}{
\begin{tabular}{lccccccccc@{\hspace{6pt}}ccccccccc}
\toprule
& \multicolumn{9}{c}{\textbf{Observation-only}} & \multicolumn{9}{c}{\textbf{Mixed-interventional}} \\
\cmidrule(lr){2-10}\cmidrule(lr){11-19}
\textbf{Method} & \multicolumn{3}{c}{$d=5$} & \multicolumn{3}{c}{$d=10$} & \multicolumn{3}{c}{$d=20$} & \multicolumn{3}{c}{$d=5$} & \multicolumn{3}{c}{$d=10$} & \multicolumn{3}{c}{$d=20$} \\
\cmidrule(lr){2-10}\cmidrule(lr){11-19}
 & F1 & SHD & SID & F1 & SHD & SID & F1 & SHD & SID & F1 & SHD & SID & F1 & SHD & SID & F1 & SHD & SID \\
\midrule
RandomRegress & 0.32\scorestd{0.19} & 7.1\scorestd{2.0} & 10.0\scorestd{3.3} & 0.28\scorestd{0.10} & 34.0\scorestd{7.2} & 43.3\scorestd{7.9} & 0.18\scorestd{0.06} & 152.8\scorestd{38.7} & 181.8\scorestd{28.2} & -- & -- & -- & -- & -- & -- & -- & -- & -- \\
DAS & \underline{0.94}\scorestd{0.12} & \underline{0.5}\scorestd{1.0} & \underline{0.8}\scorestd{1.8} & 0.84\scorestd{0.11} & 5.0\scorestd{3.9} & 7.4\scorestd{6.4} & 0.69\scorestd{0.14} & 26.5\scorestd{16.6} & 43.9\scorestd{21.7} & -- & -- & -- & -- & -- & -- & -- & -- & -- \\
LiNGAM & \textbf{0.99}\scorestd{0.03} & \textbf{0.1}\scorestd{0.2} & \textbf{0.0}\scorestd{0.2} & \textbf{0.99}\scorestd{0.02} & \textbf{0.2}\scorestd{0.5} & \textbf{0.2}\scorestd{0.6} & \textbf{0.99}\scorestd{0.03} & \textbf{1.1}\scorestd{2.8} & \textbf{1.5}\scorestd{4.6} & -- & -- & -- & -- & -- & -- & -- & -- & -- \\
PC & 0.64\scorestd{0.27} & 2.1\scorestd{1.7} & 4.8\scorestd{3.8} & 0.54\scorestd{0.17} & 9.8\scorestd{5.2} & 22.0\scorestd{10.8} & 0.43\scorestd{0.14} & 30.9\scorestd{11.9} & 90.1\scorestd{32.0} & -- & -- & -- & -- & -- & -- & -- & -- & -- \\
CDIS & 0.47\scorestd{0.35} & 3.1\scorestd{1.9} & 4.6\scorestd{3.3} & 0.52\scorestd{0.18} & 10.6\scorestd{5.1} & 21.9\scorestd{11.6} & 0.41\scorestd{0.12} & 33.0\scorestd{11.9} & 98.8\scorestd{33.0} & 0.66\scorestd{0.32} & 2.0\scorestd{1.9} & 3.7\scorestd{3.7} & 0.58\scorestd{0.19} & 9.0\scorestd{5.4} & 19.8\scorestd{11.2} & 0.42\scorestd{0.14} & 32.3\scorestd{12.6} & 95.1\scorestd{33.6} \\
GIES & 0.69\scorestd{0.29} & 1.9\scorestd{2.2} & 4.8\scorestd{4.7} & 0.67\scorestd{0.24} & 8.7\scorestd{7.9} & 19.6\scorestd{15.3} & 0.58\scorestd{0.20} & 39.9\scorestd{28.1} & 94.3\scorestd{52.7} & \underline{0.98}\scorestd{0.06} & \underline{0.1}\scorestd{0.4} & 0.2\scorestd{0.9} & \underline{0.95}\scorestd{0.10} & \underline{1.3}\scorestd{2.9} & 2.5\scorestd{5.8} & \underline{0.91}\scorestd{0.11} & \underline{7.8}\scorestd{12.5} & 22.1\scorestd{27.2} \\
IGSP & 0.72\scorestd{0.26} & 1.7\scorestd{1.9} & 4.2\scorestd{4.4} & 0.73\scorestd{0.21} & 6.3\scorestd{6.0} & 16.5\scorestd{14.4} & 0.63\scorestd{0.15} & 29.7\scorestd{18.1} & 82.4\scorestd{42.0} & 0.71\scorestd{0.28} & 1.6\scorestd{1.7} & 3.7\scorestd{3.9} & 0.73\scorestd{0.18} & 6.0\scorestd{4.9} & 15.3\scorestd{11.4} & 0.62\scorestd{0.15} & 30.2\scorestd{16.5} & 83.5\scorestd{41.2} \\
DAGMA & 0.92\scorestd{0.13} & 0.6\scorestd{1.1} & 1.0\scorestd{1.8} & 0.92\scorestd{0.09} & 2.0\scorestd{2.4} & 3.2\scorestd{4.1} & \underline{0.94}\scorestd{0.05} & \underline{4.1}\scorestd{3.7} & 11.1\scorestd{9.8} & -- & -- & -- & -- & -- & -- & -- & -- & -- \\
NOTEARS & 0.85\scorestd{0.17} & 1.2\scorestd{1.5} & 1.4\scorestd{1.9} & 0.83\scorestd{0.10} & 4.3\scorestd{2.9} & 6.0\scorestd{5.3} & 0.85\scorestd{0.07} & 10.7\scorestd{5.8} & 18.6\scorestd{11.6} & -- & -- & -- & -- & -- & -- & -- & -- & -- \\
NOTEARS-MLP & 0.82\scorestd{0.18} & 1.5\scorestd{1.6} & 1.9\scorestd{2.3} & 0.67\scorestd{0.16} & 7.9\scorestd{4.3} & 13.5\scorestd{8.1} & 0.57\scorestd{0.13} & 26.6\scorestd{10.6} & 63.2\scorestd{22.7} & -- & -- & -- & -- & -- & -- & -- & -- & -- \\
NoDAGS & -- & -- & -- & -- & -- & -- & -- & -- & -- & 0.92\scorestd{0.10} & 0.9\scorestd{1.1} & 0.5\scorestd{1.1} & 0.71\scorestd{0.18} & 7.2\scorestd{4.5} & 8.6\scorestd{7.7} & 0.21\scorestd{0.23} & 36.2\scorestd{11.9} & 79.3\scorestd{28.5} \\
SEA & 0.74\scorestd{0.22} & 2.0\scorestd{1.8} & 4.0\scorestd{3.7} & 0.82\scorestd{0.10} & 4.8\scorestd{3.1} & 9.4\scorestd{7.0} & 0.80\scorestd{0.11} & 15.2\scorestd{10.0} & 39.1\scorestd{24.2} & 0.81\scorestd{0.19} & 1.1\scorestd{1.2} & 2.7\scorestd{2.9} & 0.77\scorestd{0.12} & 7.0\scorestd{5.4} & 10.0\scorestd{6.9} & 0.69\scorestd{0.09} & 29.0\scorestd{16.2} & 48.0\scorestd{19.7} \\
SDCD & 0.53\scorestd{0.25} & 3.3\scorestd{2.1} & 7.0\scorestd{4.4} & 0.41\scorestd{0.15} & 14.2\scorestd{5.4} & 35.4\scorestd{12.5} & 0.38\scorestd{0.09} & 44.4\scorestd{13.3} & 143.6\scorestd{36.6} & 0.73\scorestd{0.25} & 1.8\scorestd{1.9} & 3.8\scorestd{3.5} & 0.83\scorestd{0.13} & 4.0\scorestd{3.1} & 8.7\scorestd{8.8} & 0.56\scorestd{0.11} & 30.3\scorestd{10.2} & 94.4\scorestd{32.5} \\
DCDI & 0.42\scorestd{0.10} & 10.0\scorestd{0.0} & 13.6\scorestd{2.3} & 0.30\scorestd{0.06} & 45.0\scorestd{0.0} & 64.6\scorestd{6.6} & 0.19\scorestd{0.04} & 190.0\scorestd{0.0} & 294.7\scorestd{22.6} & 0.39\scorestd{0.10} & 10.0\scorestd{0.0} & 13.9\scorestd{2.2} & 0.28\scorestd{0.05} & 45.0\scorestd{0.0} & 66.4\scorestd{7.2} & 0.19\scorestd{0.04} & 190.0\scorestd{0.0} & 295.2\scorestd{25.2} \\
AVICI & 0.28\scorestd{0.23} & 4.6\scorestd{2.2} & 9.2\scorestd{4.4} & 0.41\scorestd{0.18} & 12.9\scorestd{5.4} & 25.9\scorestd{10.2} & 0.39\scorestd{0.12} & 33.9\scorestd{10.5} & 78.9\scorestd{23.6} & 0.96\scorestd{0.06} & 0.3\scorestd{0.6} & \underline{0.1}\scorestd{0.6} & 0.92\scorestd{0.07} & 2.7\scorestd{2.7} & \underline{1.0}\scorestd{2.3} & 0.85\scorestd{0.08} & 12.3\scorestd{8.1} & \underline{7.2}\scorestd{7.3} \\
\method{} & \textbf{0.99}\scorestd{0.02} & \textbf{0.1}\scorestd{0.3} & \textbf{0.0}\scorestd{0.2} & \underline{0.95}\scorestd{0.07} & \underline{1.8}\scorestd{3.1} & \underline{0.6}\scorestd{1.6} & 0.93\scorestd{0.07} & 6.4\scorestd{6.0} & \underline{3.3}\scorestd{6.5} & \textbf{1.00}\scorestd{0.01} & \textbf{0.0}\scorestd{0.2} & \textbf{0.0}\scorestd{0.0} & \textbf{0.99}\scorestd{0.02} & \textbf{0.5}\scorestd{0.8} & \textbf{0.0}\scorestd{0.0} & \textbf{0.97}\scorestd{0.03} & \textbf{2.7}\scorestd{2.7} & \textbf{0.1}\scorestd{0.5} \\
\bottomrule
\end{tabular}
}
\end{table*}

%% file: figs/tables/main/appendix/table_detailed_mul-noise.tex
\begin{table*}[htbp]
\centering
\small
\caption{Detailed results for \texttt{mul\_noise} across different graph sizes ($d$) and metrics.}
\label{tab:detail-mul-noise}
\setlength{\tabcolsep}{4pt}
\resizebox{\textwidth}{!}{
\begin{tabular}{lccccccccc@{\hspace{6pt}}ccccccccc}
\toprule
& \multicolumn{9}{c}{\textbf{Observation-only}} & \multicolumn{9}{c}{\textbf{Mixed-interventional}} \\
\cmidrule(lr){2-10}\cmidrule(lr){11-19}
\textbf{Method} & \multicolumn{3}{c}{$d=5$} & \multicolumn{3}{c}{$d=10$} & \multicolumn{3}{c}{$d=20$} & \multicolumn{3}{c}{$d=5$} & \multicolumn{3}{c}{$d=10$} & \multicolumn{3}{c}{$d=20$} \\
\cmidrule(lr){2-10}\cmidrule(lr){11-19}
 & F1 & SHD & SID & F1 & SHD & SID & F1 & SHD & SID & F1 & SHD & SID & F1 & SHD & SID & F1 & SHD & SID \\
\midrule
RandomRegress & 0.33\scorestd{0.20} & 6.6\scorestd{2.1} & 9.2\scorestd{3.5} & 0.26\scorestd{0.13} & 33.8\scorestd{9.5} & 42.3\scorestd{11.1} & 0.18\scorestd{0.06} & 150.2\scorestd{43.3} & 177.3\scorestd{31.8} & -- & -- & -- & -- & -- & -- & -- & -- & -- \\
DAS & \underline{0.82}\scorestd{0.19} & \underline{1.5}\scorestd{1.5} & \underline{2.0}\scorestd{2.6} & 0.74\scorestd{0.13} & 6.9\scorestd{3.8} & 10.2\scorestd{6.7} & 0.64\scorestd{0.09} & 27.5\scorestd{11.6} & 46.4\scorestd{18.0} & -- & -- & -- & -- & -- & -- & -- & -- & -- \\
LiNGAM & 0.58\scorestd{0.30} & 3.1\scorestd{2.2} & 3.9\scorestd{2.9} & 0.51\scorestd{0.22} & 11.1\scorestd{5.3} & 15.7\scorestd{8.5} & 0.46\scorestd{0.14} & 35.1\scorestd{10.8} & 61.7\scorestd{21.9} & -- & -- & -- & -- & -- & -- & -- & -- & -- \\
PC & 0.60\scorestd{0.26} & 2.3\scorestd{1.6} & 5.2\scorestd{4.0} & 0.53\scorestd{0.16} & 8.6\scorestd{3.7} & 21.8\scorestd{10.8} & 0.48\scorestd{0.11} & 27.1\scorestd{8.9} & 90.2\scorestd{35.0} & -- & -- & -- & -- & -- & -- & -- & -- & -- \\
CDIS & 0.49\scorestd{0.32} & 3.0\scorestd{1.6} & 4.9\scorestd{3.1} & 0.51\scorestd{0.18} & 9.2\scorestd{3.6} & 20.8\scorestd{10.7} & 0.46\scorestd{0.11} & 28.3\scorestd{8.9} & 92.7\scorestd{32.3} & 0.60\scorestd{0.28} & 2.5\scorestd{1.7} & 4.7\scorestd{4.0} & 0.54\scorestd{0.16} & 9.2\scorestd{4.2} & 22.4\scorestd{11.4} & 0.46\scorestd{0.12} & 28.7\scorestd{10.3} & 95.9\scorestd{33.2} \\
GIES & 0.65\scorestd{0.25} & 2.1\scorestd{1.9} & 5.0\scorestd{3.8} & 0.64\scorestd{0.18} & 8.0\scorestd{5.1} & 19.9\scorestd{12.9} & 0.65\scorestd{0.16} & 24.9\scorestd{15.0} & 82.6\scorestd{43.2} & \underline{0.96}\scorestd{0.11} & \underline{0.4}\scorestd{0.7} & \underline{0.2}\scorestd{0.9} & \underline{0.92}\scorestd{0.08} & \underline{2.7}\scorestd{2.7} & 2.0\scorestd{3.2} & \underline{0.83}\scorestd{0.10} & 16.0\scorestd{10.8} & 27.7\scorestd{20.4} \\
IGSP & 0.63\scorestd{0.26} & 2.2\scorestd{1.8} & 5.1\scorestd{3.9} & 0.61\scorestd{0.17} & 7.8\scorestd{4.5} & 20.2\scorestd{12.1} & 0.54\scorestd{0.13} & 28.5\scorestd{11.8} & 94.9\scorestd{36.7} & 0.51\scorestd{0.31} & 2.9\scorestd{2.1} & 6.3\scorestd{4.7} & 0.58\scorestd{0.19} & 8.7\scorestd{4.8} & 21.8\scorestd{12.6} & 0.49\scorestd{0.14} & 32.1\scorestd{13.5} & 105.7\scorestd{42.0} \\
DAGMA & 0.78\scorestd{0.24} & 1.6\scorestd{1.7} & 2.2\scorestd{2.3} & 0.76\scorestd{0.16} & 5.8\scorestd{4.2} & 9.1\scorestd{6.8} & 0.62\scorestd{0.17} & 25.6\scorestd{11.0} & 53.0\scorestd{27.0} & -- & -- & -- & -- & -- & -- & -- & -- & -- \\
NOTEARS & 0.16\scorestd{0.22} & 4.6\scorestd{1.6} & 5.8\scorestd{2.2} & 0.31\scorestd{0.19} & 11.9\scorestd{3.4} & 18.0\scorestd{5.7} & 0.43\scorestd{0.13} & 29.8\scorestd{6.7} & 64.3\scorestd{15.8} & -- & -- & -- & -- & -- & -- & -- & -- & -- \\
NOTEARS-MLP & 0.35\scorestd{0.34} & 3.8\scorestd{2.0} & 4.5\scorestd{2.7} & 0.47\scorestd{0.19} & 10.7\scorestd{3.8} & 15.7\scorestd{6.3} & 0.54\scorestd{0.11} & 31.1\scorestd{14.7} & 56.9\scorestd{18.4} & -- & -- & -- & -- & -- & -- & -- & -- & -- \\
NoDAGS & -- & -- & -- & -- & -- & -- & -- & -- & -- & 0.92\scorestd{0.12} & 0.8\scorestd{0.9} & 0.3\scorestd{0.6} & 0.78\scorestd{0.15} & 6.1\scorestd{4.3} & 6.2\scorestd{6.0} & 0.41\scorestd{0.27} & 31.0\scorestd{13.2} & 69.6\scorestd{34.5} \\
SEA & 0.70\scorestd{0.27} & 1.9\scorestd{1.4} & 3.5\scorestd{3.5} & \underline{0.77}\scorestd{0.11} & \underline{5.7}\scorestd{3.3} & \underline{8.3}\scorestd{5.4} & \underline{0.68}\scorestd{0.10} & \underline{24.2}\scorestd{9.0} & \underline{36.6}\scorestd{18.3} & 0.92\scorestd{0.12} & 0.7\scorestd{1.0} & 1.2\scorestd{1.6} & 0.78\scorestd{0.12} & 6.3\scorestd{4.2} & 8.7\scorestd{6.5} & 0.64\scorestd{0.09} & 35.9\scorestd{15.3} & 43.6\scorestd{15.6} \\
SDCD & 0.51\scorestd{0.30} & 3.0\scorestd{2.1} & 6.1\scorestd{4.2} & 0.68\scorestd{0.17} & 7.2\scorestd{4.1} & 15.0\scorestd{11.9} & 0.64\scorestd{0.11} & 24.5\scorestd{9.3} & 63.8\scorestd{24.9} & 0.85\scorestd{0.16} & 1.4\scorestd{1.4} & 1.9\scorestd{2.6} & 0.78\scorestd{0.13} & 6.4\scorestd{3.9} & 7.8\scorestd{8.5} & 0.72\scorestd{0.11} & 23.8\scorestd{11.1} & 27.6\scorestd{20.4} \\
DCDI & 0.39\scorestd{0.10} & 9.9\scorestd{0.4} & 13.8\scorestd{2.2} & 0.35\scorestd{0.08} & 36.4\scorestd{8.0} & 63.4\scorestd{9.5} & 0.25\scorestd{0.04} & 161.4\scorestd{10.3} & 288.5\scorestd{21.5} & 0.43\scorestd{0.11} & 10.0\scorestd{0.1} & 13.1\scorestd{2.2} & 0.29\scorestd{0.06} & 45.0\scorestd{0.0} & 65.1\scorestd{6.4} & 0.22\scorestd{0.03} & 181.7\scorestd{6.4} & 289.7\scorestd{23.3} \\
AVICI & 0.50\scorestd{0.33} & 3.4\scorestd{2.3} & 5.7\scorestd{5.0} & 0.51\scorestd{0.16} & 9.9\scorestd{4.0} & 17.9\scorestd{8.9} & 0.40\scorestd{0.12} & 31.7\scorestd{9.2} & 80.0\scorestd{27.0} & \underline{0.96}\scorestd{0.11} & \underline{0.4}\scorestd{0.7} & \textbf{0.0}\scorestd{0.0} & 0.91\scorestd{0.07} & 2.9\scorestd{2.5} & \underline{0.2}\scorestd{0.6} & 0.81\scorestd{0.08} & \underline{14.7}\scorestd{8.0} & \underline{6.5}\scorestd{6.8} \\
\method{} & \textbf{0.97}\scorestd{0.10} & \textbf{0.2}\scorestd{0.6} & \textbf{0.1}\scorestd{0.4} & \textbf{0.93}\scorestd{0.06} & \textbf{2.3}\scorestd{2.2} & \textbf{1.4}\scorestd{2.1} & \textbf{0.89}\scorestd{0.06} & \textbf{8.9}\scorestd{5.7} & \textbf{6.0}\scorestd{6.3} & \textbf{0.98}\scorestd{0.10} & \textbf{0.2}\scorestd{0.4} & \textbf{0.0}\scorestd{0.0} & \textbf{0.98}\scorestd{0.03} & \textbf{0.7}\scorestd{1.1} & \textbf{0.0}\scorestd{0.1} & \textbf{0.93}\scorestd{0.05} & \textbf{5.7}\scorestd{4.5} & \textbf{0.9}\scorestd{1.5} \\
\bottomrule
\end{tabular}
}
\end{table*}

%% file: figs/tables/main/appendix/table_detailed_pfn.tex
\begin{table*}[htbp]
\centering
\small
\caption{Detailed results for \texttt{pfn} across different graph sizes ($d$) and metrics.}
\label{tab:detail-pfn}
\setlength{\tabcolsep}{4pt}
\resizebox{\textwidth}{!}{
\begin{tabular}{lccccccccc@{\hspace{6pt}}ccccccccc}
\toprule
& \multicolumn{9}{c}{\textbf{Observation-only}} & \multicolumn{9}{c}{\textbf{Mixed-interventional}} \\
\cmidrule(lr){2-10}\cmidrule(lr){11-19}
\textbf{Method} & \multicolumn{3}{c}{$d=5$} & \multicolumn{3}{c}{$d=10$} & \multicolumn{3}{c}{$d=20$} & \multicolumn{3}{c}{$d=5$} & \multicolumn{3}{c}{$d=10$} & \multicolumn{3}{c}{$d=20$} \\
\cmidrule(lr){2-10}\cmidrule(lr){11-19}
 & F1 & SHD & SID & F1 & SHD & SID & F1 & SHD & SID & F1 & SHD & SID & F1 & SHD & SID & F1 & SHD & SID \\
\midrule
RandomRegress & 0.26\scorestd{0.15} & 7.6\scorestd{1.7} & 10.4\scorestd{3.4} & 0.18\scorestd{0.10} & 34.8\scorestd{10.6} & 44.5\scorestd{10.5} & 0.14\scorestd{0.08} & 153.3\scorestd{49.2} & 181.6\scorestd{33.9} & -- & -- & -- & -- & -- & -- & -- & -- & -- \\
DAS & 0.33\scorestd{0.28} & 4.5\scorestd{1.9} & 8.2\scorestd{4.4} & 0.25\scorestd{0.21} & 14.2\scorestd{5.7} & 35.0\scorestd{12.7} & 0.20\scorestd{0.19} & 49.3\scorestd{26.6} & 138.3\scorestd{52.9} & -- & -- & -- & -- & -- & -- & -- & -- & -- \\
LiNGAM & 0.14\scorestd{0.22} & 4.4\scorestd{1.2} & 8.0\scorestd{2.0} & 0.12\scorestd{0.16} & 12.9\scorestd{4.5} & 31.2\scorestd{8.5} & 0.16\scorestd{0.16} & 38.8\scorestd{18.4} & 111.1\scorestd{31.5} & -- & -- & -- & -- & -- & -- & -- & -- & -- \\
PC & 0.49\scorestd{0.28} & 2.8\scorestd{1.3} & 5.8\scorestd{3.0} & 0.44\scorestd{0.19} & 9.2\scorestd{4.1} & 27.0\scorestd{8.3} & 0.32\scorestd{0.16} & 31.6\scorestd{17.7} & 111.7\scorestd{30.1} & -- & -- & -- & -- & -- & -- & -- & -- & -- \\
CDIS & 0.38\scorestd{0.36} & 3.0\scorestd{1.4} & 5.6\scorestd{2.8} & 0.36\scorestd{0.20} & 9.7\scorestd{4.1} & 29.0\scorestd{7.6} & 0.33\scorestd{0.17} & 30.9\scorestd{17.8} & 109.6\scorestd{31.2} & 0.42\scorestd{0.31} & 3.1\scorestd{1.4} & 5.9\scorestd{2.6} & 0.40\scorestd{0.23} & 8.8\scorestd{4.1} & 28.5\scorestd{8.1} & 0.32\scorestd{0.16} & 30.0\scorestd{16.7} & 105.7\scorestd{35.4} \\
GIES & \underline{0.61}\scorestd{0.31} & \underline{2.4}\scorestd{1.8} & \underline{4.0}\scorestd{3.3} & \underline{0.55}\scorestd{0.25} & \underline{8.3}\scorestd{5.8} & \underline{21.3}\scorestd{13.9} & \underline{0.51}\scorestd{0.23} & 31.4\scorestd{25.9} & \underline{83.7}\scorestd{53.3} & \underline{0.80}\scorestd{0.23} & 1.6\scorestd{1.6} & \underline{1.4}\scorestd{2.1} & \underline{0.74}\scorestd{0.22} & 6.4\scorestd{6.6} & \underline{10.7}\scorestd{12.0} & \underline{0.67}\scorestd{0.20} & 24.9\scorestd{24.1} & \underline{49.5}\scorestd{45.0} \\
IGSP & 0.52\scorestd{0.33} & 2.6\scorestd{1.6} & 5.2\scorestd{3.4} & 0.44\scorestd{0.23} & 9.2\scorestd{4.8} & 24.2\scorestd{9.6} & 0.35\scorestd{0.16} & 36.8\scorestd{24.8} & 112.6\scorestd{38.4} & 0.42\scorestd{0.32} & 3.1\scorestd{1.5} & 6.1\scorestd{3.1} & 0.45\scorestd{0.23} & 8.7\scorestd{4.9} & 25.0\scorestd{11.0} & 0.28\scorestd{0.21} & 31.0\scorestd{19.3} & 100.5\scorestd{39.6} \\
DAGMA & 0.30\scorestd{0.32} & 3.7\scorestd{1.6} & 7.6\scorestd{3.7} & 0.21\scorestd{0.26} & 11.9\scorestd{4.6} & 32.3\scorestd{9.3} & 0.22\scorestd{0.26} & 36.8\scorestd{18.6} & 111.9\scorestd{42.3} & -- & -- & -- & -- & -- & -- & -- & -- & -- \\
NOTEARS & 0.22\scorestd{0.31} & 3.7\scorestd{1.3} & 6.6\scorestd{2.5} & 0.18\scorestd{0.23} & 11.2\scorestd{3.9} & 28.0\scorestd{8.4} & 0.18\scorestd{0.20} & 33.9\scorestd{18.1} & 103.5\scorestd{32.6} & -- & -- & -- & -- & -- & -- & -- & -- & -- \\
NOTEARS-MLP & 0.24\scorestd{0.31} & 3.7\scorestd{1.2} & 6.7\scorestd{2.7} & 0.21\scorestd{0.25} & 12.6\scorestd{6.8} & 30.0\scorestd{9.8} & 0.15\scorestd{0.18} & 38.0\scorestd{19.8} & 109.9\scorestd{31.6} & -- & -- & -- & -- & -- & -- & -- & -- & -- \\
NoDAGS & -- & -- & -- & -- & -- & -- & -- & -- & -- & 0.52\scorestd{0.32} & 2.8\scorestd{1.4} & 4.6\scorestd{2.2} & 0.42\scorestd{0.25} & 8.0\scorestd{3.3} & 23.6\scorestd{8.4} & 0.24\scorestd{0.19} & 28.2\scorestd{16.5} & 93.0\scorestd{32.0} \\
SEA & 0.26\scorestd{0.24} & 3.8\scorestd{1.3} & 8.4\scorestd{2.2} & 0.27\scorestd{0.21} & 12.0\scorestd{4.5} & 33.1\scorestd{8.3} & 0.26\scorestd{0.18} & 37.5\scorestd{17.6} & 114.6\scorestd{36.6} & 0.40\scorestd{0.35} & 3.6\scorestd{2.0} & 6.8\scorestd{4.3} & 0.27\scorestd{0.23} & 12.6\scorestd{4.2} & 34.2\scorestd{11.7} & 0.22\scorestd{0.17} & 46.7\scorestd{18.0} & 124.5\scorestd{45.1} \\
SDCD & 0.31\scorestd{0.25} & 4.3\scorestd{1.8} & 8.3\scorestd{3.3} & 0.33\scorestd{0.24} & 12.7\scorestd{6.7} & 33.7\scorestd{12.3} & 0.30\scorestd{0.17} & 52.0\scorestd{29.7} & 135.5\scorestd{49.4} & 0.63\scorestd{0.30} & 2.8\scorestd{2.2} & 4.1\scorestd{4.0} & 0.58\scorestd{0.23} & 9.1\scorestd{5.9} & 19.9\scorestd{12.5} & 0.47\scorestd{0.20} & 34.8\scorestd{18.3} & 83.1\scorestd{43.1} \\
DCDI & 0.34\scorestd{0.02} & 10.0\scorestd{0.1} & 12.7\scorestd{0.5} & 0.22\scorestd{0.05} & 45.0\scorestd{0.0} & 59.8\scorestd{5.3} & 0.16\scorestd{0.08} & 189.9\scorestd{0.4} & 270.3\scorestd{28.6} & 0.35\scorestd{0.04} & 10.0\scorestd{0.0} & 12.7\scorestd{0.5} & 0.21\scorestd{0.05} & 45.0\scorestd{0.0} & 60.4\scorestd{5.8} & 0.15\scorestd{0.07} & 190.0\scorestd{0.0} & 282.3\scorestd{32.7} \\
AVICI & 0.44\scorestd{0.34} & 3.1\scorestd{1.8} & 5.3\scorestd{3.1} & 0.41\scorestd{0.27} & 9.9\scorestd{6.8} & 24.3\scorestd{8.1} & 0.36\scorestd{0.24} & \underline{28.5}\scorestd{17.7} & 96.0\scorestd{35.0} & 0.78\scorestd{0.29} & \underline{1.2}\scorestd{1.4} & 2.4\scorestd{2.6} & 0.69\scorestd{0.30} & \underline{4.6}\scorestd{3.7} & 15.1\scorestd{11.7} & 0.62\scorestd{0.28} & \underline{16.5}\scorestd{14.7} & 65.5\scorestd{41.0} \\
\method{} & \textbf{0.64}\scorestd{0.33} & \textbf{2.0}\scorestd{1.6} & \textbf{3.7}\scorestd{3.0} & \textbf{0.68}\scorestd{0.29} & \textbf{5.1}\scorestd{4.5} & \textbf{16.0}\scorestd{11.4} & \textbf{0.69}\scorestd{0.28} & \textbf{13.8}\scorestd{11.4} & \textbf{66.4}\scorestd{41.5} & \textbf{0.92}\scorestd{0.19} & \textbf{0.4}\scorestd{1.0} & \textbf{1.0}\scorestd{1.9} & \textbf{0.89}\scorestd{0.20} & \textbf{1.6}\scorestd{2.4} & \textbf{7.1}\scorestd{10.8} & \textbf{0.89}\scorestd{0.16} & \textbf{5.2}\scorestd{7.8} & \textbf{23.9}\scorestd{30.4} \\
\bottomrule
\end{tabular}
}
\end{table*}